\documentclass[copyedit,nonblindrev]{informs3}

\usepackage[utf8]{inputenc} %
\usepackage[T1]{fontenc}    %
\usepackage{inconsolata}       %
\usepackage[scaled=.90]{helvet}     %

\usepackage{microtype}    %
\usepackage{enumitem}       %
\usepackage{setspace}
\usepackage[dvipsnames]{xcolor}
\usepackage{xspace}
\usepackage{appendix}

\usepackage{amsmath}
\usepackage{amsfonts}       %
\usepackage{mathtools}
\usepackage{bm}
\usepackage{bbm}
\usepackage{amssymb}
\usepackage{nicefrac}       %

\usepackage{graphicx}
\usepackage[noend]{algpseudocode}
\usepackage{algorithm}
\usepackage{tabularx}       %
\makeatletter
\newcommand{\multiline}[1]{\begin{tabularx}{\dimexpr\linewidth-\ALG@thistlm}[t]{@{}X@{}}#1\end{tabularx}}
\makeatother
\usepackage{verbatim}
\usepackage{booktabs}       %
\usepackage{hhline}
\usepackage{array}
\usepackage{multirow}

\definecolor{DarkBlue}{rgb}{0.0,0.0,0.55}
\PassOptionsToPackage{hypertexnames=false}{hyperref}  %
\usepackage[backref = false, bookmarks, colorlinks = true, plainpages = false, citecolor = DarkBlue , urlcolor = DarkBlue, filecolor = DarkBlue, linkcolor = DarkBlue]{hyperref}
\usepackage{url}            %

\usepackage[nameinlink,capitalize]{cleveref}

\crefformat{equation}{#2(#1)#3}
\Crefformat{equation}{#2(#1)#3}
\Crefformat{figure}{#2Figure #1#3}
\Crefformat{assumption}{#2Assumption #1#3}
\Crefformat{ALC@unique}{#2Line #1#3}

\newcommand{\pref}[1]{\cref{#1}}

\makeatletter
\let\save@mathaccent\mathaccent
\newcommand*\if@single[3]{%
  \setbox0\hbox{${\mathaccent"0362{#1}}^H$}%
  \setbox2\hbox{${\mathaccent"0362{\kern0pt#1}}^H$}%
  \ifdim\ht0=\ht2 #3\else #2\fi
  }
\newcommand*\rel@kern[1]{\kern#1\dimexpr\macc@kerna}
\newcommand*\widebar[1]{\@ifnextchar^{{\wide@bar{#1}{0}}}{\wide@bar{#1}{1}}}
\newcommand*\uwidebar[1]{\@ifnextchar_{{\under@bar{#1}{0}}}{\under@bar{#1}{1}}}
\newcommand*\wide@bar[2]{\if@single{#1}{\wide@bar@{#1}{#2}{1}}{\wide@bar@{#1}{#2}{2}}}
\newcommand*\under@bar[2]{\if@single{#1}{\under@bar@{#1}{#2}{1}}{\under@bar@{#1}{#2}{2}}}
\newcommand*\wide@bar@[3]{%
  \begingroup
  \def\mathaccent##1##2{%
    \let\mathaccent\save@mathaccent
    \if#32 \let\macc@nucleus\first@char \fi
    \setbox\z@\hbox{$\macc@style{\macc@nucleus}_{}$}%
    \setbox\tw@\hbox{$\macc@style{\macc@nucleus}{}_{}$}%
    \dimen@\wd\tw@
    \advance\dimen@-\wd\z@
    \divide\dimen@ 3
    \@tempdima\wd\tw@
    \advance\@tempdima-\scriptspace
    \divide\@tempdima 10
    \advance\dimen@-\@tempdima
    \ifdim\dimen@>\z@ \dimen@0pt\fi
    \rel@kern{0.6}\kern-\dimen@
    \if#31
      \overline{\rel@kern{-0.6}\kern\dimen@\macc@nucleus\rel@kern{0.4}\kern\dimen@}%
      \advance\dimen@0.4\dimexpr\macc@kerna
      \let\final@kern#2%
      \ifdim\dimen@<\z@ \let\final@kern1\fi
      \if\final@kern1 \kern-\dimen@\fi
    \else
      \overline{\rel@kern{-0.6}\kern\dimen@#1}%
    \fi
  }%
  \macc@depth\@ne
  \let\math@bgroup\@empty \let\math@egroup\macc@set@skewchar
  \mathsurround\z@ \frozen@everymath{\mathgroup\macc@group\relax}%
  \macc@set@skewchar\relax
  \let\mathaccentV\macc@nested@a
  \if#31
    \macc@nested@a\relax111{#1}%
  \else
    \def\gobble@till@marker##1\endmarker{}%
    \futurelet\first@char\gobble@till@marker#1\endmarker
    \ifcat\noexpand\first@char A\else
      \def\first@char{}%
    \fi
    \macc@nested@a\relax111{\first@char}%
  \fi
  \endgroup
}
\newcommand*\under@bar@[3]{%
  \begingroup
  \def\mathaccent##1##2{%
    \let\mathaccent\save@mathaccent
    \if#32 \let\macc@nucleus\first@char \fi
    \setbox\z@\hbox{$\macc@style{\macc@nucleus}_{}$}%
    \setbox\tw@\hbox{$\macc@style{\macc@nucleus}{}_{}$}%
    \dimen@\wd\tw@
    \advance\dimen@-\wd\z@
    \divide\dimen@ 3
    \@tempdima\wd\tw@
    \advance\@tempdima-\scriptspace
    \divide\@tempdima 10
    \advance\dimen@-\@tempdima
    \ifdim\dimen@>\z@ \dimen@0pt\fi
    \rel@kern{0.6}\kern-\dimen@
    \if#31
      \underline{\rel@kern{-0.6}\kern\dimen@\macc@nucleus\rel@kern{0.4}\kern\dimen@}%
      \advance\dimen@0.4\dimexpr\macc@kerna
      \let\final@kern#2%
      \ifdim\dimen@<\z@ \let\final@kern1\fi
      \if\final@kern1 \kern-\dimen@\fi
    \else
      \underline{\rel@kern{-0.6}\kern\dimen@#1}%
    \fi
  }%
  \macc@depth\@ne
  \let\math@bgroup\@empty \let\math@egroup\macc@set@skewchar
  \mathsurround\z@ \frozen@everymath{\mathgroup\macc@group\relax}%
  \macc@set@skewchar\relax
  \let\mathaccentV\macc@nested@a
  \if#31
    \macc@nested@a\relax111{#1}%
  \else
    \def\gobble@till@marker##1\endmarker{}%
    \futurelet\first@char\gobble@till@marker#1\endmarker
    \ifcat\noexpand\first@char A\else
      \def\first@char{}%
    \fi
    \macc@nested@a\relax111{\first@char}%
  \fi
  \endgroup
}
\makeatother

\makeatletter
\newcommand{\thickhline}{%
    \noalign {\ifnum 0=`}\fi \hrule height 1pt
    \futurelet \reserved@a \@xhline
}
\newcolumntype{"}{@{\hskip\tabcolsep\vrule width 1pt\hskip\tabcolsep}}
\makeatother
\DeclarePairedDelimiter{\abs}{\lvert}{\rvert}
\DeclarePairedDelimiter{\brk}{[}{]}
\DeclarePairedDelimiter{\crl}{\{}{\}}
\DeclarePairedDelimiter{\prn}{(}{)}

\DeclarePairedDelimiter{\ceil}{\lceil}{\rceil}
\DeclarePairedDelimiter{\floor}{\lfloor}{\rfloor}

\newcommand{\Prob}{\operatorname{\mathbb{P}}}

\newcommand{\wt}[1]{\widetilde{#1}}
\newcommand{\wh}[1]{\widehat{#1}}
\newcommand{\wb}[1]{\widebar{#1}}

\def\ddefloop#1{\ifx\ddefloop#1\else\ddef{#1}\expandafter\ddefloop\fi}
\def\ddef#1{\expandafter\def\csname bb#1\endcsname{\ensuremath{\mathbb{#1}}}}
\ddefloop ABCDEFGHIJKLMNOPQRSTUVWXYZ\ddefloop
\def\ddef#1{\expandafter\def\csname c#1\endcsname{\ensuremath{\mathcal{#1}}}}
\ddefloop ABCDEFGHIJKLMNOPQRSTUVWXYZ\ddefloop
\def\ddef#1{\expandafter\def\csname b#1\endcsname{\ensuremath{\boldsymbol{#1}}}}
\ddefloop ABCDEFGHIJKLMNOPQRSTUVWXYZwxyz\ddefloop
\def\ddef#1{\expandafter\def\csname b#1\endcsname{\ensuremath{\boldsymbol{\csname #1\endcsname}}}}
\ddefloop {alpha}{beta}{gamma}{delta}{epsilon}{varepsilon}{zeta}{theta}{vartheta}{iota}{kappa}{lambda}{mu}{nu}{xi}{pi}{varpi}{rho}{varrho}{sigma}{varsigma}{tau}{upsilon}{phi}{varphi}{chi}{psi}{omega}{Gamma}{Delta}{Theta}{Lambda}{Xi}{Pi}{Sigma}{varSigma}{Upsilon}{Phi}{Psi}{Omega}{ell}\ddefloop

\newcommand{\Ind}{\mathbbm{1}}    %

\newcommand{\eps}{\epsilon}

\newcommand{\ldef}{\vcentcolon=}

\newcommand{\Ber}{\textrm{Ber}}
\newcommand{\BwSC}{\textsf{BwSC}\xspace}
\newcommand{\MAB}{\textsf{MAB}\xspace}
\newcommand{\GRF}{\textsf{GRF}\xspace}
\newcommand{\algoo}{\textsf{LS-SE}\xspace}
\newcommand{\algon}{\textsf{AdaLS}\xspace}
\newcommand{\algog}{\textsf{HS-SE}\xspace}
\newcommand{\algoa}{\textsf{AS-SE}\xspace}
\newcommand{\lbargu}{\textsf{RECAP}\xspace}
\newcommand{\UB}{\textsf{U-BwSC}\xspace}
\newcommand{\GB}{\textsf{G-BwSC}\xspace}
\newcommand{\DB}{\textsf{D-BwSC}\xspace}
\renewcommand{\qed}{\tag*{$\Box$}}

\usepackage{color-edits}
\usepackage[dvipsnames]{xcolor}
\addauthor{yx}{Red}

\usepackage{natbib}
 \bibpunct[, ]{(}{)}{,}{a}{}{,}%
 \def\bibfont{\footnotesize}%

\theoremstyle{TH}       %
\newtheorem{theorem}{Theorem}
\newtheorem{lemma}{Lemma}
\newtheorem{proposition}{Proposition}
\newtheorem{corollary}{Corollary}

\newtheorem{definition}{Definition}
\newtheorem{observation}{Fact}
\newtheorem{example}{Example}
\ECRepeatTheorems

\EquationsNumberedThrough    %

\MANUSCRIPTNO{}

\begin{document}

\TITLE{\Large Phase Transitions in Bandits with Switching Constraints}
\ARTICLEAUTHORS{
\AUTHOR{David Simchi-Levi ~~~~~~~~~~ Yunzong Xu}
\AFF{Institute for Data, Systems, and Society, Massachusetts Institute of Technology, Cambridge, MA 02139\\dslevi@mit.edu, yxu@mit.edu}
}

\date{\today}

\ABSTRACT{
{
We consider the classical stochastic multi-armed bandit problem with a constraint that limits the total cost incurred by switching between actions to be no larger than a given \emph{switching budget}. For this problem, we prove matching upper and lower bounds on the optimal (i.e., minimax) regret, and provide efficient rate-optimal algorithms. Surprisingly, the optimal regret of this problem exhibits a non-conventional growth rate in terms of the time horizon and the number of arms. Consequently, we discover surprising ``phase transitions'' regarding how the \emph{optimal regret rate} changes with respect to the switching budget: when the number of arms is fixed, there are equal-length phases, where the optimal regret rate remains (almost) the same within each phase and exhibits abrupt changes between phases; when the number of arms grows with the time horizon, such abrupt changes become subtler and may disappear, but a generalized notion of phase transitions involving certain new measurements still exists. The results enable us to fully characterize the trade-off between the regret rate and the incurred switching cost in the stochastic multi-armed bandit problem, contributing new insights to this fundamental problem. Under the general switching cost structure, the results reveal interesting connections between bandit problems and graph traversal problems, such as the shortest Hamiltonian path problem.}
}

\maketitle

\section{Introduction}
\label{sec:intro}
The multi-armed bandit (\MAB) problem is one of the most fundamental problems in online machine learning, with diverse applications ranging from pricing and online advertising to clinical trails. Over the past several decades, it has been a very active research area spanning different disciplines, including computer science, operations research, statistics and economics.

In a traditional multi-armed bandit problem, the learner (i.e., decision-maker) is allowed to switch freely between actions, and an effective learning policy may incur frequent switching --- indeed, the learner's task is to balance the exploration-exploitation trade-off, and both exploration (i.e., acquiring new information) and exploitation (i.e., optimizing decisions based on up-to-date information) require switching. However, in many real-world scenarios, it is costly to switch between different alternatives, and a learning policy with limited switching behavior is preferred. The learner thus has to consider the cost of switching in her learning task.

In this paper, we introduce the \emph{Bandits with Switching Constraints} (\BwSC) problem. We note that most previous research in multi-armed bandits has modeled the switching cost as a penalty in the learner's objective, and hence the learner's switching behavior is a complete output of the learning algorithm. However, in many real-world applications, there are strict limits on the learner's switching behavior, which should be modeled as a \emph{hard constraint}, and hence the learner's allowable level of switching is an input to the algorithm. In addition, while most prior research assumes specific structures on switching costs (e.g., unit or homogeneous costs), in reality, switching between different pairs of actions may incur heterogeneous costs that do not follow any parametric form. These gaps motivate us to propose the \BwSC framework, which includes a hard constraint imposed on the total switching cost.

In addition to its strong modeling power and practical significance, the \BwSC problem is theoretically important, as it is a natural framework to study the fundamental trade-off between the best achievable regret rate and the {maximum} incurred switching cost in the classical multi-armed bandit problem. In particular, it enables characterizing important switching patterns associated with any effective exploration-exploitation policies. Thus, the study of the \BwSC problem leads to a series of new results for the classical multi-armed bandit problem.

\subsection{Motivating Examples}
The \BwSC framework has numerous applications, including dynamic pricing, online assortment optimization, online marketplaces, clinical trails, labor markets, supply chain management, etc. We describe some representative examples below. %

\textbf{Dynamic pricing with demand learning.} Dynamic pricing with demand learning has proven its effectiveness in revenue management (\citealt{den2015dynamic}). However, it is well known that in practice, sellers often face business constraints that prevent them from conducting extensive price experimentation and making frequent price changes; see \cite{cheung2017dynamic}, \cite{chen2019parametric} and \cite{chen2020data} for discussions of multiple practical reasons. The seller's sequential decision-making problem can be modeled as a \BwSC problem, where changing from each price to another price incurs some cost, and there is a limit on the total cost incurred by price changes. Here, a high switching cost between two prices implies that the corresponding price change is highly undesirable, while a low switching cost implies that the corresponding price change is generally acceptable.

{\textbf{Promotion {and assortment} strategies in retail and financial services.} Similar to the example of dynamic pricing, many retailers and financial service providers have started to use online learning  techniques to dynamically adjust their promotion strategies (e.g., deals, referral programs, sign-up bonus offers) or product assortments based on sequentially collected data. In many scenarios, frequent changes of public offerings not only increase operational and marketing costs (e.g., inventory and advertising costs) but also lead to customer dissatisfaction and  negative public image (\citealt{simchi2008designing}). 
The sequential promotion planning problem, and the sequential assortment planning problem (with a fixed number of assortment candidates\footnote{Here we refer to the setting that a retailer chooses one assortment from a few assortment candidates (which captures many retailers' practice under complicated business constraints). A different setting in literature allows one to consider exponentially many assortment candidates under an MNL choice model, see \cite{agrawal2019mnl,dong2020multinomial}.}), can both be modeled as \BwSC.}

{\textbf{Sequential experiments in online marketplaces.} Consider an online e-commerce platform (e.g., Uber, Airbnb) choosing a  mechanism (e.g., a surge pricing algorithm, a listing ranking rule) among several alternatives. %
It is common practice for platforms to  conduct sequential experiments of mechanisms %
using bandit approaches to optimize long-term revenue. However, frequent changes of mechanisms may be highly undesirable for a marketplace, because not only the platform (e.g., Uber) but also the \emph{market participants} (e.g., drivers and riders) may suffer from switching costs: each time the platform  announces a new mechanism, the market participants will make efforts to adapt to the new mechanism (e.g., if Uber announces that trips completed during select hours each day earn extra rewards, then a driver may be incentivized to change his work schedule); as a result, market participants will get annoyed when they find that the mechanism changes frequently (which means that they have to re-develop their business strategies frequently). Therefore, platforms usually have to limit their number of mechanism changes in sequential experiments.
}

\subsection{Problem Formulation}\label{sec:model}\label{sec:formulation}
We now introduce our model. Consider a $K$-armed bandit problem where a learner chooses actions from a fixed set $[K]=\{1,\dots,K\}$. There is a total of $T$ rounds ($T\ge K$). In each round $t\in[T]$, the learner first chooses an action $a_t\in[K]$, then observes and collects a reward $X^t(a_t)\in\mathbb{R}$. For each action $k\in[K]$, the reward of action $k$ is i.i.d. drawn from an (unknown) distribution $\cD_k$ with (unknown) expected value $\mu_k$. We assume that the distributions $\cD_1,\dots,\cD_K$ are standardized sub-Gaussian.\footnote{This is a standard assumption in the stochastic bandit literature. Note that the class of sub-Gaussian distributions is sufficiently wide as it contains Gaussian, Bernoulli and all bounded distributions.} 
Without loss of generality, we assume $\sup_{i,j\in[K]}|\mu_i-\mu_j|\in[0,1]$.

In the \BwSC problem, the learner incurs a switching cost $c_{i,j}\ge0$ each time she switches from action $i$ to action $j$ ($i,j\in[K]$).\footnote{We allow $c_{i,j}=\infty$, which means that switching from $i$ to $j$ is prohibited. We also allow $c_{i,j}\ne c_{j,i}$, which means that the switching costs are asymmetric.} In particular, $c_{i,i}=0$ for $i\in[K]$. There is a pre-specified \textit{switching budget} $S\ge0$ representing the maximum amount of switching costs that the learner can incur in total. Once the total switching cost exceeds the switching budget $S$, the learner cannot switch her actions any more. The learner's goal is to maximize the expected total reward over $T$ rounds. %

\subsubsection{Admissible Policies}%

Let $\pi$ denote the learner's (non-anticipating) learning policy; specifically, $\pi$ is a sequence  $\prn{\pi_1,\dots,\pi_T}$, where $\pi_t$ establishes a probability kernel acting from the (measurable) space of historical actions and observations before round $t$  to the (measurable) space of actions at round $t$. Let $a_t$ denote the (random) action selected by policy $\pi$ at round $t$, and $X^t(a_t)$ denote the (random) reward observed by policy $\pi$ at round $t$ (note that both $a_t$ and $X^t(a_t)$ depend on the underlying distributions $\cD=(\cD_1,\dots,\cD_k)$). 
Let $\mathbb{P}^\pi_{\cD}$  be the probability measure induced by  the random variables ${\prn*{a_1,X^t(a_1)},\dots,\prn*{a_T,X^T(a_T)}}$, and $\bbE^\pi_\cD[\cdot]$ be the associated expectation operator. 

According to our model, we only need to restrict our attention to the \emph{$S$-switching-budget} policies, which take $S$, $K$ and $T$ as input and are defined below.\footnote{Note that here we do not make any assumption on the learner's behavior. In particular, we do not require the learner to intentionally pick an $S$-switching-budget policy --- the switching constraint makes the learner's policy automatically equivalent to an $S$-switching-budget policy.}
\begin{definition}%
A policy $\pi$ is said to be an $S$-switching-budget policy if for all $\cD$,
\[
\mathbb{P}_{\cD}^\pi\left[\sum_{t=1}^{T-1}c_{a_t,a_{t+1}}\le S\right]=1.
\]
\end{definition}
Let $\Pi_S$ denote the set of all $S$-switching-budget policies, which is also the admissible policy class of the \BwSC problem.

\subsubsection{Regret and Optimal Regret}\label{sec:regret}
The performance of a learning policy is measured against a clairvoyant policy that maximizes the expected total reward given foreknowledge of the \emph{environment} (i.e., underlying distributions) $\cD$. Let $k^*=\arg\max_{k\in[K]}\mu_i$ and $\mu^*=\max_{k\in[K]}\mu_k$. If a clairvoyant knows $\cD$ in advance, then she would choose the ``optimal'' action $k^*$ for every round and her expected total reward would be $T\mu^*$. We define the \emph{regret} of policy $\pi$ as the worst-case difference between the expected performance of the optimal clairvoyant policy and the expected performance of policy $\pi$:
\[
R^\pi(T)\ldef\sup_{\cD}\left\{T\mu^*-\mathbb{E}_{\cD}^\pi\left[\sum_{t=1}^T X^t({a_t})\right]\right\}=\sup_{\cD}\left\{T\mu^*-\mathbb{E}_{\cD}^\pi\left[\sum_{t=1}^T \mu_{a_t}\right]\right\},
\]
which is a non-negative function of the policy $\pi$, the number of actions $K$, and the \emph{horizon} (i.e., number of rounds) $T$;  occasionally, we will use the notation $R^\pi(K,T)$ to highlight its dependence on $K$. Furthermore, the \emph{optimal} (i.e., minimax) regret of \BwSC is defined as \[R_{S}^*(T)\ldef\inf_{\pi\in\Pi_S}R^\pi(T),\]
which is a non-negative function of the switching budget $S$, the number of actions $K$, and the horizon $T$; occasionally, we will use the notation $R_{S}^*(K,T)$ to highlight its dependence on $K$. 
Note that the optimal regret is an intrinsic quantity that can help us to characterize the  statistical complexity of the \BwSC problem.

\textbf{Remark.} There are two notions of regret in the stochastic bandit literature. The $R^\pi(T)$ regret that we consider is called the \emph{distribution-independent} (or \emph{worst-case}) regret, as it does not depend on $\cD$. On the other hand, one can also define the \textit{distribution-dependent} (or \emph{instance-dependent}) regret $R_{\cD}^\pi(T)=T\mu^*-\mathbb{E}_{\cD}^\pi\left[\sum_{t=1}^T \mu_{a_t}\right]$ that depends on $\cD$. %
Unlike the classical \MAB problem where there are policies that simultaneously achieve near-optimal bounds under both regret notions, in the \BwSC problem, due to the limited switching budget, finding a policy that simultaneously achieves near-optimal bounds under both regret notions is usually impossible. Thus in the main body of the paper, we focus on the distribution-independent regret. However, in Appendix \ref{app:dd}, we extend our results to the distribution-dependent regret.

\subsubsection{Research Questions}\label{sec:bwsc-mab}

Two fundamental tasks in the study of bandits are: (i) to understand the \emph{growth rate} of the optimal regret (i.e., ``optimal regret rate'') as $T$ grows, or as both $K$ and $T$ grow, and (ii) to design efficient algorithms that attain near-optimal regret. In this paper, we seek to address both of these challenges for \BwSC. Moreover, motivated by the relationship between \BwSC and \MAB,\footnote{Note that \BwSC and \MAB share the same definition of $R^\pi(T)$, and the only difference between \BwSC and \MAB is the existence of a switching constraint $\pi\in\Pi_S$, determined by $(c_{i,j})\in\overline{\mathbb{R}}_{\ge0}^{K\times K}$ and $S\in\overline{\mathbb{R}}_{\ge0}$ (when $S=\infty$, \BwSC degenerates to \MAB).} we seek to understand how the switching constraint fundamentally affects the statistical nature of bandits. Altogether, our central questions are:
\begin{enumerate}
    \item What is the statistical complexity (i.e., optimal regret rate) of \BwSC?
    \item Can we design practical algorithms to attain the optimal regret rate?
    \item How does the optimal regret rate of \BwSC changes with respect to the switching budget $S$, and how is it affected by the structure of switching costs $(c_{i,j})$?
\end{enumerate}

\subsection{Main Results and Technical Highlights}\label{sec:results}

The main contributions of this paper lie in  fully addressing the above three research questions for \BwSC under the unit switching cost structure,  partially addressing the three questions for \BwSC under the general switching cost structure, and  discovering surprising ``phase transition'' behavior of the optimal regret (under both unit and general switching cost structures). We devise a series of efficient algorithms which attain sharp regret upper bounds, and introduce a highly non-trivial five-step method which provides matching (or nearly matching) lower bounds.
As a by-product, we develop a novel information-theoretic inequality, namely the \emph{Generalized Reverse Fano-type} inequality, which plays a critical role in our five-step method.

We summarize our main results and technical contributions as follows.

\textbf{Effective algorithms for the \text{U-BwSC} problem.} We first study the \BwSC problem under the most fundamental switching cost structure --- the unit switching cost structure: $c_{i,j}=1$ for all $i\ne j$. The problem is referred to as the \emph{unit-switching-cost}  \BwSC problem (or \textsf{U-BwSC} for short), and can be interpreted as ``\MAB with limited number of switches.'' As a preliminary attempt, we present a simple and intuitive algorithm, called \algoo, which builds on the ``batched elimination'' framework recently developed by \cite{perchet2016batched} and \cite{gao2019batched}, and ensures the following regret:
    \begin{equation}\label{eq:algoo-reg}
    \wt{\cO}(1)\cdot K^{1-\frac{1}{2-2^{-q(S,K)}}}T^{\frac{1}{2-2^{-q(S,K)}}},
    \end{equation}
where $q(S,K)\ldef\floor*{\frac{S-1}{K-1}}$ is the \emph{quotient} of the \emph{Euclidian division} of $(S-1)$ by $(K-1)$.\footnote{See \href{https://en.wikipedia.org/wiki/Euclidean_division}{https://en.wikipedia.org/wiki/Euclidean\_division} for a definition of the Euclidian division. We use $\floor{\cdot}$ to denote the floor function; see \cref{sec:notations} for details.}  

{The \algoo algorithm, though being very simple, has several drawbacks, including a potentially large waste of the switching budget, and overly low \emph{adaptivity} (i.e., it learns from data in an overly infrequent manner; see \cref{sec:improved}). To overcome these drawbacks, we design a new algorithm, \algon, which builds on and improves upon \algoo by (i) adopting a novel combinatorial and randomized exploration strategy and (ii) deciding when to make switches in a more data-driven fashion. %
These two features enable \algon to make better use of the switching budget and enjoy higher adaptivity.} 

We show that \algon attains an improved regret bound of
    \begin{equation}\label{eq:algon-reg}
    \wt{\cO}(1)\cdot \max\crl*{\frac{\prn*{K-{r}(S,K)}^{2-\frac{1}{2-2^{-q(S,K)}}}}{K}T^{\frac{1}{2-2^{-q(S,K)}}},\ {{K}^{1-\frac{1}{2-2^{-q(S,K)-1}}}}T^{\frac{1}{2-2^{-q(S,K)-1}}}},
    \end{equation}
where $r(S,K)\ldef(S-1)\%(K-1)$ is the \emph{remainder} of the \emph{Euclidean division} of $(S-1)$ by $(K-1)$.\footnote{We use $\%$ to denote the modulo operation; see \cref{sec:notations} for details.} 
{Since $0\le r(S,K)\le K-2$, the rate of \pref{eq:algon-reg} is at most $K^{1-\frac{1}{2-2^{-q(S,K)}}}T^{\frac{1}{2-2^{-q(S,K)}}}$, which is the same as \pref{eq:algoo-reg}, and at least $\max\crl*{K^{-1}T^{\frac{1}{2-2^{-q(S,K)}}}, K^{1-\frac{1}{2-2^{-q(S,K)-1}}}T^{\frac{1}{2-2^{-q(S,K)-1}}}}$, which can be much smaller than \pref{eq:algoo-reg} when $K$ is large, as the first term in ``$\max$'' has a sharp $K^{-1}$ factor and the second term in ``$\max$'' has a smaller order of $T$.
This implies that \algon reduces the regret of \algoo by a multiplicative factor of at least $\Omega(1)$ and at most ${\cO}(K^{3/2})$. See  \pref{tb:version1} for detailed illustrations and comparisons of regret bounds \pref{eq:algoo-reg} and \pref{eq:algon-reg}, and \pref{sec:improved} for more explanations.}

\textbf{Tight lower bound for the U-BwSC problem.} The \algon algorithm, though being a significantly refined version of the \algoo algorithm, still {seems to} leave plenty of room for improvement. In particular, it only improves the regret rate when $K$ is permitted to grow with $T$, failing to {directly} improve the regret's dependence on the most important parameter $T$ when $K=\wt\cO(1)$. Several challenging questions remain open: 
Is it possible to directly improve the regret in terms of $T$? Can the dependence on $K$ be further improved? What is the fundamental limit of the  \textsf{U-BwSC} problem? We settle these questions by establishing a strong (and quite surprising) information-theoretic lower bound that directly match the upper bound \cref{eq:algon-reg} for any $S$, any $K$, and any $T$ --- specifically, we show that \emph{no} admissible policy can avoid a regret lower bound of
\begin{equation}\label{eq:ubwsc-lb}
    \wt{\Omega}(1)\cdot \max\crl*{\frac{\prn*{K-{r}(S,K)}^{2-\frac{1}{2-2^{-q(S,K)}}}}{K}T^{\frac{1}{2-2^{-q(S,K)}}},\ {{K}^{1-\frac{1}{2-2^{-q(S,K)-1}}}}T^{\frac{1}{2-2^{-q(S,K)-1}}}},
\end{equation}
which implies that \algon is optimal up to logarithmic factors. The proof of the lower bound is highly non-trivial. The methodological contributions in the lower bound proof will be elaborated shortly.

 The tight lower bound proved for \textsf{U-BwSC} also motivate new insights about the classical \MAB. In particular, it implies that $\Omega(K\log\log T)$ switches are \textit{necessary} to achieve  $\wt{\cO}(\sqrt{KT})$ (near-optimal) regret in \MAB, which appears to be a fundamental yet new result.

\textbf{Phase transitions associated with the optimal regret.} Combining \cref{eq:algon-reg} and \cref{eq:ubwsc-lb}, we completely characterize the optimal regret  of the \UB problem. {The characterization reveals surprising findings:  when $K=\wt\cO(1)$, the quotient function $q(S,K)$ (as a floor function) uniquely determines the optimal regret rate; when  $K$ grows with $T$ in a non-negligible way, the remainder function $r(S,K)$ also affects the optimal regret rate through the dependence on $K$, but only when $r(S,K)$ is large enough such that $r(S,K)=K-o(K)$.}  
To the best of knowledge, this is the \emph{first} example of an online learning setting where (i) a floor function naturally arises in the exponent of $T$ in the optimal regret, and (ii) the optimal regret exhibits a ``combinatorial'' growth rate which is surprisingly characterized by an Euclidean division. %

{As a consequence of these findings, we discover surprising \emph{phase transitions} regarding how the optimal regret rate changes with respect to the switching budget $S$, which can be summarized by the following two cases: when $K$ is fixed (and $T$ grows), there are equal-length phases defined by $S$, where the optimal regret rate remains the same (up to logarithmic factors) within each phase and exhibits abrupt changes between phases; when $K$ grows with $T$ in a non-negligible manner, such abrupt changes become subtler and may disappear, but a  generalized form of phase transitions involving the ``budget-to-arm ratio'' (BAR) still exist. We will provide a rigorous and detailed treatments of phase transitions in \cref{sec:sbrto}.}

\textbf{Extensions to general switching cost structures.} We extend the results obtained in the \UB problem to the general \BwSC problem. Specifically, we study two types of general switching cost structures (with one being symmetric and the other being asymmetric): (i) the \emph{general symmetric switching cost} structure (corresponding to the \GB problem), where $c_{i,j}=c_{j,i}$ $(i\ne j)$ can be any non-negative real number; and (ii) the \emph{departure cost} structure (corresponding to the \DB problem), where $c_{i,j}=c_i$ for all $j\ne i$ and $c_i$ can be any non-negative real number, i.e., the  switching cost between any pair of actions only depends on  the action that the learner departs from. For both \GB and \DB, we design efficient algorithms, and prove corresponding lower bounds on regret. Under the condition of $K=\wt{\cO}(1)$, we show that our regret upper and lower bounds almost match for \GB, and exactly match for \DB (both in terms of the dependence on $T$); the optimal regret again exhibits \emph{phase transitions}. %
Our results in this part make conceptual contributions by revealing an interesting connection between  bandit problems and \emph{graph traversal} problems. %

\textbf{Methodological contributions in the lower bound analysis.}  As we mentioned, the proof of the lower bound \cref{eq:ubwsc-lb} requires significant technical effort, and contains several technical highlights of this paper. In particular, to show that the  quotient function $q(S,K)$ \emph{necessarily} appear in the exponent of $T$  and the remainder function $r(S,K)$  \emph{necessarily} affects the order of $K$ through the $(K-r(S,K))$ term, we develop a host of new techniques, largely from first principles, to characterize how the switching constraint  affects the learning dynamics of an \emph{arbitrary} admissible policy, and how concrete classes of certain learning dynamics (represented by  \emph{risky events}) lead to fundamental performance limits. These techniques combine ideas from information theory, probability theory, statistics and combinatorics, and are integrated in a five-step proof program called \lbargu.

As an aside, we establish a novel and powerful information-theoretic inequality, namely the \emph{Generalized Reverse Fano-type} (\GRF) inequality, which is of independent interest. The inequality overcomes several limitations of the classical \emph{Fano's} inequality and its  generalizations by (i) acting in a reverse direction (or equivalently, providing {sharp} lower bounds on the average probability of multiple ``low-probability'' events), (ii) working with arbitrary events (not necessarily forming a partition), (iii) working with arbitrary measures (not necessarily on a same measurable space), (iv) working with the \emph{reverse KL divergence}, and (v) allowing arbitrary choices of  ``refrence'' measures that could vary with events. Our lower bound proof clearly demonstrates the advantages of the \GRF inequality: all of the above five features play critical roles in the \lbargu method.

We believe that the \GRF inequality, together with the  ideas and techniques arising in the \lbargu method, can find broader applications in learning theory, information theory, and mathematical statistics. For this reason, we will use a separate section (\cref{sec:methodology}) in this paper to introduce the \GRF inequality and present the \lbargu method in detail.

\subsubsection{Comparison with Prior Work}\label{sec:comparison}
We provide a detailed comparison of our results and some closely related results in prior literature, including those obtained in the literature on ``batched bandits''  (\citealt{perchet2016batched,gao2019batched}) and those presented in a preliminary version of this paper (\citealt{simchi2019phase}). %

\textbf{Comparison with the results on batched bandits.}\label{sec:batch} The \UB problem is related to the ``batched (multi-armed) bandit'' problem  (\citealt{perchet2016batched,gao2019batched}). The $M$-batched bandit problem is defined as follows: given a classical \MAB, assumes that the learner must split her learning process into $M$ batches and is only able to observe data (i.e., realized rewards) from a given batch after the entire batch is completed. This implies that all actions within a batch are determined at the beginning of this batch. Here $M$ can be viewed as a quantity measuring the learner's \textit{adaptivity}, i.e., her ability to learn from her data and adapt to the environment. An $M$-batch policy is defined as a policy that only observes the available data for $M-1$ times through the entire horizon. \cite{perchet2016batched} study the above problem in the two-armed case. They propose an $M$-batch policy with $\wt{\Theta}\prn*{T^{\frac{1}{1-2^{1-M}}}}$ regret, and show that no $M$-batch policy can attain a better regret rate under the ``static grid'' restriction which requires the policy to \emph{pre-determine} the batch sizes before the learning process. \cite{gao2019batched} extend their algorithm and results to the general $K$-armed case, and show that even without the ``static grid'' restriction (i.e., even when the batch sizes can be \emph{adaptively chosen} in a batch-by-batch manner), no $M$-batch policy can attain an better regret rate. The statistical complexity of the batched bandit problem is thus completely characterized.

The batched bandit problem and the \UB problem are two closely related but fundamentally different problems: while the batched bandit problem explicitly limits the number of times of making observations (i.e., adaptivity), the \UB problem only limits the number of times of action changes, and (importantly!) \emph{allows unlimited number of times of making observations}. As we shall see in \pref{sec:lsla}, the batched bandit model is strictly more restricted than the \UB model (i.e., the \UB model is more relaxed than the batched bandit model), in the sense that the admissible policy class of \UB is much richer and contains more efficient policies. This difference allows \UB to enjoy a fundamentally smaller optimal regret rate when $K$ is large; see \pref{sec:lsla} for a detailed discussion on the relationship and difference of the two problems.

We note that existing results and techniques of batched bandits {cannot} provide a satisfying solution to the \UB problem, neither in terms of designing rate-optimal algorithms nor in terms of  establishing fundamental limits. While it is relatively easier to obtain an $S$-switching-budget policy by modifying a $(q(S,K)+1)$-batch policy (see \pref{sec:ub} and the \algoo algorithm), such approach suffers from several drawbacks and is generally sub-optimal when $K$ is large, as an $S$-switching-budget policy can utilize data more frequently than a $(q(S,K)+1)$-batch policy and achieve better regret (see \pref{sec:improved}). As a result, we have to develop new algorithmic ideas to design a more advanced algorithm (the \algon algorithm) to achieve the optimal regret rate of the \UB problem. 

{More importantly, since \UB is more relaxed than the batched bandit problem, a regret lower bound for the batched bandit problem cannot imply a regret lower bound for \UB. Therefore, we need to establish  new lower bounds for \UB, which also imply lower bounds for the batched bandit problem.  
In fact, from an information-theoretic perspective, when we aim for lower bounds, dealing with a switching constraint (which does not impose any constraint on the number of queries of information) is considerably more challenging than dealing with a batch constraint (which directly restricts the ability to inquire information): %
the challenge is that an $S$-switching-budget policy can continuously gain new information at every round; as a result, it may be difficult to establish sharp impossibility results for such a policy via standard information-theoretic arguments}. We address this challenge by establishing the \lbargu method from first principles. We remark that the lower bound results in our paper are strong. Indeed, both the lower bounds of \cite{perchet2016batched} and \cite{gao2019batched} can be seen as %
corollaries of our lower bound \pref{eq:ubwsc-lb}; see \pref{sec:lsla}. %

\textbf{Advances over the  conference version.}  We would like to point out that a preliminary version of the current paper, \cite{simchi2019phase}, appeared in the 33rd Conference on Neural Information Processing Systems (NeurIPS 2019), and the current paper is a significantly enhanced version of it. In particular, the conference version only studies the \BwSC problem in the case of $K=\wt\cO(1)$, where one only needs to care about the regret's dependence on $T$ and does not need to care about the regret's dependence on $K$. As a result, the conference version  leaves an $\cO(K^{5/2})$ gap between the upper and lower bounds for the \UB problem. The current paper significantly improves upon \cite{simchi2019phase} in the following aspects:
\begin{enumerate}
    \item We propose a new algorithm (\algon) for the \UB problem, which improves the regret upper bound in the conference version by a multiplicative factor of at least $\Omega(1)$ and at most $\cO(K^{3/2})$.
    \item We develop a new lower bound proof program (\lbargu) which builds on the lower bound proof ideas of the conference version. The new proof program is more powerful and strengthens  the \UB lower bound in the conference version by a multiplicative factor of at least $\Omega(K)$ and at most $\cO(K^{5/2})$. Combining the new upper and lower bounds, we close the  $\cO(K^{5/2})$ gap in  \cite{simchi2019phase} and completely characterize the optimal regret's dependence on $K$ for the \UB problem. In developing the \lbargu method, we introduce a new information-theoretic inequality (the \GRF inequality) which is of independent interest.
    \item We present a more precise description of phase transitions in different regimes. 
    \item In \pref{sec:general}, we study the \DB problem which is uncovered in the conference version.
\end{enumerate}

\subsection{Notations and Organization}\label{sec:notations}

Let $\bbN$ (resp. $\bbN_{>0}$) be the set of all non-negative (resp. positive) integers. 
For all $n_1,n_2\in\mathbb{N}$ such that $n_1\le n_2$, we use $[n_1]$ to denote the set $\{1,\dots,n_1\}$, and use $[n_1:n_2]$ (resp. $(n_1:n_2]$) to denote the set $\{n_1,n_1+1,\dots,n_2\}$ (resp. $\{n_1+1,\dots,n_2\}$). For all $x\ge0$, we use $\lfloor x\rfloor$ to denote the largest integer less than or equal to $x$. For ease of presentation, we define $\lfloor x\rfloor=0$ for all $x<0$. For all $m\in\bbN, n\in\bbN_{>0}$, we define $m\%n\ldef m-n\floor*{m/n}$ (i.e., the remainder of the Euclidean division of $m$ by $n$). For all $m,n\in\bbR$, let $m \vee n\ldef \max\crl{m,n}$ and $m \wedge n\ldef \min\crl{m,n}$. Throughout the paper, we adopt non-asymptotic big-oh notation: for functions
	$f,g:\cX\to\bbR_{+}$, we write $f=\cO(g)$ (resp. $f=\Omega(g)$) if there exists some constant
	$C>0$ such that $f(x)\leq{}Cg(x)$ (resp. $f(x)\geq{}Cg(x)$)
        for all $x\in\cX$. We write $f=\wt{\cO}(g)$ if
        $f=\cO(g\cdot\mathrm{polylog}(T))$, $f=\wt{\Omega}(g)$ if $f=\Omega(g/\textrm{polylog}(T))$, and
        $f=\wt{\Theta}(g)$ if $f=\wt{\cO}(g)$ and $f=\wt{\Omega}(g)$. %
	We use $f\asymp g$ as shorthand for $f=\wt{\Theta}(g)$. We write $f=o(g)$ if $\lim_{x\rightarrow\infty}{f(x)}/{g(x)}=0$.

The rest of the paper is organized as follows. In \cref{sec:litrev}, we review other related literature. In \cref{sec:unit}, we discuss the unit-switching-cost model. In \cref{sec:general}, we discuss two general-switching-cost models. In \cref{sec:methodology}, we introduce the \GRF inequality and our lower bound proof method. We discuss some natural extensions in   \cref{sec:conclusion}.

\section{Related Literature}\label{sec:litrev}
\vspace{0.5em}

\subsection{Stochastic MAB with Switching Costs}
The stochastic \MAB problem has been extensively studied for more than fifty years. It is well known that the optimal distribution-dependent regret is  $\Theta(K\log T)$  (\citealt{lai1985asymptotically}) and the optimal distribution-independent regret is  $\Theta(\sqrt{KT})$ (\citealt{auer2002finite}). We point out two excellent surveys  \cite{lattimore2020bandit} and \cite{slivkins2019introduction} for more reference about this topic.

There is rich literature focusing on stochastic \MAB with switching costs.\footnote{It is worth noting that there is also a vast literature on \emph{adversarial} \MAB with switching costs. In particular, \cite{dekel2014bandits} prove a striking $\wt{\Omega}(K^{1/3}T^{2/3})$ lower bound for this problem, indicating a fundamental difference between the roles of switching costs in stochastic \MAB and in adversarial \MAB.}  Most of the papers model the switching cost as a penalty in the learner's objective, i.e., they  measure a policy's regret and incurred switching cost using the same metric and the objective is to minimize the sum of these two terms (e.g., \citealt{agrawal1988asymptotically,agrawal1990multi,brezzi2002optimal,cesa2013online}; there are other variations with discounted rewards \citealt{banks1994switching,asawa1996multi,bergemann2001stationary}, see \citealt{jun2004survey} for a survey). Though this conventional ``switching penalty'' model has attracted significant research interest in the past, it has two limitations. First, under this model, the learner's total switching cost is an output determined by the algorithm. However, in many real-world applications, there are strict limits on the learner's total switching cost, which should be modeled as a \textit{hard constraint}, and hence the learner's switching budget should be an input that helps determine the algorithm. In particular, while the  algorithm in \cite{cesa2013online} developed for the ``switching penalty'' model can  achieve $\wt{\cO}(\sqrt{KT})$ (near-optimal) regret with $\cO(K\log\log T)$ switches, if the learner wants a policy that always incurs finite switching cost independent of $T$, then prior literature does not provide an answer. Second, the ``switching penalty'' model has fundamental weakness in studying the trade-off between the regret rate and the incurred switching cost in stochastic \MAB{} ---  since the $\log\log T$-type bound on the incurred switching cost of a policy is negligible compared with the $\sqrt{T}$-type bound on its best achievable regret, when adding the two terms up, the term associated with incurred switching cost is always dominated by the regret (in terms of the growth rate), thus no trade-off can be identified. %
As a result, to the best of our knowledge, prior literature has not characterized the fundamental trade-off between the regret rate and the incurred switching cost in stochastic \MAB.

The \BwSC framework addresses the issues associated with the ``switching penalty'' model in several ways. First, it introduces a hard constraint on the total switching cost, enabling us to design good policies that {guarantee} limited switching cost. While $\cO(K\log\log T)$ switches has proven to be sufficient for a learning policy to achieve near-optimal regret in \MAB, in \BwSC, we are mostly interested in the setting of finite or $o(K\log\log T)$ switching budget, which is highly relevant in practice. Second, by focusing on rewards in the objective function and incurred switching cost in the switching constraint, the \BwSC framework enables the characterization of the fundamental trade-off between regret and {maximum} incurred switching cost in \MAB. Third, while most prior research assumes specific structures on switching costs (e.g., unit or homogeneous costs), \BwSC allows general switching costs, which makes it a powerful modeling framework.

\subsection{Online Learning with Limited Switches.}
This paper is not the first one to study online learning problems with limited switches.\footnote{Here, ``online learning with limited switches'' refers to online learning problems with constraints on the \emph{learner}'s number of changes of \emph{decisions}. There is another line of research studying (non-stationary) online learning problems with constraints on \emph{nature}'s number of changes of \emph{environments} (e.g., \citealt{herbster1998tracking,jun2017online}). Though this line of research focuses on completely different learning challenges (i.e., non-stationarity), it is conceptually relevant as it shares the same flavor of characterizing the regret using a ``budget for making changes.''} In Indeed, a few authors have realized the practical significance of limited switching budget. For example, \cite{cheung2017dynamic} consider a dynamic pricing model where the demand function is unknown but belongs to a known finite set, and a pricing policy is allowed to make at most $m$ price changes. Their constraint on the total number of price changes is motivated by collaboration with Groupon, a major e-commerce marketplace in North America. In such an environment, Groupon limits the number of price changes,  either because of implementation constraints, or for fear of confusing customers and receiving negative customer feedback. They propose a pricing policy that guarantees $\cO(\log^{(m)}T)$ (or $m$ iterations of
the logarithm) regret with at most $m$ price changes, and report that in a field experiment, this pricing policy with a single price change increases revenue and market share significantly. \cite{chen2019parametric} study the joint pricing and inventory control problem with unknown demand and limited price changes. Assuming that the demand function is drawn from a parametric class of functions, they develop a finite-price-change policy based on maximum likelihood estimation (MLE) that achieves the optimal regret rate. \cite{chen2020data} also study the dynamic pricing and inventory control problem with limited price changes, but in a more challenging setting with \emph{censored} demand. They prove matching upper and lower bounds on the optimal regret, and devise an MLE-based policy to achieve the optimal regret rate. 

We note that all of \cite{cheung2017dynamic,chen2019parametric,chen2020data} focus on specific revenue management problems, and their results rely on certain assumptions that are specialized to their models. %
The \BwSC model in our paper has a different flavor, in the sense that it is very generic and requires very few assumptions. The results and techniques of our paper are thus very different from the above papers. Also, the switching constraint in the \BwSC problem is more general than the price-change constraints in previous models.

In the Bayesian bandit setting, \cite{guha2013approximation} (see also \citealt{guha2009multi} for a conference version) study the ``bandits with metric switching costs'' problem that allows a constraint involving metric switching costs. Using competitive ratio  as the performance metric and assuming Bayesian priors, they develop a 4-approximation algorithm for the problem. The competitive ratio is measured against an optimal online policy that does not know the true distributions. As pointed out by the authors, the optimal online policy can be directly determined by a dynamic program, so the main challenge in their model is a computational one. Our work is different, as we are using regret as our performance metric, and we are competing with an optimal clairvoyant policy that knows the true distributions --- a much stronger benchmark. Our problem thus involves both statistical and computational challenges. In fact, the algorithm in \cite{guha2013approximation} cannot avoid a linear regret when applied to the \BwSC problem.

In the adversarial bandit setting, \cite{altschuler2018online} study the adversarial \MAB problem with limited number of switches, which can be viewed as an adversarial counterpart of the \UB problem. For any policy that makes no more than $S\le T$ switches, they prove that the optimal regret is $\wt{\Theta}(T\sqrt{K}/\sqrt{S})$. Since we are considering a different setting from them (our problem is stochastic while their problem is adversarial), the results and techniques in our paper are fundamentally different from their paper. In particular, while any fixed-switching-budget policy cannot avoid linear regret in the adversarial setting, in the stochastic setting, a fixed number of switches may already be able to guarantee sublinear regret (assuming $K$ is fixed). Moreover, while the optimal regret rate in \cite{altschuler2018online} decreases smoothly as $S$ increases from $0$ to $T$, in the stochastic setting, we identify surprising behavior of the optimal regret rate  as $S$ increases from $0$ to $\Theta(K\log\log T)$, which, to the best of our knowledge, has not been identified in the bandit literature before.

\section{Unit Switching Costs}\label{sec:unit}
In this section, we consider the unit-switching-cost \BwSC problem (abbreviated as \UB), where $c_{i,j}=1$ for all $i\ne j$. In this case, since every switch incurs a unit cost, the switching budget $S$ can be interpreted as the maximum number of switches that the learner can make in total. Without loss of generality, in this section we assume that $S$ is a non-negative integer, and refer to an $S$-switching-budget policy as an $S$-switch policy. Note that the \UB problem can be simply interpreted as ``\MAB with limited number of switches.''

The section is organized as follows. In \cref{sec:ub}, we present a simple and intuitive algorithm and an initial upper bound on regret. 
In \cref{sec:improved}, we propose a refined algorithm that attains an improved upper bound on regret. 
In \cref{sec:lb}, we establish a matching lower bound on regret, indicating that the algorithm in \cref{sec:improved} is rate-optimal. In \cref{sec:sbrto}, we discuss several surprising findings in \UB, namely ``phase transitions'' of the optimal regret.
In \cref{sec:lsla}, we discuss the relationship between limited switches and limited adaptivity in bandit problems. %

\textbf{Algorithmic notations.} We adopt the following notations to facilitate the descriptions of our algorithms. For any execution of an algorithm, for any action $i\in[K]$, for any round $t\in[T]$, let $N_i(t)$ denote the number of plays of action $i$ up to round $t$ (inclusive), $\wb{\mu}_i(t)$ denote the average \emph{observed} reward of action $i$ up to round $t$ (for notational convenience, we define $\wb{\mu}_i(0)=-\infty$), and
\begin{align}\label{eq:confbounds}
    \texttt{UCB}_{i}(t)\ldef\wb{\mu}_{i}(t)+\sqrt{\frac{6\log T}{N_i(t)}},~~~~\texttt{LCB}_{i}(t)\ldef\wb{\mu}_{i}(t)-\sqrt{\frac{6\log T}{N_i(t)}}
\end{align}
denote the \emph{upper confidence bound} and  \emph{lower confidence bound} of action $i$ up to round $t$, respectively.

\subsection{The LS-SE Algorithm}\label{sec:ub}

As a preliminary attempt, we provide a simple and intuitive algorithm for \UB, namely the \textit{Limited-Switch Successive Elimination} (\algoo) algorithm; see \cref{alg:lsse} for details.\footnote{Note that in \cref{line:3} and \cref{line:5} of \cref{alg:lsse}, $\frac{t_{l}-t_{l-1}}{|A_l|}$ might be fractional. For ease of presentation, we defer the rigorous treatment of such (minor) rounding issues to \cref{app:rounding}. The same principle applies to \pref{alg:hsse}.} The algorithm builds on the ``batched elimination'' framework recently developed by \cite{perchet2016batched} and \cite{gao2019batched} for the batched bandit problem, which splits the $T$ rounds into a given number of pre-determined batches and successively eliminates  ``poorly performing'' actions (based on confidence bounds)  in a batch-by-batch manner; the key ingredient of this framework is a delicate batch schedule (i.e., splitting rule) that strikes a balance between exploration and exploitation given a limited number of batches (cf. Section 4 of \citealt{perchet2016batched}). Since we are studying a different problem, directly applying a batched bandit algorithm to the \UB problem may not work --- in batched bandits, the number of batches is a given constraint, while in \UB, the switching budget is the given constraint. We thus add two ingredients into the \algoo algorithm: (i) an index $q(S,K)$ suggesting how many batches should be used to split the entire horizon, and (ii) a switching rule ensuring that the total number of switches across all $K$ actions cannot exceed the  budget $S$. %

\begin{algorithm}[htbp]\linespread{0.7}\selectfont{}
\caption{Limited-Switch Successive Elimination (\algoo)}
\label{alg:lsse}
{\bf Input:} Switching budget $S$, number of actions $K$, horizon $T$. \\
{\bf Initialization:} Compute $q(S,K)=\left\lfloor\frac{S-1}{K-1}\right\rfloor$. Divide the entire time horizon $T$ into $q(S,K)+1$ epochs: $(t_0:t_1],(t_1:t_2],\dots,(t_{q(S,K)}:t_{q(S,K)+1}]$, where the endpoints are defined by $t_0=0$ and
$$
t_j=\left\lfloor K^{1-\frac{2-2^{-(j-1)}}{2-2^{-q(S,K)}}}{T}^{\frac{2-2^{-(j-1)}}{2-2^{-q(S,K)}}}\right\rfloor,~~\forall j=1,\dots,q(S,K)+1.
$$%
Let $A_1=[K]$. Let $a_0$ be a random action in $[K]$.\\
{\bf Policy:}
\begin{algorithmic}[1]
\For{$l=1,\dots,q(S,K)$}
\If{$a_{t_{l-1}}\in A_{l}$}
\For{$i=a_{t_{l-1}}$ and {then} $i\in A_l\setminus\crl{a_{t_{l-1}}}$} {\color{blue}\Comment{starting from $i=a_{t_{l-1}}$ is critical}}
\State{Choose action $i$ for ${\frac{t_{l}-t_{l-1}}{|A_l|}}$ \emph{consecutive} rounds.}\label{line:3}
\EndFor
\Else %
\For{$i\in A_l$}
\State{Choose action $i$ for ${\frac{t_{l}-t_{l-1}}{|A_l|}}$ \emph{consecutive} rounds.}\label{line:5}
\EndFor
\EndIf
\State\multiline{Mark the last chosen action as $a_{t_l}$.}
\State{Elimination:}\label{line:elimination}
compute $\texttt{UCB}_{i}(t_l)$ and $\texttt{LCB}_{i}(t_l)$ for all $i\in A_l$ and let {\color{blue}\Comment learn from data
}
\[A_{l+1}=\crl*{i\in A_l \mid \texttt{UCB}_{i}(t_l)\ge \max_{j\in A_l} \texttt{LCB}_{j}(t_l)}.\]
\EndFor
\State{For $l=q(S,K)+1$, compute an action in $A_l$ that maximizes $\wb{\mu}_i(t_l)$. Keep choosing this action until round $T$.}
\end{algorithmic}
\end{algorithm}

\textbf{Intuition about LS-SE.} The algorithm divides the $T$ rounds into $q(S,K)+1=\left\lfloor \frac{S-1}{K-1}\right\rfloor+1$ epochs in advance, where an epoch corresponds to a batch in batched bandits. Note that there is no adaptivity \emph{within} each epoch: decisions are determined at the beginning of the epoch and do not depend on the rewards observed in this epoch. The epoch schedule follows the batch schedules given by \cite{perchet2016batched} and \cite{gao2019batched}, with slight differences in the dependence on $K$.\footnote{The batch schedule of \cite{perchet2016batched} does not involve $K$  because they only study the two-armed case. The batch schedule of \cite{gao2019batched} does not involve $K$ because they allow $\sup_{i,j\in[K]}\abs{\mu_i-\mu_j}\in[0,\sqrt{K}]$. Our epoch schedule is optimized for the (usual) setting of $\sup_{i,j\in[K]}\abs{\mu_i-\mu_j}\in[0,1]$ and leads to better regret in this setting.}  This schedule ensures that $t_1\asymp \frac{t_2}{\sqrt{t_1/K}}\asymp\cdots\asymp\frac{T}{\sqrt{t_{q(S,K)}/K}}$; combined with the celebrated \emph{successive elimination} strategy (see \cref{line:elimination}) in bandits, the schedule ensures that exploration and exploitation are balanced and the (worst-case) regret incurred during each epoch is at the same level. %
In addition, our two new ingredients (the index and the switching rule) guarantee the following properties:
\begin{itemize}
    \item Limited switches within each epoch: In epoch $l$, only $|A_l|-1\le K-1$ switches happen. %
    \item At most one switch between two consecutive epochs: If the last action chosen in epoch $l$ remains in $A_{l+1}$ ($l<q(S,K)$), then it will be the first action chosen in epoch $l+1$, and no switch occurs between these two epochs. If the last action chosen in epoch $l$ is eliminated from $A_{l+1}$, then epoch $l+1$ starts from another action in $A_{l+1}$, and one switch occurs between these two epochs.
    \item No switch within the last epoch:  In the last epoch, only the empirical best action {is} chosen.
    \item At most $S$ switches in $T$ rounds: by combining the above three properties with $q(S,K)=\floor*{\frac{S-1}{K-1}}$, one can show that the total number of switches is at most $q(S,K)(K-1)+1\le S$.
\end{itemize}

We show that \algoo is indeed an $S$-switch policy, and ensures the following upper bound on regret. 
The proof is standard\footnote{The regret analysis of \algoo is similar to the analysis of \cite{gao2019batched}; we present it for completeness. A difference is that we obtain slightly better dependence on $K$ under the condition $\sup_{i,j\in[K]}\abs{\mu_i-\mu_j}\in[0,1]$.} and deferred to \cref{app:proof-thm1}.
\begin{proposition}\label{thm:up-unit}
Let ${\pi}$ be the \algoo policy, then $\pi\in\Pi_S$. There exists an absolute constant $C\ge0$ such that for all $K\ge1$, $S\ge0$ and $T\ge K$,
\[
R^{{\pi}}(K,T)\le C{(\log K\log T)}K^{1-\frac{1}{2-2^{-q(S,K)}}} T^{\frac{1}{2-2^{-q(S,K)}}},
\]
where $q(S,K)=\left\lfloor\frac{S-1}{K-1}\right\rfloor$.
\end{proposition}

\textbf{Remark.} \cref{thm:up-unit} implies that for the classical \MAB problem, $\cO(K\log\log T)$ times of switches are sufficient for a learner to achieve the optimal $\wt{\cO}(\sqrt{KT})$ regret, which recovers a well-known result of \cite{cesa2013online} (see also \citealt{perchet2016batched,gao2019batched}).

\subsection{The AdaLS Algorithm}\label{sec:improved}
The \algoo algorithm, though being very simple, has several drawbacks that may degrade its performance.
Specifically:
\begin{itemize}
    \item The \algoo policy does not make full use of its switching budget. Consider the case of $S=2K-2$. Since $q(2K-2,K)=\floor*{\frac{2K-3}{K-1}}=1=q(K,K)$, the \algoo policy will just run as if it could only make $K$ switches, despite the fact that it can actually make $2K-2$ switches --- in this case, nearly half of the switching budget will never be used. Intuitively, an effective learning policy should make full use of its switching budget. It seems that by tracking and allocating the switching budget in a more careful way, one can achieve lower regret.
    \item The \algoo policy has (unnecessarily) low adaptivity. Note that the \algoo policy is a batched policy  that utilizes data  in a very restrictive way: it only learns from data at the end of each epoch, for at most $q(S,K)=\floor*{\frac{S-1}{K-1}}$ times. For example,  consider the case of $S=2K-2$. The \algoo policy will observe the data only once throughout the entire horizon. This is a  waste of a policy's information acquisition ability in \BwSC, where the learner is more flexible than in batched bandits, and can observe data at \emph{every} round. %
    Intuitively, data should be utilized to save switches and reduce regret, and one would expect that an effective policy will have a higher degree of \textit{adaptivity}, that is, it should learn from the available data and adapt to the environment more frequently than \algoo. %
\end{itemize}

To overcome the above drawbacks, we design a new algorithm, namely the \emph{Adaptive Limited-Switch} (\algon) algorithm; see \cref{alg:new} for details. \algon builds on and improves upon \algoo by (i) adopting a novel combinatorial and randomized exploration strategy and (ii) deciding when to make switches in a more data-driven fashion. These two features enable \algon to make better use of its switching budget and enjoy higher adaptivity (i.e., learn from data more frequently). We explain the key ideas of \algon  below.

\begin{algorithm}[t]\linespread{0.95}\selectfont{}
\caption{Adaptive Limited-Switch Policy (\algon)}
\label{alg:new}
{\bf Input:} Switching budget $S$, number of actions $K$, horizon $T$, tuning parameter $\lambda=1/2$. \\
{\bf Initialization:} Compute $q(S,K)=\left\lfloor\frac{S-1}{K-1}\right\rfloor$ and $r(S,K)=(S-1)\%(K-1)$. Define $\wh{r}(S,K)=\max\crl{r(S,K)+1-q(S,K),0}$. Define $t_0^{(1)}=t_0^{(2)}=0$ and 
\[
{t}_{j}^{(1)}=\floor*{\prn*{K-r(S,K)}^{1-\frac{2-2^{1-j}}{2-2^{-q(S,K)}}}{{T}}^{\frac{2-2^{1-j}}{2-2^{-q(S,K)}}}},~~\forall j=1,\dots,q(S,K)+1,
\]
\[
t_j^{(2)}=\floor*{K^{1-\frac{2-2^{1-j}}{2-2^{-q(S,K)-1}}}{{T}}^{\frac{2-2^{1-j}}{2-2^{-q(S,K)-1}}}},~~\forall j=1,\dots,q(S,K)+2.
\]
Let $A_1=[K]$. Let $A_1^{(2)}$ be a subset of $A_1$ obtained by uniformly sampling $\wh{r}(S,K)$ actions from $A_1$ \emph{without replacement} (thus $\abs{A_1^{(2)}}=\wh{r}(S,K)$). Let $A_1^{(1)}=A_1\setminus A_1^{(2)}$. Let $a_0$ be a random action in $A_1^{(1)}$.\\
{\bf Policy:}
\begin{algorithmic}[1]
\For{$l=1,\dots,q(S,K)$}
\State\multiline{Starting from an arbitrary action in $A_l^{(2)}$, choose each action in $A_l^{(2)}$ for $n_{l}^{(2)}=\floor{\frac{t_{l}^{(2)}-T_{l-1}^{(1)}}{\abs{A_l}}}$ \emph{consecutive} rounds. Mark the last round as $T_l^{(2)}$.}\label{line:nonadaptive}
\If{$a_{T_{l-1}^{(1)}}\in A_{l}^{(1)}$}\label{line:start}
\For{$i=a_{T_{l-1}^{(1)}}$ and then $i\in A_l^{(1)}\setminus\crl{a_{T_{l-1}^{(1)}}}$} {\color{blue}\Comment{starting from $i=a_{T_{l-1}^{(1)}}$ is critical}}
\State{Choose action $i$ for $n_l^{(2)}$ \emph{consecutive} rounds. Mark the last round as $T_{l,i}^{(1)}$.}\label{line:repeat_s}
\If{$\texttt{UCB}_i(T_{l,i}^{(1)})\ge \max_{j\in A_l}\texttt{LCB}_j(T_{l,i}^{(1)})$}{\color{blue}\Comment learn from data
}\label{line:learn1}
\State{Choose action $i$ for additional $\max\crl*{\floor{\frac{\lambda{t}_l^{(1)}-T_l^{(2)}}{\abs{A_l^{(1)}}}}-n_l^{(2)},0}$ \emph{consecutive} rounds.}\label{line:repeat_t}
\EndIf
\EndFor
\Else %
\For{$i\in A_l^{(1)}$}
\State{The same steps as \pref{line:repeat_s,line:learn1,line:repeat_t}.}{\color{blue}\Comment learn from data}\label{line:learn2}
\EndFor
\EndIf\label{line:end}
\State{Mark the last round as $T_l^{(1)}$, and mark the last chosen action as $a_{T_l^{(1)}}$.}
\State{Elimination:}\label{line:elimination2}
compute $\texttt{UCB}_{i}(T_l^{(1)})$ and $\texttt{LCB}_{i}(T_l^{(1)})$ for all $i\in A_l$, and let {\color{blue}\Comment learn from data
}
\[A_{l+1}^{(1)}=\crl*{i\in A_l^{(1)} \mid \texttt{UCB}_{i}(T_l^{(1)})\ge {\max_{j\in A_l} \max\crl{\texttt{LCB}_{j}(T_l^{(1)}), \texttt{LCB}_{j}(T_{l,i}^{(1)})}}},\]\[A_{l+1}^{(2)}=\crl*{i\in A_l^{(2)} \mid \texttt{UCB}_{i}(T_l^{(1)})\ge \max_{j\in A_l} \max\crl{\texttt{LCB}_{j}(T_l^{(1)}),\texttt{LCB}_{j}(T_{l,i}^{(1)})}},\]
\Statex~~~~~and $A_{l+1}=A_{l+1}^{(1)}\cup A_{l+1}^{(2)}$.
\EndFor
\For{$l=q(S,K)+1$}
\State\multiline{Starting from an arbitrary action in $A_l^{(2)}$, choose each action in $A_l^{(2)}$ for $n_{l}^{(2)}=\floor{\frac{t_{l}^{(2)}-T_{l-1}^{(1)}}{\abs{A_l}}}$ \emph{consecutive} rounds. Mark the last round as $T_l^{(2)}$.}\label{line:last}
\State Elimination: compute $\texttt{UCB}_{i}(T_l^{(2)})$ and $\texttt{LCB}_{i}(T_l^{(2)})$ for all $i\in A_l$, and let {\color{blue}\Comment learn from data
}\label{line:last-elim}
\[A_{l+1}=\crl*{i\in A_l \mid \texttt{UCB}_{i}(T_l^{(2)})\ge \max_{j\in A_l} \texttt{LCB}_{j}(T_l^{(2)})}.\]
\State Compute an action in $A_l$ that maximizes $\wb{\mu}_i(T_l^{(2)})$. Keep choosing this action until round $T$.\label{line:commit}
\EndFor
\end{algorithmic}
\end{algorithm}

\textbf{Two key indices.} At initialization, \algon performs the Euclidian division of $(S-1)$ by $(K-1)$ and obtains two key indices: the quotient $q(S,K)$ and the remainder $r(S,K)\in\crl{0,\cdots,K-2}$. While the quotient $q(S,K)$ is already used by \algoo to determine the number of epochs (which ensures that \algoo makes at most $q(S,K)(K-1)+1$ switches), the remainder $r(S,K)=S-1-q(S,K)(K-1)$ is a new index that reflects the ``abandoned switching budget'' of \algoo, i.e., the amount of switching budget that will never be used by \algoo. Intuitively, the larger $r(S,K)$ is, the more \algon can (hopefully) improve upon \algoo by making better use of the switching budget.

\textbf{Random partition of the action set.} An important goal of \algon is to make better use of the switching budget when $r(S,K)$ is large. This is however a non-trivial task:  since $r(S,K)\le K-2$, the additional switching budget does not allow for visiting all actions in $[K]$; that is, we may only  conduct additional exploration for a \emph{subset} of actions in $[K]$, and the selection of this subset requires careful consideration. We address this issue by utilizing the idea of \emph{randomization}: at initialization, we randomly split $[K]$ into two subsets $A_1^{(1)}$ and $A_1^{(2)}$, with $\abs{A_1^{(1)}}=K-\wh{r}(S,K)$ and $\abs{A_1^{(2)}}=\wh{r}(S,K)$ (we will explain the configuration of $\wh{r}(S,K)$ shortly). Then, in the execution of the policy, we treat the actions in $A_1^{(1)}$ and $A_1^{(2)}$ differently, allowing  the actions in $A_1^{(2)}$ to be explored more frequently than the actions in $A_1^{(1)}$. Specifically, before the last switch (where we commit to the empirical best action; see \cref{line:commit}), we allow \algon to switch to each action in $A_1^{(1)}$ for at most $q(S,K)$ times, while allowing it to switch to each action in $A_1^{(2)}$ for at most $q(S,K)+1$ times (as a comparison, \algoo switches to every action in $[K]$ for at most $q(S,K)$ times before the last switch). By letting $\wh{r}(S,K)=\max\crl{r(S,K)+1-q(S,K),0}$, we enable \algon to make good use of the switching budget while never going over it: if $\wh{r}(S,K)>0$  (i.e., $r(S,K)$ is large enough), then \algon will  make up to $q(S,K)(K-\wh{r}(S,K))+(q(S,K)+1)\wh{r}(S,K)-1+1=S$ switches; otherwise, \algon will behave similar to \algoo and  make up to $q(S,K)(K-1)+1<S$ switches.  See \pref{app:illustration} for an illustration of how \algon makes switches. 
We remark that it is crucial to determine $A_1^{(1)}$ and $A_1^{(2)}$  \emph{randomly} rather than deterministically, as randomization enables  better worst-case performance. %

\textbf{Combinatorial exploration scheme.} At initialization, we define two series of time points $\prn{t_j^{(1)}}_{j=1}^{q(S,K)+1}$ and $\prn{t_j^{(2)}}_{j=1}^{q(S,K)+2}$, which provide (rough) guidance on how we should balance the exploration and exploitation for actions in $A_1^{(1)}$ and $A_1^{(2)}$. These two series are similar but different from the series $\prn{t_j}_{j=1}^{q(S,K)+1}$ defined in \cref{alg:lsse} due to (i) we allow \algon to switch to the actions in $A_1^{(2)}$ more frequently and (ii) we need to consider the interplay between the two classes of actions. Then, \algon runs in $q(S,K)+1$ epochs. In each epoch $l\in[q(S,K)]$, \algon first explores each  action in $A_l^{(2)}$ (which consists of all \emph{un-eliminated} action in $A_1^{(2)}$) for an equal number of rounds (see \cref{line:nonadaptive}), then explores each action in  $A_l^{(1)}$ (which consists of all \emph{un-eliminated} action in $A_1^{(1)}$) for a \emph{data-dependent} number of rounds (see \cref{line:start} to \cref{line:end}), and finally conducts elimination to determine $A_{l+1}^{(1)}$ and $A_{l+1}^{(2)}$ (see \cref{line:elimination2}). At the last epoch $q(S,K)+1$, \algon will first explore all actions in $A_{q(S,K)+1}^{(2)}$ (see \cref{line:last}), then conducts elimination, and finally commits to the empirical best action (see \cref{line:commit}). Notably, in every elimination step of \algon, the confidence bounds of different actions are at different scales, because they were explored non-uniformly. Compared with \algoo where each un-eliminated action is uniformly explored in epoch $l\in[q(S,K)]$, the exploration scheme of \algon requires more delicate design (which is combinatorial in nature) because we need to ensure that the elimination based on \emph{confidence bounds with different scales} are effective. 

\textbf{Higher adaptivity via more frequent queries to the data.} A significant difference between \algon and \algoo is that \algon utilizes data more frequently and is not a batched policy --- while \algon runs in $q(S,K)+1$ epochs, each epoch does not correspond to a batch because \emph{actions selected during epoch $l$ depends on the latest data collected during epoch $l$} (see \cref{line:learn1,line:learn2,line:last-elim}, where \algon utilizes the latest data to determine whether to switch or not). Moreover, the actual epoch schedule $\prn{T_l^{(1)}}_{l=1}^{q(S,K)+1}$ is also data-dependent, i.e., \algon decides when to start and end each epoch only \emph{after} gradual access to the data. Such additional adaptivity is critical for \algon to achieve better performance; otherwise, one cannot have careful control over the exploration and may over-explore the actions in $A_1^{(1)}$ (as they are visited relatively less frequently).

We provide a rigorous analysis of \algon, verify that it is an $S$-switch policy, and show that it attains an improved regret bound; see the statement in \cref{thm:ub-n}. The proof of \cref{thm:ub-n} (which closely follows the intuition that we provide above) is considerably more challenging than \cref{thm:up-unit}; see \cref{app:upper} for details.

\begin{theorem}\label{thm:ub-n}
Let ${\pi}$ be the \algon policy, then $\pi\in\Pi_S$. There exists an absolute constant $C\ge0$ such that for all $K\ge1$, $S\ge0$ and $T\ge K$,
\[
R^\pi(K,T)\le C(\log T)^2\cdot \max\crl*{\frac{\prn*{K-{r}(S,K)}^{2-\frac{1}{2-2^{-q(S,K)}}}}{K}T^{\frac{1}{2-2^{-q(S,K)}}},\ {{K}^{1-\frac{1}{2-2^{-q(S,K)-1}}}}T^{\frac{1}{2-2^{-q(S,K)-1}}}},
\]
where $q(S,K)=\left\lfloor\frac{S-1}{K-1}\right\rfloor$ and $r(S,K)=(S-1)\%(K-1)$.
\end{theorem}

To illustrate the regret guarantee given by \pref{thm:ub-n}, we use it to calculate the exact regret rates (in terms of both $K$ and $T$) of \algon under different concrete values of $S$; see \cref{tb:version1} for details. As benchmarks, we also calculate the exact regret rates of \algoo using \cref{thm:up-unit},  and compare the regret of \algoo and \algon under each single value of $S$; see the detailed comparisons in \cref{tb:version1}.

We make two observations. First, \algon shares the same regret rate as \algoo when $K-r(S,K)=\Omega(K)$; see the second column  in \pref{tb:version1}, where the regret rate of both \algon and \algoo is ${K^{1-\frac{1}{2-2^{-q(S,K)}}}T^{\frac{1}{2-2^{-q(S,K)}}}}$ (within logarithmic factors). This implies that \algon does not attain a fundamentally better rate than \algoo when $K=\wt\cO(1)$ or when $r(S,K)$ is not close to $K$. %
Second, \algon can attain a significantly better regret rate when $r(S,K)=K-o(K)$; see the last two columns in \pref{tb:version1}, where the regret rate of \algon is always better than \algoo. Note that the closer $r(S,K)$ is to $K$, the better \algon's regret rate can be. In particular, when $r(S,K)=K-2$ (i.e, when $S$ is a \emph{multiple} of $K-1$), \algon attains a rate of ${K^{-1}T^{\frac{1}{2-2^{-q(S,K)}}}\vee K^{1-\frac{1}{2-2^{-q(S,K)-1}}}T^{\frac{1}{2-2^{-q(S,K)-1}}}}$ --- if the first term dominates, then \algon improves upon the regret of \algoo by a multiplicative factor of $\wt\Theta(K^{2-\frac{1}{2-2^{-q(S,K)}}})$; if the second term dominates (which is not uncommon when $K$ is large), then the regret of \algon has a better growth rate in $T$, which enables it to perform arbitrarily better than \algoo when $T\rightarrow\infty$.

\begin{table}[htbp] \caption{Regret of \algoo and \algon under different switching budgets.\protect\footnotemark \\Here $\eps\in(0,1)$ is an arbitrary constant independent of $K$ and $T$ (it can be arbitrarily close to $0$, as long as it is fixed).}
		\label{tb:version1}%
		\begin{center}
			{
				\begin{tabular}{l||c|c|c}
					\toprule 	\hline 
					
					\multicolumn{4}{l}{  $S\in\crl{0,1,\dots,K-1}$}                                                                                \\ \hline %
					$S$                                 & $0,1,\dots,(1-\eps)(K-1)$   &  $K-1-\wt\Theta\prn*{K^{\delta}},~\delta\in(0,1)$ & $K-1$ \\ \hline
					\algoo & $\wt\Theta\prn*{T}$ & $\wt\Theta\prn*{T}$ & $\wt\Theta\prn*{T}$ \\ \hline
					\algon & $\wt{\Theta}\prn*{T}$    & $\wt{\Theta}\prn*{K^{\delta-1}T \vee K^{\frac{1}{3}}T^{\frac{2}{3}}}$   & $\wt{\Theta}\prn*{K^{-1}T \vee K^{\frac{1}{3}}T^{\frac{2}{3}}}$        \\ \hline
					\hline
					\multicolumn{4}{l}{  $S\in\crl{K,\dots,2K-2}$}                                                                                \\ \hline %
					$S$                                 & $K,K+1,\dots,(1-\eps)(2K-2)$   &  $2K-2-\wt\Theta(K^{\delta}),~\delta\in(0,1)$ & $2K-2$ \\ \hline 
					\algoo & $\wt{\Theta}\prn*{K^{\frac{1}{3}}T^{\frac{2}{3}}}$ & $\wt{\Theta}\prn*{K^{\frac{1}{3}}T^{\frac{2}{3}}}$ & $\wt{\Theta}\prn*{K^{\frac{1}{3}}T^{\frac{2}{3}}}$ \\ \hline
					\algon & $\wt{\Theta}\prn*{K^{\frac{1}{3}}T^{\frac{2}{3}}}$    & $\wt{\Theta}\prn*{K^{\frac{4}{3}\delta-1}T^{\frac{2}{3}} \vee K^{\frac{3}{7}}T^{\frac{4}{7}}}$   & $\wt{\Theta}\prn*{K^{-1}T^{\frac{2}{3}} \vee K^{\frac{3}{7}}T^{\frac{4}{7}}}$         \\ \hline
					\hline 
 					\multicolumn{4}{l}{ $S\in\crl{2K-1,\dots,3K-3}$}                                                                                \\ \hline %
 					$S$       & {$2K-1,2K,\dots,(1-\eps)(3K-3)$ } &  $3K-3-\wt\Theta(K^{\delta}),~\delta\in(0,1)$   &  $3K-3$  \\  \hline
 					\algoo & $ \wt\Theta\prn*{K^{\frac{3}{7}}T^{\frac{4}{7}}}$ & $ \wt\Theta\prn*{K^{\frac{3}{7}}T^{\frac{4}{7}}}$ & $ \wt\Theta\prn*{K^{\frac{3}{7}}T^{\frac{4}{7}}}$ \\ \hline
 					\algon  &{$ \wt\Theta\prn*{K^{\frac{3}{7}}T^{\frac{4}{7}}}$}  & $\wt{\Theta}\prn*{K^{\frac{10}{7}\delta-1}T^{\frac{4}{7}} \vee K^{\frac{7}{15}}T^{\frac{8}{15}}}$  &  $\wt{\Theta}\prn*{K^{-1}T^{\frac{4}{7}} \vee K^{\frac{7}{15}}T^{\frac{8}{15}}}$    \\ \hline
					\hline 
 					\multicolumn{4}{l}{ $S\in\crl{3K-2,\dots,4K-4}$}                                                                                \\ \hline %
 					$S$       & {$3K-2,3K-1,\dots,(1-\eps)(4K-4)$ } &  $4K-4-\wt\Theta(K^{\delta}),~\delta\in(0,1)$   &  $4K-4$  \\  \hline
 					\algoo & $ \wt\Theta\prn*{K^{\frac{7}{15}}T^{\frac{8}{15}}}$ & $ \wt\Theta\prn*{K^{\frac{7}{15}}T^{\frac{8}{15}}}$ & $ \wt\Theta\prn*{K^{\frac{7}{15}}T^{\frac{8}{15}}}$ \\ \hline
 					\algon  &{$ \wt\Theta\prn*{K^{\frac{7}{15}}T^{\frac{8}{15}}}$}  & $\wt{\Theta}\prn*{K^{\frac{22}{15}\delta-1}T^{\frac{8}{15}} \vee K^{\frac{15}{31}}T^{\frac{16}{31}}}$  &  $\wt{\Theta}\prn*{K^{-1}T^{\frac{8}{15}} \vee K^{\frac{15}{31}}T^{\frac{16}{31}}}$
					\\ \hline \bottomrule 
			\end{tabular}}
		\end{center}
\end{table}
\footnotetext{We make two remarks for \pref{tb:version1}. First, for brevity, we only present the regret rates for $S\in[0,4K-4]$ in \cref{tb:version1} --- regret rates for larger $S$ follows the same pattern. Second, since it is quite easy to show \emph{algorithm-specific} lower bounds for \algoo and \algon which match the upper bounds in \cref{thm:up-unit} and \cref{thm:ub-n} respectively (up to logarithmic factors), we directly use $\wt\Theta$ (rather than $\wt\cO$) to describe the regret of \algoo and \algon; of course, the \emph{algorithm-specific} lower bounds for \algoo and \algoo do not imply fundamental limits for \emph{other}  algorithms (which can be arbitrarily more complicated) --- proving an universal \emph{algorithm-independent} lower bound is the task of \pref{sec:lb}. 
}

\subsection{Lower Bound on Regret}\label{sec:lb}

The \algon algorithm, though being a significantly refined version of the \algoo algorithm, still {seems to} leave plenty of room for improvement. For example, while \algon has higher adaptivity than \algoo, it learns from data for at most $Kq(S,K)+1$ times, leaving an open question of whether one can utilize even more adaptivity to achieve lower regret.  
Moreover, as discussed in \pref{sec:improved},  \algon only improves the regret with the help of $K$, failing to \emph{directly} improve the regret's dependence on the most important parameter $T$ when $K=\wt\cO(1)$. This motivates the following natural  questions: 
Is it possible to directly improve the regret in terms of $T$? Can the dependence on $K$ be further improved? What is the fundamental limit of the  \textsf{U-BwSC} problem?

We answer the above questions by establishing a strong (and quite surprising) information-theoretic lower bound on the regret incurred by \emph{any} admissible policy; see \pref{thm:lb-unit}. The lower bound directly match the upper bound in \cref{thm:ub-n}, indicating that \algon is optimal up to logarithmic factors. Notably, our lower bound holds for any $K$, any $S$, any $T$, thus is substantially stronger than a specific lower bound demonstrated for special choices of $S,K,T$.

\begin{theorem}\label{thm:lb-unit}
There exists an absolute constant $C>0$ such that for all $K> 1,S\ge0, T\ge 2K$ and for all policy $\pi\in\Pi_S$,
\begin{equation*}
    R^\pi(K,T)\ge\frac{C}{\log T}\cdot \max\crl*{\frac{\prn*{K-{r}(S,K)}^{2-\frac{1}{2-2^{-q(S,K)}}}}{K}T^{\frac{1}{2-2^{-q(S,K)}}},\ {{K}^{1-\frac{1}{2-2^{-q(S,K)-1}}}}T^{\frac{1}{2-2^{-q(S,K)-1}}}},
\end{equation*}
where $q(S,K)=\left\lfloor\frac{S-1}{K-1}\right\rfloor$ and $r(S,K)=(S-1)\%(K-1)$.
\end{theorem}

As we introduced in \pref{sec:results}, the proof of \pref{thm:lb-unit} is  non-trivial, and will be elaborated on in a separate section (\cref{sec:methodology}). Combining \cref{thm:ub-n} and \cref{thm:lb-unit}, we completely characterize the optimal regret of \UB as follows.

\begin{corollary}For all $S\ge0, K>1, T\ge 2K$, we have
\[R^*_S(K,T)=\wt{\Theta}(1)\cdot \max\crl*{\frac{\prn*{K-{r}(S,K)}^{2-\frac{1}{2-2^{-q(S,K)}}}}{K}T^{\frac{1}{2-2^{-q(S,K)}}},\ {{K}^{1-\frac{1}{2-2^{-q(S,K)-1}}}}T^{\frac{1}{2-2^{-q(S,K)-1}}}}.\]
If $K=\wt{\cO}(1)$, then we have $R^*_S(T)=\wt{\Theta}\prn*{T^{\frac{1}{2-2^{-q(S,K)}}}}$.
\end{corollary}

\subsubsection{On the Necessity of Switching in MAB}\label{sec:nec}
The lower bound in \pref{thm:lb-unit} also leads to new results for the classical \MAB problem.

\begin{corollary}\label{cor:sc} The following properties hold for the classical \MAB: (i) $\Theta(K\log\log T)$ switches are necessary and sufficient for achieving $\wt{\cO}(\sqrt{KT})$ regret, (ii) for any fixed $N\in\bbN_{>0}$, $N(K-1)+1$ switches are necessary and sufficient for achieving $\wt{\cO}(K^{1-\frac{1}{2-2^{-N}}}T^\frac{1}{2-2^{-N}})$ regret, and (iii) $\Omega(K)$ switches are necessary for achieving sublinear regret.
\end{corollary}

Note that the number of switches stated in \pref{cor:sc} refers to the maximum number of switches that a policy can make. While \cite{cesa2013online} has proposed policies that achieve  $\wt{\cO}(\sqrt{KT})$ (near-optimal) regret with $\cO(K\log\log T)$ switches, 
no prior work has answered the question of how many switches are \textit{necessary} for a near-optimal learning policy in \MAB. To the best of our knowledge, this paper is the first  to show $\Omega(K\log\log T)$  lower bound on the number of switches.

\subsection{Phase Transitions}\label{sec:sbrto}

\pref{cor:sc} provides a non-asymptotic (i.e., finite-time) characterization on the optimal regret of the \UB problem. While such non-asymptotic characterization is very general (e.g., it holds for arbitrary $S,K,T$ and does not rely on any assumption on their orders), if we want to obtain deeper insights  on the optimal regret's \emph{growth rate} (which is easier to understand when defined as concrete limiting behavior), then the asymptotic regime may be more appropriate in terms of making our statements rigorous and precise.  
In this subsection, we use asymptotics to rigorously define the {optimal regret rate}, and characterize the trade-off between the optimal regret rate and the switching budget. As we shall see, the trade-off reveals surprising \emph{phase transitions}\footnote{The terminology of phase transitions originates from physics (see, e.g., \citealt{domb2000phase}), and has been used in various fields in probability theory and statistics. We note that most of rigorous definitions of phase transitions in probability theory and statistics require asymptotics; see, e.g., \citealt{wainwright2009sharp,bayati2015universality}.} (to be defined shortly). For ease of presentation, all the ``rates'' defined in this subsection do not involve logarithmic factors.

\subsubsection{Phase Transitions in the Fixed-$K$ Asymptotic Regime}\label{sec:phase1} We first consider the most natural asymptotic regime where we let the time horizon $T\rightarrow\infty$ and keep the number of arms $K$ fixed; we refer to this regime as the ``fixed-$K$'' asymptotic regime. For any fixed switching budget $S\ge0$, we are interested in the growth rate of the optimal regret $R^*_S(T)$ as $T\rightarrow\infty$. Following the convention of statistics and machine learning (see, e.g., \citealt{tsybakov2008introduction}), we define the {optimal regret rate} (i.e., minimax rate) as the power function that best approximates $R_S^*(T)$ as $T\rightarrow\infty$; see below.
\begin{definition}\label{def:asy1}For any fixed $K>1,S\ge0$, there exists a unique constant $p\in[0,1]$ such that
\[
\lim_{T\rightarrow\infty}\frac{R_S^*(T)}{T^{p+\eps}}=0,~~\lim_{T\rightarrow\infty}\frac{R_S^*(T)}{T^{p-\eps}}=\infty,~~~~\forall \eps>0.
\]
We call $T^p$ the optimal regret rate under switching budget $S$, and $p$ the optimal regret rate exponent. 
\end{definition}
Note that an equivalent definition is to directly let $p\ldef \lim_{T\rightarrow\infty}\frac{\log R^*_S(T)}{\log T}$ and let the power function $T^p$ be the optimal regret rate (see \citealt{hu2020smooth}).

By \pref{cor:sc}, $R^*_S(T)=\wt{\Theta}\prn*{T^{\frac{1}{2-2^{-\floor*{(S-1)/(K-1)}}}}}$ when $K$ is fixed. To the best of knowledge, this is the first time that a floor function naturally arises in the exponent of $T$ in the optimal regret of an online learning problem. Consequently, we know that the optimal regret rate under switching budget $S$ is ${T^\frac{1}{2-2^{-\floor*{(S-1)/(K-1)}}}}$, which exhibits   surprising \textit{phase transitions} described below.
\begin{definition}[Phases \& Transition Points]
In the fixed-$K$ regime, we call the interval $[(j-1)(K-1)+1,j(K-1)+1)$ the $j$-th phase, and call $j(K-1)+1$ the $j$-th transition point ($j\in\bbN_{>0}$).
\end{definition}

\begin{observation}[Phase Transitions]
 As $S$ increases from $0$ to infinity, $S$ will leave the $j$-th phase and enter the $(j+1)$-th phase at the $j$-th transition point ($j\in\bbN_{>0}$). Each time $S$ arrives at a transition point, the optimal regret rate will change abruptly, and then remain the same until $S$ arrives the next transition point.%
\end{observation}

\begin{table}[htbp]
   \caption{Optimal Regret Rate under Different Switching Budgets for a Fixed $K$} 
   \label{tab:example}
   \small %
   \centering %
   \begin{tabular}{lcccccc} %
   \toprule
   $S$ & $[0,K)$ & $[K,2K-1)$ & $[2K-1,3K-2)$ &$[3K-2,4K-3)$ & $[4K-3,5K-4)$ & $[5K-4,6K-5)$\\ 
   \midrule
   Rate & $T$ & $T^{2/3}$ & $T^{4/7}$ & $T^{8/15}$ & $T^{16/31}$& $T^{32/63}$\\
   \bottomrule
   \end{tabular}
\end{table}

Phase transitions are illustrated in \pref{tab:example}. This phenomenon seems counter-intuitive, as it suggests that in the fixed-$K$ regime, increasing switching budget would not help reduce the best achievable regret rate, as long as the budget does not reach the next transition point. Moreover,  the abrupt change happens at each transition point is very interesting --- at this point, a minimal difference in the switching budget can fundamentally change the statistical nature of the problem.
 Along with phase transitions, we also observe an interesting property: the length of each phase is always equal to $K-1$. This property is elegant and reveals some favorable features of the \UB problem: as we will show in \pref{sec:general}, this property does not hold under the general switching cost structure.

\textbf{Remark.} While phase transitions are intriguing and theoretically interesting, we would like to make some comments on the scope of the above reuslts. First, one should keep in mind that for any $S$, the above analysis concerns the growth rate (i.e. scaling behavior) of $R^*_S(T)$ as $T$ grows (which is a statistical property), rather than the numerical value of $R^*_S(T)$ for a  specific $T$. 
In particular, we are less interested in describing how  $R_S^*(T)$ changes with respect to $S$ for a fixed, specifically chosen $T$; instead, we seek to understand how the minimax rate of the problem (which reflects the ``learnability'' of the problem when the sample size grows) changes with respect to $S$. Second, phase transitions are more relevant to practice when $S$ belongs to the first 4 or 5 phases. Indeed, the regret rate exponent in \pref{tab:example} decreases dramatically as $S$ goes over more phases --- when $S$ is relatively large, the difference between two phases may be too small to make the rate reduction happens in the ``ideal'' asymptotic world really make a difference in reality. Recall that in the non-asymptotic regime where $S$ can depend on $T$, $\cO(\log\log T)$ switches are sufficient for one to achieve  $\wt{\cO}(\sqrt{T})$ regret --- this also indicates that the most interesting transitions should happen when $S$ is small.

\subsubsection{Phase Transition in the Growing-$K$ Asymptotic Regime} We now consider a second asymptotic regime which allows $K$ to grow with $T$ in a moderate rate, which corresponds to the ``growing-dimension'' asymptotic regime in statistics (\citealt{portnoy1984asymptotic,portnoy1988asymptotic}). Specifically, we consider the following ``growing-$K$'' asymptotic regime: $K,T\rightarrow\infty$ and $K/T^\alpha\rightarrow c$ for some $\alpha\in(0,1), c\in(0,\infty)$. By \pref{cor:sc}, $\Omega(K)$ switches are necessary for achieving sublinear regret in \MAB; thus, in the ``growing-$K$'' regime of \UB, a fixed $S$ cannot avoid $\Omega(T)$ regret, and the values of $S$ that we are most interested in should range from $\Omega(K)$ to $o(K\log\log T)$. This indicates that in the growing-$K$ regime, $S$ should be naturally understood as ``a function of $K$'': the dependence between $S$ and $K$ is necessary, while $S/K$ should have no or extremely small dependence on $T$. We thus only consider $S$ such that $S/K\rightarrow \theta$ for some constant $\theta\in[0,\infty)$, and we call $\theta\ldef\lim_{K\rightarrow\infty}S/K$  the ``budget-to-arm ratio'' ({BAR}). The optimal regret rate in the ``growing-$K$'' regime can be then defined similar to \pref{def:asy1} (with $K$ and $S$ scales proportional to $T$), or equivalently by calculating $p\ldef\lim_{T\rightarrow\infty}\frac{\log R_S^*(K,T)}{\log T}$ and denote $T^p$ as the optimal regret rate. %

The first finding in the growing-$K$ regime is that the (original form of) phase transitions described in \pref{sec:phase1} may not hold any more, and the existence of ``abrupt rate changes'' depends on the magnitude of $K$ relative to $T$. To see this, let us focus on a small range of $S$ at the end of the first phase and at the start of the second phase: $S=K-1-\wt\Theta(K^{\delta})$ with $\delta\in(0,1)$, $S=K-1$ and $S=K$. By \pref{tb:version1} and simple calculation, the corresponding optimal regret rate exponents for them are $\max\crl*{1-\alpha(1-\delta),\frac{\alpha+2}{3}}$, $\max\crl*{1-\alpha,\frac{\alpha+2}{3}}$ and $\frac{\alpha+2}{3}$ respectively. By letting $\delta$ move smoothly from 1 to 0, one can find that the optimal regret rate exponent associated with $S=K-1-\Theta(K^{\delta})$ smoothly decays from $1$ to $\max\crl*{1-\alpha,\frac{\alpha+2}{3}}$: while $S$ is always in the first phase defined in \pref{sec:phase1}, the optimal regret rate does not ``remain the same'' any more. Moreover, whether there is an abrupt change when $S$ moves from $K-1$ to $K$ depends on the magnitude of $\alpha$. If $\alpha<\frac{1}{4}$, then there is still an abrupt change in the optimal regret rate (e.g., if $\alpha=0.1$, then the rate jumps from $T^{0.9}$ to $T^{0.7}$); if $\alpha\ge\frac{1}{4}$, then the optimal regret rate under $S=K-1$ remains unchanged when $S$ reaches the next transition point. One can keep conducting such analysis for other phases, and find similar examples on the ending range of each phase where $r(S,K)=K-o(K)$ --- interestingly, these are also the ranges that \algon significantly improves the regret; see the last two columns in \pref{tb:version1}.

The second finding in the growing-$K$ regime is that when we consider how the optimal regret rate changes with respect to the budget-to-arm ratio  $\theta$ (rather than $S$), we can still discover phase transitions similar to \pref{sec:phase1}. Indeed, the counter-examples described in the above paragraphs correspond to the ranges of $S$ where the remainder function $r(S,K)$ moves from $K-o(K)$ to $K-2$ to $0$, i.e., the ranges where $S$ moves from $N(K-1)-o(K)$ to $N(K-1)$ to $N(K-1)+1$ for some integer $N\in\bbN_{>0}$. All valid $S$ in these ranges have the property that the BAR $\theta=\lim_{K\rightarrow}S/K=N\in\bbN_{>0}$. This indicates that the complicated behavior of the optimal regret rate described in the above paragraph might only happen in scenarios where the BAR $\theta$ is exactly a positive integer. In fact, for all $S$ such that $\theta$ exists and $\theta$ is not an integer (e.g., consider $S=\floor{2.5K}$, then $\theta=2.5$ is not an integer), one can find that the (unique) optimal regret rate exponent is $\alpha+\frac{1}{2-2^{-\floor{\theta}}}(1-\alpha)$, which contains a floor function.
Consequently, we can define the interval $(j-1,j)$ as the $j$-th phase $(j\in\bbN_{>0})$ for $\theta$, and discover phase transitions of the optimal regret rate as illustrated in \pref{tab:example1}. A difference in this new notion of phase transitions is that in the growing-$K$ regime, each ``transition point''  $j\in\bbN_{>0}$ can be associated with infinitely many optimal regret rates which interpolate between the rates of the previous and the next phase (see the previous paragraph for the example of $\theta=1$). 
\begin{table}[htbp]
   \caption{Optimal Regret Rate under Different BAR $\theta$ when $K$ Grows as $T^{\alpha}$} 
   \label{tab:example1}
   \small %
   \centering %
   \begin{tabular}{lccccccc} %
   \toprule
      $\theta$ & $[0,1)$ & $1$ & $(1,2)$ &$2$ & $(2,3)$ & $3$ & (3,4)\\ 
   \midrule
   Rate & $T$ & $-$ & $T^{(2+\alpha)/3}$ & $-$ & $T^{(4+3\alpha)/7}$& $-$& $T^{(8+7\alpha)/15}$\\
   \bottomrule
   \end{tabular}
\end{table}

\subsection{Relationship between Limited {Switches} and Limited Adaptivity}\label{sec:lsla}

In this subsection, we discuss the relationship between limited switches and limited adaptivity in bandit problems. As discussed {in Section \ref{sec:batch}}, {in the \UB problem, the constraint is on the number of switches and is defined in the ``action world,'' hence the learner has full adpativity. By contrast, in the batched bandit problem, the constraint is on adaptivity and is defined in the ``observation world,'' hence the learner has full switching power}. Since the two constraints in the two problems are defined in two different ``worlds,'' the relationship between the two problems is interesting.

We first claim that the \UB problem can be seen as a strict relaxation of the 
batched bandit problem (no matter with the ``static grid'' restriction, like \citealt{perchet2016batched}, or without such restriction, like \citealt{gao2019batched}), 
in the sense that \UB admits more flexible policies and can enjoy a fundamentally better optimal regret rate. The \algoo and \algon algorithms help establish this claim. First, one can easily show that any $M$-batch policy that achieves certain regret in the $M$-batch $K$-armed bandit problem can be transformed, using the \algoo ingredients and randomization, to an $S$-switch policy that achieves exactly the same regret in the $S$-switch $K$-armed \UB problem, as long as $q(S,K)=M-1$, i.e., $S\in[(M-1)(K-1)+1:M(K-1)]$. This implies that the admissible policy class of the $S$-switch \UB problem essentially contains  the admissible policy class of the $(q(S,K)+1)$-batch bandit problem as a subset. Moreover, since an $S$-switch policy can utilize data much more flexibly than an $(q(S,K)+1)$-batch policy (see \pref{sec:improved}),  $(q(S,K)+1)$-batch algorithms \emph{necessarily} suffer from sub-optimal rates for \UB in general. Note that the regret lower bound for the $(q(S,K)+1)$-batch bandit problem is $\wt{\Omega}\prn*{K^{1-\frac{1}{2-2^{-q(S,K)}}}T^{\frac{1}{2-2^{-q(S,K)}}}}$\footnote{This is shown in \cite{gao2019batched} with slight difference dependence on $K$ as they allow $\sup_{i,j}\abs{\mu_i-\mu_j}=\sqrt{K}$. Since our setting is $\sup_{i,j}\abs{\mu_i-\mu_j}\in[0,1]$, we write the corresponding ``right'' dependence on $K$ for ease of comparison.}, which is a fundamental limit for all $(q(S,K)+1)$-batch algorithms; \algon's regret bound \pref{eq:algon-reg} surpasses this limit, indicating that the admissible policy class of \UB indeed contains lower-regret policies, and \UB can have a fundamentally better optimal regret rate.

On the other hand, the performance improvement that one can benefit from the above relaxation also has a limit, as demonstrated by our lower bound (\pref{thm:lb-unit}). In fact, one can find that the optimal regret rate of the $S$-switch \UB (\pref{cor:sc}) \emph{interpolates} between the optimal regret rates of the $(q(S,K)+1)$-batch and the $(q(S,K)+2)$-batch bandit problems; moreover, when $K=\wt\cO(1)$, the optimal regret rate of the $S$-switch \UB coincides with the optimal regret rate of the $(q(S,K)+1)$-batch bandit problem. The above findings provide very useful managerial insights:
{\begin{enumerate}[itemindent=\dimexpr\labelwidth+\labelsep\relax,leftmargin=0pt]
    \item The switching constraint is a more relaxed constraint than the batch constraint, and enable better performance guarantees when $K$ is large. \item Limiting switches (in the ``action'' world) implicitly limits adaptivity (in the ``observation'' world), in the sense that the optimal $S$-switch policy's regret rate lies between the optimal  $(q(S,K)+1)$-batch and $(q(S,K)+2)$-batch policies' regret rates; when $K=\wt\cO(1)$, the regret rate of the optimal $S$-switch policy and the optimal $(q(S,K)+1)$-batch policy coincide (up to logarithmic factors). 
\end{enumerate}}

Finally, we would like to point out that our lower bound result (\pref{thm:lb-unit}) is theoretically stronger and more general than the lower bounds for batched bandits. Indeed, since any $M$-batch policy can be transformed to an equivalent $((M-1)(K-1)+1)$-switch policy, any lower bound proved for \UB implies a lower bound for batched bandits. In particular, by directly plugging $S=(M-1)(K-1)+1$ in \pref{thm:lb-unit} (this value of $S$ actually corresponds to the ``easier-to-prove'' part of \pref{thm:lb-unit}, as it is the start of a ``phase''), one obtains the minimax lower bound result of  \cite{gao2019batched} as a corollary (the order of $K,T$ will be the same under their conditions).

\section{General Switching Costs}\label{sec:general}
We now proceed to the general setting of \BwSC, where $c_{i,j}$ can be any non-negative real number and even $\infty$ ($i\ne j$)  . %
For this general setting, a new and fundamental resarch question is how the structure of switching costs $(c_{i,j})$ affects the statistical nature of \BwSC. Since this question is interesting and challenging even when  $K=\wt\cO(1)$, in this section, we only seek to derive algorithms and regret bounds that are effective when $K=\wt\cO(1)$. We believe that the techniques developed in \pref{sec:unit} should be helpful for one to obtain refined algorithms and results (for general switching costs) when $K$ is large, but we leave it for future work. In what follows, we consider two switching cost structures: a general symmetric one in \pref{sec:sym} and an asymmetric one in \pref{sec:asym}.

\subsection{Symmetric Switching Costs}\label{sec:sym}
We first consider the \emph{general symmetric switching cost} structure where $c_{i,j}=c_{j,i}$ for all $i,j\in[K]$. The corresponding \BwSC problem is referred to as the \GB problem. 
To start with, we need to enhance the framework of Section \ref{sec:formulation} to better represent the switching costs. We do this by representing switching costs via a weighted graph. Let $G=(V,E)$ be a (weighted) complete graph, where $V=[K]$ (i.e., each vertex corresponds to an action), and the edge between $i$ and $j$ is assigned a weight $c_{i,j}$ ($\forall i\ne j$). We call the weighted graph $G$ the \textit{switching graph}. %
In this section, we assume the switching costs satisfy the triangle inequality: $\forall i,j,l\in[k]$, $c_{i,j}\le c_{i,l}+c_{l,j}$. We relax this assumption in Appendix \ref{app:relax}.

The results in \pref{sec:unit}  suggest that a simple and  effective learning strategy (when $K=\wt\cO(1)$) is to 
repeatedly visit all actions for many times and then commit to the best action, in a manner similar to \algoo. This indicates that in the \GB problem, one should consider how to repeatedly visit all vertices in the switching graph, in a most economical way to stay within budget. %
This implies a connection between \GB and the celebrated shortest Hamiltonian path problem. 
Motivated by this connection, we propose the \textit{Hamiltonian-Switching Successive Elimination} (\algog) algorithm, and present it in  \pref{alg:hsse}. The algorithm enhances the original \algoo algorithm by adding an additional ingredient: a pre-specified switching order determined by the shortest Hamiltonian path of the switching graph $G$. Note that while the shortest Hamiltonian path problem is NP-hard, solving this problem is entirely an ``offline'' step in the \algog algorithm, i.e., for a given switching graph, the learner only needs to solve this problem once. %
We also comment that one may use the techniques presented in \pref{sec:improved} to design a refined algorithm (analogous to \algon) that achieves better performance; however, we leave this for future work, as the simple \algog algorithm is already sufficient for revealing important properties of \GB when $K=\wt\cO(1)$.

\begin{algorithm}[t]\linespread{0.7}\selectfont{}
\caption{Hamiltonian-Switching Successive Elimination (\algog)}
\label{alg:hsse}
{\bf Input:}  Switching budget $S$, switching graph $G$, horizon $T$.\\
{\bf Initialization:} Let $A_1=[K]$. Find a shortest Hamiltonian path in $G$: ${i_1}\rightarrow\dots\rightarrow {i_{K}}$. Denote the total weight of the shortest Hamiltonian path as $H$. Compute $q'(S,G)=\left\lfloor\frac{S-\max_{i,j\in[K]}c_{i,j}}{H}\right\rfloor$. Divide the entire time horizon $T$ into $q'(S,G)+1$ epochs: $(t_0:t_1],(t_1:t_2],\dots,(t_{q'(S,G)}:t_{q'(S,G)+1}]$, where the endpoints are defined by $t_0=0$ and
\[
t_j=\left\lfloor K^{1-\frac{2-2^{-(j-1)}}{2-2^{-q'(S,G)}}}{T}^{\frac{2-2^{-(j-1)}}{2-2^{-q'(S,G)}}}\right\rfloor,~~\forall j=1,\dots,q'(S,G)+1.
\]
{\bf Policy:}
\begin{algorithmic}[1]
\For{$l=1,\dots,q'(S,G)$}
\If{$l$ is odd}
\For{$i=i_1,\dots,i_K$}{\color{blue}\Comment{along the direction of ${i_1}\rightarrow\dots\rightarrow {i_{K}}$}}\label{line:direction}
\State{If $i\in A_l$ (i.e., uneliminated), choose action $i$ for ${\frac{t_{l}-t_{l-1}}{|A_l|}}$ \emph{consecutive} rounds.}
\EndFor
\Else
\For{$i=i_K,\dots,i_1$}{\color{blue}\Comment{along the direction of ${i_K}\rightarrow\dots\rightarrow {i_{1}}$} (the reverse of the above)}
\State{If $i\in A_l$ (i.e., uneliminated), choose action $i$ for ${\frac{t_{l}-t_{l-1}}{|A_l|}}$ \emph{consecutive} rounds.}
\EndFor
\EndIf
\State{Elimination:}
compute $\texttt{UCB}_{i}(t_l)$ and $\texttt{LCB}_{i}(t_l)$ for all $i\in A_l$ and let {\color{blue}\Comment learn from data
}
\[A_{l+1}=\crl*{i\in A_l \mid \texttt{UCB}_{i}(t_l)\ge \max_{j\in A_l} \texttt{LCB}_{j}(t_l)}.\]
\EndFor
\State{For $l=q'(S,K)+1$, compute an action in $A_l$ that maximizes $\wb{\mu}_i(t_l)$. Keep choosing this action until round $T$.}
\end{algorithmic}
\end{algorithm}

Let $H$ denote the total weight of the shortest Hamiltonian path of $G$. It is not difficult to verify that \algog is an $S$-switching-budget policy and ensures the following upper bound on regret; see \pref{app:proof-thm3} for a proof.

\begin{theorem}\label{thm:gub}
Let $\pi$ be the \algog policy, then $\pi\in\Pi_S$. There exists an absolute constant $C\ge0$ such that for all $G$, $K=|G|$, $S\ge 0$, $T\ge K$,
\[
R^\pi(T)\le C{(\log K\log T)}K^{1-\frac{1}{2-2^{-q'(S,G)}}}  T^{\frac{1}{2-2^{-q'(S,G)}}},
\]
where $q'(S,G)=\left\lfloor\frac{S-\max_{i,j\in[K]}{c_{i,j}}}{H}\right\rfloor$.
\end{theorem}

In \pref{thm:glb}, we provide a lower bound that is very close to the above upper bound. The proof of \pref{thm:glb} builds on the proof of \pref{thm:lb-unit}, but has two notable differences: (i) it involves several new techniques to deal with the general switching cost structure, and (ii) it pays less attention to the dependence on $K$; see \pref{app:lb-gbwcs} for details.

\begin{theorem}\label{thm:glb}
There exists an absolute constant $C>0$ such that for all $G, K=|G|>1, S\ge 0, T\ge 2K$ and for all policy $\pi\in\Pi_S$,
\[
R^\pi(K,T)\ge\frac{C}{K\log T}\cdot T^{\frac{1}{2-2^{-q''(S,G)}}},
\]
where $q''(S,G)=\left\lfloor\frac{S-\max_{i\in[K]}\min_{j\ne i}c_{i,j}}{H}\right\rfloor$.
\end{theorem}

Let us focus on the case of  $K=\wt\cO(1)$ and compare the upper and lower bounds given by \pref{thm:gub} and \pref{thm:glb}. When the switching costs satisfy the condition $\max_{i,j\in[K]}{c_{i,j}}=\max_{i\in[K]}\min_{j\ne i}c_{i,j}$, we have $q'(S,G)=q''(S,G)$, thus the two bounds directly match (up to $\text{polylog}(T)$). This reveals an interesting fact: when $\max_{i,j\in[K]}{c_{i,j}}=\max_{i\in[K]}\min_{j\ne i}c_{i,j}$, the optimal regret rate of \GB is completely characterized by the floor function $\left\lfloor\frac{S-\max_{i,j\in[K]}{c_{i,j}}}{H}\right\rfloor$, which further depends on $H$. The fact implies that the length of the shortest Hamiltonian path is indeed a fundamental quantity associated with the \GB problem, and conveys an important message: the structure of switching costs may affect the optimal regret rate of \BwSC through some key quantities associated with  graph traversal problems. We now provide a concrete switching cost structure satisfying the condition $\max_{i,j\in[K]}{c_{i,j}}=\max_{i\in[K]}\min_{j\ne i}c_{i,j}$ below. 
\begin{example}[Isolated Action Model]
Consider a set of $K-1$ ``close'' or ``similar'' actions $\crl{1,\cdots,K-1}$, and another ``isolated'' action $K$, such that $c_{K,1}=\cdots=c_{K,K-1}\ge \max_{i,j\in[K-1]}c_{i,j}$ (i.e., action $K$ is isolated from other actions such that its distance to every other action is a large constant). This model always satisfies the condition $\max_{i,j\in[K]}{c_{i,j}}=\max_{i\in[K]}\min_{j\ne i}c_{i,j}$, and subsumes the unit-switching-cost model as a special case. As an example, in promotion planning, $1,\cdots,K-1$ can be different variants of a standard promotion strategy, while $K$ can be an aggressive clearance strategy. 
\end{example}

When the condition $\max_{i,j\in[K]}{c_{i,j}}=\max_{i\in[K]}\min_{j\ne i}c_{i,j}$ is not satisfied, for any switching graph $G$, the above two bounds still match for a wide range of $S$:
\[
\left[0,H+\max_{i\in[k]}\min_{j\ne i}c_{i,j}\right)\bigcup \left\{\bigcup_{n=1}^{\infty}\left[nH+\max_{i,j\in[k]}c_{i,j},(n+1)H+\max_{i\in[k]}\min_{j\ne i}c_{i,j}\right)\right\}.
\]
Even when $S$ is not in this range, we still have $q'(S.G)\le q''(S,G)\le q'(S.G)+1$ for any $G$ and any $S$, which means that the difference between the two indices is at most 1 and the regret bounds are always very close. In fact, it can be shown that as $S$ increases, the gap between the upper and lower bounds decreases \textit{doubly exponentially}. Therefore, the \algog algorithm is quite effective for the  \GB problem when $K=\wt\cO(1)$.

\subsection{Asymmetric Switching Costs: The Departure Cost Structure}\label{sec:asym}

We now consider another switching cost structure that allows asymmetry. Since the general asymmetric case is only more complicated than the case studied in \pref{sec:sym}, we consider a special case of asymmetric switching costs: there exists $\bm{c}=(c_1,\dots,c_K)\in\bbR_{\ge0}^K$ such that $c_{i,j}=c_i$ for all $i\in[K],j\ne i$. That is, the  switching cost between any pair of actions only depends on  the action that the learner departs from. We refer to this switching cost structure as the \emph{departure cost} structure (with $c_i$ called the departure cost of action $i$), and the corresponding \BwSC problem as the \DB problem. As we shall see, for this fairly general problem, we can fully characterize the optimal regret when $K=\wt\cO(1)$.

We provide an algorithm (\algoa) for the \DB problem; see \pref{alg:asse}. The algorithm follows the same main steps of \algog, but has some important differences in the initialization step: it calculates the key actions $i_1, i_K$ and the key index $q(S,\bm{c})$ differently. We provide some intuition for this new configuration. First, we can still construct a switching graph $G$ associated with the switching costs, but this time being a \emph{directed} graph. Since $i_K$ is the action with the maximum departure cost (denoted by $c^{(1)}$), the path $i_1\rightarrow \cdots\rightarrow i_K$ is a shortest Hamiltonian path of the switching graph $G$. The choice of $i_K$ is thus consistent with \algog. However, since the switching costs are asymmetric now, one cannot guarantee that the reverse path $i_K\rightarrow\cdots\rightarrow i_1$ also has a small length. In \pref{alg:asse}, based on the departure cost structure, we optimize the reverse path by letting $i_1$ be the action with the second largest departure cost (denoted by $c^{(2)}$). The determination of the key index $q(S,\bm{c})$ is a little more complicated, as we need to consider the alternation of two directions; thanks to the departure cost structure, it is easy to compute.

\begin{algorithm}[htbp]\linespread{0.7}\selectfont{}
\caption{Asymmetric-Switching Successive Elimination (\algoa)}
\label{alg:asse}
{\bf Input:}  Switching budget $S$, switching costs $\bm{c}$, horizon $T$.\\
{\bf Initialization:} Let $A_1=[K]$. Find an action $i_K\in\arg\max_{i\in[K]}c_i$ and an action $i_1\in\arg\max_{i\in[K]\setminus\crl{i_1}}c_i$. Let $(i_2,\dots,i_{K-1})$ be an arbitrary permutation of $[K]\setminus\crl{i_1,i_K}$. 
Let $\Sigma=\sum_{i=1}^K c_i$, $c^{(1)}=c_{i_K}$ and $c^{(2)}=c_{i_1}$. Compute $q(S,\bm{c})=\max\crl*{1+2\left\lfloor\frac{S-\Sigma}{2\Sigma-c^{(1)-c^{
(2)}}}\right\rfloor,2\floor*{\frac{S-c^{(2)}}{2\Sigma-c^{(1)}-c^{(2)}}}}$. Divide the entire time horizon $T$ into $q(S,\bm{c})+1$ epochs: $(t_0:t_1],(t_1:t_2],\dots,(t_{q(S,\bm{c})}:t_{q(S,\bm{c})+1}]$, where  $t_0=0$ and
\[
t_j=\left\lfloor K^{1-\frac{2-2^{-(j-1)}}{2-2^{-q(S,\bm{c})}}}{T}^{\frac{2-2^{-(j-1)}}{2-2^{-q'(S,\bm{c})}}}\right\rfloor,~~\forall j=1,\dots,q(S,\bm{c})+1.
\]
{\bf Policy:} The same as Lines 1 to 9  of \pref{alg:hsse}.
\end{algorithm}

\begin{theorem}\label{thm:asym}
When $K=\wt\cO(1)$, the optimal regret of \DB is $\wt\Theta\prn*{T^{\frac{1}{2-2^{-q(S,\bm{c})}}}}$, where $q(S,\bm{c})$ is given by \pref{alg:asse}. Furthermore, the \algoa algorithm attains this regret rate.
\end{theorem}

\pref{thm:asym} shows that when $K=\wt\cO(1)$, the optimal regret rate of \DB can be completely characterized by the key index $q(S,\bm{c})$, and \algoa is rate-optimal. This reveals surprising phase transitions similar to \pref{sec:phase1}; the difference is that each phase may not have an equal length anymore. See \pref{tab:example2} for an illustration (the quantities $\Sigma,c^{(1)},c^{(2)}$ are given in \pref{alg:asse}).

\begin{table}[htbp]
   \caption{Optimal Regret Rate (of \DB) under Different Switching Budgets for Fixed $K$ and $\bm{c}$} 
   \label{tab:example2}
   \small %
   \centering %
   \begin{tabular}{lcccc} %
   \toprule
   $S$ & $[0,\Sigma)$ & $[\Sigma,2\Sigma-c^{(1)})$ & $[2\Sigma-c^{(1)},3\Sigma-c^{(1)}-c^{(2)})$ &$[3\Sigma-c^{(1)}-c^{(2)},4\Sigma-2c^{(1)}-c^{(2)})$\\ 
   \midrule
   Rate & $T$ & $T^{2/3}$ & $T^{4/7}$ & $T^{8/15}$ \\
   \bottomrule
   \end{tabular}
\end{table}

\section{Reverse Fano-type Inequalities and Lower Bound Analysis}\label{sec:methodology}

This section provides an overview of the methodological contributions associated with our proof of \pref{thm:lb-unit}. We first introduce the \GRF inequality, then present our lower bound approach.

\subsection{Reverse Fano-type Inequalities}

\emph{Fano’s} inequality  is a fundamental information-theoretical tool for developing algorithm-independent impossibility results in statistics and machine learning. In one of its most classical forms, it states that for any sequence of $N\ge2$ probability measures $\bbP_1,\dots\bbP_N$ on the same measurable space $(\Omega,\cF)$, and any sequence of events $E_1,\cdots,E_N$ forming a partition of $\Omega$, it holds that
\begin{equation}\label{eq:fano}
    \frac{1}{N}\sum_{i=1}^N \Prob_i(E_i)\le \frac{\frac{1}{N}\sum_{i=1}^N D_{\rm KL}(\bbP_i\parallel\bbQ)+\log 2}{\log N},
\end{equation}
where $\bbQ$ is an arbitrary measure on $(\Omega,\cF)$, and $D_{\rm KL}(\cdot\parallel\cdot)$ stands for the \emph{KL divergence}. Fano's inequality has important consequences for various problems in various fields; see \cite{scarlett2019introductory} for a survey. For example, in multiple hypothesis testing, by considering events of the form $E_i = \crl{\psi = i}$ where $\psi:\Omega\rightarrow[N]$ is a \emph{test}, \pref{eq:fano} provides a sharp lower bound on the  average error probability  $\frac{1}{N}\sum_{i=1}^N\bbP_i(\psi\ne i)$ for any test $\psi$. 

Many variants of Fano's inequality have been derived in the literature; see \cite{scarlett2019introductory} and \cite{gerchinovitz2020fano} for overviews. However, to our knowledge, existing literature does not provide a reverse version of \pref{eq:fano}, i.e., an inequality that establishes a sharp lower bound on $\frac{1}{N}\sum_{i=1}^N \Prob_i(E_i)$ \textbf{for any} $E_1,\dots,E_N$ forming a partition, which corresponds to a sharp upper bound on $\frac{1}{N}\sum_{i=1}^N\bbP_i(\psi\ne i)$ {for any} test $\psi$ in multiple hypothesis testing. While there are indeed some existing inequalities sometimes referred to as ``reverse Fano's inequalities'' in the literature (e.g., \citealt{chu1966inequalities,tebbe1968uncertainty}), and some other related inequalities are implied by the  recent work of \cite{gerchinovitz2020fano}, these inequalities either fail to lower bound $\frac{1}{N}\sum_{i=1}^N \Prob_i(E_i)$ \textbf{for any} $E_1,\dots,E_N$ forming a partition, or suffer from sub-optimal dependence on $N$; see \pref{app:fanoexp} for detailed discussions. We fill  this gap by developing a reverse version of \pref{eq:fano} and significantly generalizing it to a much stronger version, i.e., the \GRF inequality; see \pref{prop:grf-m}. The proof builds on the general framework developed by \cite{gerchinovitz2020fano}, with some new techniques to obtain better dependence on $N$ via localized versions of Pinsker's inequality; see \pref{app:tools}.

\begin{proposition}[Generalized Reverse Fano-type Inequality]\label{prop:grf-m}
Let $D(\cdot\parallel\cdot)$ be the KL divergence or the reverse KL divergence (see \pref{app:tools} for definitions). Let $(\Omega_1,\cF_1), \dots, (\Omega_N,\cF_N)$ be an arbitrary sequence of measurable spaces. For any $i\in[N]$, let 
$\Prob_i$ and $\bbQ_i$ be arbitrary probability measures on  $(\Omega_i,\cF_i)$, and $E_i\in\cF_i$ be an arbitrary event. %
We have
\begin{equation}
    \frac{1}{N}\sum_{i=1}^N \Prob_i(E_i)\ge \frac{1}{N}\sum_{i=1}^N \bbQ_i(E_i)-\sqrt{2\cdot{\frac{1}{N}\sum_{i=1}^N \bbQ_i(E_i)}\cdot\frac{1}{N}\sum_{i=1}^N D(\Prob_i\parallel \bbQ_i)}.
\end{equation}
\end{proposition}

\pref{prop:grf-m} is fairly general and enjoys several advantages: (i) it acts in the reverse direction of \pref{eq:fano}, thus enables new applications, (ii) the events $E_1,\dots,E_N$ do \textbf{not} need to form a partition, (iii) the probability measures $\bbP_1,\dots\bbP_N$ can be defined on different measurable spaces, (iv) $D(\cdot\parallel\cdot)$ can be the \emph{reverse} KL divergence, and (v) the probability measures $\bbQ_1,\cdots,\bbQ_N$ can vary with events and do not need to be fixed. All of the above advantages will be utilized in our lower bound analysis.

\subsection{The Five-Step Approach to Establish \pref{thm:lb-unit}}\label{sec:sketch}
Given any $K>1$, $S\ge0$ and $T\ge 2K$, we focus on the setting of $\cD_k=\mathcal{N}(\mu_k,1)$, $\forall k\in[K]$, where  $\mathcal{N}(\mu_k,1)$ denotes the Gaussian distribution with mean $\mu_k$ and variance 1. Since in this setting the underlying environment (i.e., reward distributions) $\cD$ is completely specified by a vector $\bmu=(\mu_1,\cdots,\mu_K)\in\mathbb{R}^{K}$, for simplicity, we directly use the vector $\bmu$ to represent the environment. %

\textbf{Notations.} For any environment $\bmu$, let $X_{\bmu}^{t}(k)\sim\mathcal{N}(\mu_k,1)$ denote the i.i.d. random reward of each action $k$ at round $t$ ($k\in[K],t\in[T]$). For any policy $\pi\in\Pi_S$, for any environment $\bmu$, for any $t\in[T]$, we use $a_t$ and  $X^t_{\bmu}(a_t)$ to denote the random action selected by  and the random reward observed by policy $\pi$ at round $t$ under environment $\bmu$, respectively. %
Let $\bbP_{\bmu}^{\pi}$ be the probability measure induced by random variables ${\prn*{a_1,X_{\bmu}^1(a_1)},\dots,\prn*{a_T,X_{\bmu}^T(a_T)}}$, and $\bbE_{\bmu}^\pi$ be the associated expectation operator. Let $R_{\bmu}^{\pi}(T)\ldef T\mu^*-\bbE_{\bmu}^{\pi}\brk*{\sum_{t=1}^T{\mu}_{a_t}}$ be policy $\pi$'s distribution-dependent regret under environment $\bmu$.

\textbf{Outline.} Since the desired lower bound \pref{eq:ubwsc-lb} becomes the standard $\wt\Omega(\sqrt{KT})$   lower bound 
when $S=\Omega(K\log\log T)$ (see \pref{app:lower}), we focus on the more interesting case of $S=\cO(K\log\log T)$. For any such $S,K,T$,  we seek to explicitly construct a family of environments $\Phi$, such that for any $S$-switch policy $\pi\in\Pi_S$, the ``average-case regret'' $\frac{1}{|\Phi|}\sum_{\bmu\in\Phi}R_{\bmu}^{\pi}(T)$ is lower bounded by \pref{eq:ubwsc-lb} --- this implies that the worst-case regret $R^{\pi}(T)$ is also lower bounded by \pref{eq:ubwsc-lb}.  
In our proof, we construct two classes of environments to show the two parts in the ``max'' of \pref{eq:ubwsc-lb} respectively. In this section, we focus on the more challenging part --- the $\wt{\Omega}\prn*{\frac{(K-r(S,K))^{2-\frac{1}{2-2^{-q(S,K)}}}}{K}T^{\frac{1}{2-2^{-q(S,K)}}}}$ lower bound.

Our lower bound proof program consists of five steps:
\begin{enumerate}
    \item \textbf{R}isky \textbf{E}vents
    \item \textbf{C}ombinatorial arguments and lower bounds under a single environment
    \item \textbf{A}lternative environments, bad events, and lower bound reductions \hfill{\color{blue}(we construct $\Phi$ here)}
    \item \textbf{P}robability space changing tricks
    \item Applying the \GRF inequality %
\end{enumerate}
Based on the initials of the first four steps, we call the program \lbargu. We provide an overview of each step below. The detailed proof can be found in \cref{app:lower}.

\textbf{Step 1: risky events.} We first define a {stopping time} $\tau$, which is the first round that the learner's number of switches reaches $S$.  
We then define a class of \emph{risky events} as follows: for any $k\in[K]$, let

$E_{1,k}^{(1)}\ldef\crl*{\text{action }k \text{ is not chosen in period }\brk*{1:t_{1}^{(1)}}},$

$E_{j,k}^{(1)}\ldef\crl*{\text{action }k \text{ is not chosen in period }\brk*{t_{j-1}^{(1)}:t_{j}^{(1)}}},~~\forall j\in[2:q(S,K)],$

$E_{q(S,K)+1,k}^{(1)}\ldef\crl*{\text{action }k \text{ is not chosen in period }\brk*{t_{q(S,K)}^{(1)}:\floor{\prn{t_{q(S,K)}^{(1)}+T}/2}}},$

$E_{q(S,K)+2,k}^{(1)}\ldef\crl*{\tau\le \floor{\prn{t_{q(S,K)}^{(1)}+T}/2},\, a_\tau=k, \text{ action }k\text{ is not chosen in period }\brk*{t_{q(S,K)}^{(1)}:\tau-1}}.$\\
By doing so, we get $\prn*{q(S,K)+2}K$ risky events (of the form $E_{j,k}^{(1)}$) in total. Note that the time points $\prn{t_j^{(1)}}_{j=1}^{q(S,K)+1}$ are fixed and given in \pref{alg:new}, and the events $\prn{E_{q(S,K)+2,k}^{(1)}}_{k=1}^K$ are defined based on the stopping time $\tau$. Such delicate design based on $\tau$ is novel and crucial; see \pref{app:REdef}.

The risky events characterize some important patterns that are ``unavoidable'' (in a certain sense, as we shall see in Step 2) for any $S$-switch policy under any environment. They are considered ``risky'' because, while they may not directly lead to large regret under an arbitrary environment, each of them would lead to larger regret under a specifically chosen environment (i.e.., the \emph{alternative} environment in Step 3) which will be included in our environment class $\Phi$.

\textbf{Step 2: combinatorial arguments and lower bounds under a single environment.}
In this step, we prove a key result (\cref{lm:first-class-m}) using  non-trivial combinatorial and probabilistic arguments. The arguments extensively exploit the properties of the switching constraint.
\begin{lemma}\label{lm:first-class-m}For any $S$-switch policy $\pi\in\Pi_S$, for any environment $\bmu$, we have
\[
\sum_{j\in[q(S,K)+2]}\sum_{k\in[K]}\bbP_{\bmu}^{\pi}\prn*{E_{j,k}^{(1)}}\ge K-\ceil*{\frac{S}{q(S,K)+1}}=\wt{\Omega}\prn*{\frac{K-r(S,K)}{K}}.
\]
\end{lemma}
\cref{lm:first-class-m} implies the following fact: under any \emph{single} environment $\bmu$, the average probability of the  risky events is $\wt{\Omega}\prn*{\frac{K-r(S,K)}{K}}$. That is, the occurrence of a  risky event is ``probabilistically unavoidable'' for any $S$-switch policy under any single environment. This result precisely characterize the potential weakness of an $S$-switch policy and reveals fundamental properties of the switching constraint.

\textbf{Step 3: alternative environments, bad events, and lower bound reductions.} In this step, we construct our environment class $\Phi\ldef\crl*{\bmu_{j,k}^{(1)}\mid j\in[q(S,K)+2],k\in[K]}$, which consists of $(q(S,K)+2)K$ judiciously chosen \emph{alternative} environments. Each alternative environment $\bmu_{j,k}^{(1)}$  is designed to make the risky event $E_{j,k}^{(1)}$ become  a \emph{bad event} whose occurrence implies large regret. We can then reduce the task of proving a lower bound on the ``average-case regret''  $\frac{1}{|\Phi|}\sum_{\bmu\in\Phi}R_{\bmu}^{\pi}(T)$ to the task of proving a lower bound on the ``average-case bad event probability''  $\frac{1}{(q(S,K)+2)K}\sum_{j\in[q(S,K)+2]}\sum_{k\in[K]}\bbP_{j,k}^{(1)}\prn*{E_{j,k}^{(1)}}$, where $\bbP_{j,k}^{(1)}\ldef\bbP^{\pi}_{\bmu_{j,k}^{(1)}}$ denotes the \emph{alternative} measure associated with policy $\pi$ and alternative environment $\bmu_{j,k}^{(1)}$.

Specifically, our construction of alternative environments is as follows. 
Let $\bm{0}=(0,\dots,0)\in\bbR^K$ be the \emph{reference} environment.  For any $j\in[q(S,K+2)]$, define a reward gap
\[
\Delta_{j}^{(1)}\ldef\begin{cases}
1, &\text{if }j=1,\\
\frac{1}{{2(q(S,K)+2)}}\sqrt{\frac{{K-r(s,k)}}{{t_{j-1}^{(1)}}}}, &\text{if }j\in[2:q(S,K)+1],\\
-\frac{1}{{2(q(S,K)+2)}}\sqrt{\frac{{K-r(s,k)}}{{t_{q(S,K)}^{(1)}}}}, &\text{if }j=q(S,K)+2.
\end{cases}
\]
For any $j\in[q(S,K)+2], k\in[K]$, define an \emph{alternative} environment $\bmu^{(1)}_{j,k}\ldef \prn*{\mu^{(1)}_{j,k;1},\dots,\mu^{(1)}_{j,k;K}}\in\bbR^K$ where
\[
\mu^{(1)}_{j,k;i}\ldef\begin{cases}\Delta_j^{(1)}, &\text{if }i=k,\\
0, &\text{otherwise.}
\end{cases}
\]
Note that each alternative environment only differs from the reference environment in one coordinate. %

In \pref{lm:bad}, we show that  the risky event $E_{j,k}^{(1)}$ is indeed a bad event under environment $\bmu_{j,k}^{(1)}$, in the sense that its occurrence implies that the regret is larger than a universal quantity $\cR_{\rm bad}(S,K,T)=\wt\Omega\prn*{(K-r(S,K))^{1-\frac{1}{2-2^{-q(S,K)}}}T^{\frac{1}{2-2^{-q(S,K)}}}}$. Thus, in order to prove the desired lower bound on $\frac{1}{|\Phi|}\sum_{\bmu\in\Phi}R_{\bmu}^{\pi}(T)$, it suffices to prove the following statement:
\begin{equation}\label{eq:obj1-m}
    \wb{p^{(1)}}\ldef\frac{1}{(q(S,K)+2)K}\sum_{j\in[q(S,K)+2]}\sum_{k\in[K]}\bbP_{j,k}^{(1)}\prn*{E_{j,k}^{(1)}}=\wt{\Omega}\prn*{\frac{K-r(S,K)}{K}},
\end{equation}

\textbf{Step 4: probability space changing tricks.} 
Let $\bbQ\ldef\bbP_{\bm{0}}^\pi$ denote the \emph{reference} measure. By applying \cref{lm:first-class-m} to the reference environment $\bm{0}$, we have
\[
\wb{q^{(1)}}\ldef\frac{1}{(q(S,K)+2)K}\sum_{j\in[q(S,K)+2]}\sum_{k\in[K]}\bbQ\prn*{E_{j,k}^{(1)}}=\wt{\Omega}\prn*{\frac{K-r(S,K)}{K}},
\]
Therefore, in order to show \cref{eq:obj1-m}, it suffices to show that $\wb{p^{(1)}}$ is close to $\wb{q^{(1)}}$. 
Note that $\wb{p^{(1)}}$ is the average of the sequence $\crl*{\bbP_{j,k}^{(1)}\prn*{E_{j,k}^{(1)}}}$, where a sequence of events $\crl*{E_{j,k}^{(1)}}$ are evaluated by a sequence of varying alternative measures $\crl*{\bbP_{j,k}^{(1)}}$, while $\wb{q^{(1)}}$ is the average of the sequence $\crl*{\bbQ\prn*{E_{j,k}^{(1)}}}$, where the same sequence of events $\crl*{E_{j,k}^{(1)}}$  are evaluated by a single and fixed reference measure $\bbQ$. Intuitively, we just need a ``change of  measure'' / information-theoretic argument --- if the alternative measures $\crl*{\bbP_{j,k}^{(1)}}$ are ``close enough'' to the reference measure $\bbQ$, then $\wb{p^{(1)}}$ is close to $\wb{q^{(1)}}$.

Unfortunately, it turns out that the divergence between $\crl*{\bbP_{j,k}^{(1)}}$ and  $\bbQ$ is too large to make the above argument work. An important reason is that such an argument directly deals with the underlying measures $\crl*{\bbP_{j,k}^{(1)}}$ and  $\bbQ$, thus 
completely overlooks the special structures of the risky event sequence $\crl*{E_{j,k}^{(1)}}$.  Therefore, we need to integrate the structural properties of risky events into our argument. We develop \emph{probability space changing tricks} to address this challenge. Specifically, 
we design two sequences of \emph{artificial} measures $\crl*{\bbP_{j,k}'}$ and  $\crl*{\bbQ_{j,k}'}$ based on the structural properties of $\crl*{E_{j,k}^{(1)}}$, such that (i) each  ${\bbP_{j,k}'}$ (resp. ${\bbQ_{j,k}'}$) is the \emph{restriction} of ${\bbP_{j,k}^{(1)}}$ (resp. $\bbQ$) to a carefully-chosen $\sigma$-algebra $\cF'_{j,k}$ which \emph{tightly} contains $E_{j,k}^{(1)}$, and (ii) the reverse KL divergence between ${\bbP_{j,k}'}$ and ${\bbQ_{j,k}'}$ is small enough. 
We can then represent $\wb{p^{(1)}}$ and $\wb{q^{(1)}}$ as the averages of $\crl*{\bbP_{j,k}'(E_{j,k}^{(1)})}$ and $\crl*{\bbQ_{j,k}'(E_{j,k}^{(1)})}$, and bound the difference between $\wb{p^{(1)}}$ and $\wb{q^{(1)}}$ by 
showing that $\crl*{\bbP_{j,k}'}$ and  $\crl*{\bbQ_{j,k}'}$ are ``close enough.''

\textbf{Step 5: applying the GRF inequality.} In the last step, we apply the \GRF inequality to provide a tight lower bound on $\wb{p^{(1)}}$ in terms of $\wb{q^{(1)}}$, thus completes the proof of \pref{eq:obj1-m} (and eventually \pref{thm:lb-unit}). We remark that we thoroughly utilizes the five advantages of the \GRF inequality in this step: (i) we need to lower bound $\wb{p^{(1)}}$ rather than lower bound $1-\wb{p^{(1)}}$ (the latter is what classical Fano-type inequalities do), (ii) our events $\crl*{E_{j,k}^{(1)}}$ (or their complements) do not form a partition, (iii) our measures $\crl*{\bbP_{j,k}'}$ are defined on different measurable spaces, (iv) we need to use the \emph{reverse} (rather than the standard) KL divergence to evaluate the ``closeness'' between $\crl*{\bbP_{j,k}'}$ and  $\crl*{\bbQ_{j,k}'}$, as we need to fix the reference environment to characterize the policy's behavior, and (v) the artificial reference measures $\crl*{\bbQ_{j,k}'}$  are not fixed.

\section{Concluding Remarks}\label{sec:conclusion}

We study the stochastic multi-armed bandit problem with a constraint on the total cost incurred by switching between actions. Under different switching cost structures, we prove matching (or almost matching) upper and lower bounds on regret and provide near-optimal algorithms for the problem. We conclude by discussing some natural extensions, which we defer to the appendix.%

First, all of the algorithms in this paper require the  knowledge of $T$. A natural question is what will happen if $T$ is not known a priori. In \pref{app:unknown}, we show that any algorithm with a switching budget independent of $T$  cannot attain a sublinear regret rate for \UB if $T$ is unknown. The result indicates that (i) the knowledge of $T$ is somehow necessary if we do not want the switching budget to grow with $T$, and (ii) $\Omega(\frac{K\log T}{\log\log T})$ switches are necessary for achieving $\wt\cO({\sqrt{KT}})$ regret in the classical \MAB problem without the knowledge of $T$.  

Second, our upper and lower bounds in \pref{sec:asym} do now always match. A natural question is whether one of them is tight. In \pref{app:tight}, we show that neither of the bounds are tight in $T$ in general (even for a fixed $K$). Our theoretical characterization in \pref{app:tight} is however very complicated and not practical. Understanding this issue better is an interesting future direction.

{
\bibliographystyle{informs2014}
\bibliography{ref}

\begin{thebibliography}{44}
\providecommand{\natexlab}[1]{#1}
\providecommand{\url}[1]{\texttt{#1}}
\providecommand{\urlprefix}{URL }

\bibitem[{Agrawal et~al.(1988)Agrawal, Hedge, \protect\BIBand{}
  Teneketzis}]{agrawal1988asymptotically}
Agrawal R, Hedge M, Teneketzis D (1988) Asymptotically efficient adaptive
  allocation rules for the multiarmed bandit problem with switching cost.
  \emph{IEEE Transactions on Automatic Control} 33(10):899--906.

\bibitem[{Agrawal et~al.(1990)Agrawal, Hegde, \protect\BIBand{}
  Teneketzis}]{agrawal1990multi}
Agrawal R, Hegde M, Teneketzis D (1990) Multi-armed bandit problems with
  multiple plays and switching cost. \emph{Stochastics and Stochastic Reports}
  29(4):437--459.

\bibitem[{Agrawal et~al.(2019)Agrawal, Avadhanula, Goyal, \protect\BIBand{}
  Zeevi}]{agrawal2019mnl}
Agrawal S, Avadhanula V, Goyal V, Zeevi A (2019) Mnl-bandit: A dynamic learning
  approach to assortment selection. \emph{Operations Research}
  67(5):1453--1485.

\bibitem[{Altschuler \protect\BIBand{} Talwar(2018)}]{altschuler2018online}
Altschuler J, Talwar K (2018) Online learning over a finite action set with
  limited switching. \emph{Conference on Learning Theory}, 1569--1573.

\bibitem[{Asawa \protect\BIBand{} Teneketzis(1996)}]{asawa1996multi}
Asawa M, Teneketzis D (1996) Multi-armed bandits with switching penalties.
  \emph{IEEE Transactions on Automatic Control} 41(3):328--348.

\bibitem[{Auer et~al.(2002)Auer, Cesa-Bianchi, \protect\BIBand{}
  Fischer}]{auer2002finite}
Auer P, Cesa-Bianchi N, Fischer P (2002) Finite-time analysis of the multiarmed
  bandit problem. \emph{Machine Learning} 47(2-3):235--256.

\bibitem[{Banks \protect\BIBand{} Sundaram(1994)}]{banks1994switching}
Banks JS, Sundaram RK (1994) Switching costs and the gittins index.
  \emph{Econometrica} 62(3):687--694.

\bibitem[{Bayati et~al.(2015)Bayati, Lelarge, \protect\BIBand{}
  Montanari}]{bayati2015universality}
Bayati M, Lelarge M, Montanari A (2015) Universality in polytope phase
  transitions and message passing algorithms. \emph{Annals of Applied
  Probability} 25(2):753--822.

\bibitem[{Bergemann \protect\BIBand{}
  V{\"a}lim{\"a}ki(2001)}]{bergemann2001stationary}
Bergemann D, V{\"a}lim{\"a}ki J (2001) Stationary multi-choice bandit problems.
  \emph{Journal of Economic Dynamics and Control} 25(10):1585--1594.

\bibitem[{Brezzi \protect\BIBand{} Lai(2002)}]{brezzi2002optimal}
Brezzi M, Lai TL (2002) Optimal learning and experimentation in bandit
  problems. \emph{Journal of Economic Dynamics and Control} 27(1):87--108.

\bibitem[{Cesa-Bianchi et~al.(2013)Cesa-Bianchi, Dekel, \protect\BIBand{}
  Shamir}]{cesa2013online}
Cesa-Bianchi N, Dekel O, Shamir O (2013) Online learning with switching costs
  and other adaptive adversaries. \emph{Advances in Neural Information
  Processing Systems}, 1160--1168.

\bibitem[{Chen \protect\BIBand{} Chao(2019)}]{chen2019parametric}
Chen B, Chao X (2019) Parametric demand learning with limited price
  explorations in a backlog stochastic inventory system. \emph{IISE
  Transactions} 1--9.

\bibitem[{Chen et~al.(2020)Chen, Chao, \protect\BIBand{} Wang}]{chen2020data}
Chen B, Chao X, Wang Y (2020) Data-based dynamic pricing and inventory control
  with censored demand and limited price changes. \emph{Operations Research}
  68(5):1445--1456.

\bibitem[{Cheung et~al.(2017)Cheung, Simchi-Levi, \protect\BIBand{}
  Wang}]{cheung2017dynamic}
Cheung WC, Simchi-Levi D, Wang H (2017) Dynamic pricing and demand learning
  with limited price experimentation. \emph{Operations Research}
  65(6):1722--1731.

\bibitem[{Christofides(1976)}]{christofides1976worst}
Christofides N (1976) Worst-case analysis of a new heuristic for the travelling
  salesman problem. Technical report, Carnegie Mellon University Management
  Sciences Research Group.

\bibitem[{Chu \protect\BIBand{} Chueh(1966)}]{chu1966inequalities}
Chu J, Chueh J (1966) Inequalities between information measures and error
  probability. \emph{Journal of the Franklin Institute} 282(2):121--125.

\bibitem[{Cormen et~al.(2009)Cormen, Leiserson, Rivest, \protect\BIBand{}
  Stein}]{cormen2009introduction}
Cormen TH, Leiserson CE, Rivest RL, Stein C (2009) \emph{Introduction to
  algorithms} (MIT Press).

\bibitem[{Dekel et~al.(2014)Dekel, Ding, Koren, \protect\BIBand{}
  Peres}]{dekel2014bandits}
Dekel O, Ding J, Koren T, Peres Y (2014) Bandits with switching costs: T 2/3
  regret. \emph{Proceedings of the forty-sixth annual ACM symposium on Theory
  of computing}, 459--467.

\bibitem[{den Boer(2015)}]{den2015dynamic}
den Boer AV (2015) Dynamic pricing and learning: historical origins, current
  research, and new directions. \emph{Surveys in operations research and
  management science} 20(1):1--18.

\bibitem[{Domb(2000)}]{domb2000phase}
Domb C (2000) \emph{Phase Transitions and Critical Phenomena}, volume~1
  (Elsevier).

\bibitem[{Dong et~al.(2020)Dong, Li, Zhang, \protect\BIBand{}
  Zhou}]{dong2020multinomial}
Dong K, Li Y, Zhang Q, Zhou Y (2020) Multinomial logit bandit with low
  switching cost. \emph{International Conference on Machine Learning},
  2607--2615 (PMLR).

\bibitem[{Gao et~al.(2019)Gao, Han, Ren, \protect\BIBand{}
  Zhou}]{gao2019batched}
Gao Z, Han Y, Ren Z, Zhou Z (2019) Batched multi-armed bandits problem.
  \emph{Advances in Neural Information Processing Systems}, 503--513.

\bibitem[{Gerchinovitz et~al.(2020)Gerchinovitz, M{\'e}nard, \protect\BIBand{}
  Stoltz}]{gerchinovitz2020fano}
Gerchinovitz S, M{\'e}nard, Stoltz G (2020) Fano’s inequality for random
  variables. \emph{Statistical Science} 35(2):178--201.

\bibitem[{Guha \protect\BIBand{} Munagala(2009)}]{guha2009multi}
Guha S, Munagala K (2009) Multi-armed bandits with metric switching costs.
  \emph{International Colloquium on Automata, Languages, and Programming},
  496--507 (Springer).

\bibitem[{Guha \protect\BIBand{} Munagala(2013)}]{guha2013approximation}
Guha S, Munagala K (2013) Approximation algorithms for bayesian multi-armed
  bandit problems. \emph{arXiv preprint arXiv:1306.3525} .

\bibitem[{Herbster \protect\BIBand{} Warmuth(1998)}]{herbster1998tracking}
Herbster M, Warmuth MK (1998) Tracking the best expert. \emph{Machine Learning}
  32(2):151--178.

\bibitem[{Hu et~al.(2020)Hu, Kallus, \protect\BIBand{} Mao}]{hu2020smooth}
Hu Y, Kallus N, Mao X (2020) Smooth contextual bandits: Bridging the parametric
  and non-differentiable regret regimes. \emph{Conference on Learning Theory},
  2007--2010 (PMLR).

\bibitem[{Jun et~al.(2017)Jun, Orabona, Wright, \protect\BIBand{}
  Willett}]{jun2017online}
Jun KS, Orabona F, Wright S, Willett R (2017) Online learning for changing
  environments using coin betting. \emph{Electronic Journal of Statistics}
  11(2):5282--5310.

\bibitem[{Jun(2004)}]{jun2004survey}
Jun T (2004) A survey on the bandit problem with switching costs. \emph{De
  Economist} 152(4):513--541.

\bibitem[{Lai \protect\BIBand{} Robbins(1985)}]{lai1985asymptotically}
Lai TL, Robbins H (1985) Asymptotically efficient adaptive allocation rules.
  \emph{Advances in Applied Mathematics} 6(1):4--22.

\bibitem[{Lattimore \protect\BIBand{}
  Szepesv{\'a}ri(2020)}]{lattimore2020bandit}
Lattimore T, Szepesv{\'a}ri C (2020) \emph{Bandit algorithms} (Cambridge
  University Press).

\bibitem[{Lawler et~al.(1985)Lawler, Lenstra, Kan, \protect\BIBand{}
  Shmoys}]{lawler1985traveling}
Lawler EL, Lenstra JK, Kan AR, Shmoys DB (1985) \emph{The traveling salesman
  problem: a guided tour of combinatorial optimization}, volume~3 (New York:
  Wiley).

\bibitem[{Ordentlich \protect\BIBand{}
  Weinberger(2005)}]{ordentlich2005distribution}
Ordentlich E, Weinberger MJ (2005) A distribution dependent refinement of
  pinsker's inequality. \emph{IEEE Transactions on Information Theory}
  51(5):1836--1840.

\bibitem[{Perchet et~al.(2016)Perchet, Rigollet, Chassang, \protect\BIBand{}
  Snowberg}]{perchet2016batched}
Perchet V, Rigollet P, Chassang S, Snowberg E (2016) Batched bandit problems.
  \emph{The Annals of Statistics} 44(2):660--681.

\bibitem[{Portnoy(1984)}]{portnoy1984asymptotic}
Portnoy S (1984) Asymptotic behavior of $m$-estimators of $p$ regression
  parameters when $p^2/n $ is large. i. consistency. \emph{The Annals of
  Statistics} 12(4):1298--1309.

\bibitem[{Portnoy(1988)}]{portnoy1988asymptotic}
Portnoy S (1988) Asymptotic behavior of likelihood methods for exponential
  families when the number of parameters tends to infinity. \emph{The Annals of
  Statistics} 356--366.

\bibitem[{Scarlett \protect\BIBand{} Cevher(2019)}]{scarlett2019introductory}
Scarlett J, Cevher V (2019) An introductory guide to fano's inequality with
  applications in statistical estimation. \emph{arXiv preprint
  arXiv:1901.00555} .

\bibitem[{Simchi-Levi et~al.(2008)Simchi-Levi, Kaminsky, Simchi-Levi,
  \protect\BIBand{} Shankar}]{simchi2008designing}
Simchi-Levi D, Kaminsky P, Simchi-Levi E, Shankar R (2008) \emph{Designing and
  managing the supply chain: concepts, strategies and case studies} (Tata
  McGraw-Hill Education).

\bibitem[{Simchi-Levi \protect\BIBand{} Xu(2019)}]{simchi2019phase}
Simchi-Levi D, Xu Y (2019) Phase transitions and cyclic phenomena in bandits
  with switching constraints. \emph{Advances in Neural Information Processing
  Systems}, 7521--7530.

\bibitem[{Slivkins(2019)}]{slivkins2019introduction}
Slivkins A (2019) Introduction to multi-armed bandits. \emph{arXiv preprint
  arXiv:1904.07272} .

\bibitem[{Talebi Mazraeh~Shahi(2017)}]{talebi2017minimizing}
Talebi Mazraeh~Shahi MS (2017) \emph{Minimizing regret in combinatorial bandits
  and reinforcement learning}. Ph.D. thesis, KTH Royal Institute of Technology.

\bibitem[{Tebbe \protect\BIBand{} Dwyer(1968)}]{tebbe1968uncertainty}
Tebbe D, Dwyer S (1968) Uncertainty and the probability of error. \emph{IEEE
  Transactions on Information theory} 14(3):516--518.

\bibitem[{Tsybakov(2008)}]{tsybakov2008introduction}
Tsybakov AB (2008) \emph{Introduction to nonparametric estimation} (Springer
  Science \& Business Media).

\bibitem[{Wainwright(2009)}]{wainwright2009sharp}
Wainwright MJ (2009) Sharp thresholds for high-dimensional and noisy sparsity
  recovery using $\ell_1$-constrained quadratic programming (lasso). \emph{IEEE
  transactions on information theory} 55(5):2183--2202.

\end{thebibliography}
}

\newpage
\appendixpage

\begin{appendices}
\crefalias{section}{appendix}
\crefalias{subsection}{appendix}
\crefalias{subsubsection}{appendix}

\noindent{\large \bf Part I. Additional Results and Explanations}

\section{Results on Distribution-Dependent Regret Bounds}\label{app:dd}
For simplicity, we only present the results of distribution-dependent regret bounds for the \UB problem. Extensions to general switching cost structures are analogous to \pref{sec:general} of the main article.

To achieve tight distribution-dependent regret bounds, we propose the \textsf{LS-SE2} algorithm and the \textsf{AdaLS2} algorithm, which are stated in \pref{alg:ddlsse} and \pref{alg:ddnew} respectively. Note that the difference between the new algorithms and the original algorithms in \pref{sec:unit} is only on the epoch schedules (which are optimized for distribution-dependent regret).

\begin{algorithm}[htbp]\linespread{0.7}\selectfont{}
\caption{Limited-Switch Successive Elimination 2(\textsf{LS-SE2})}
\label{alg:ddlsse}
{\bf Input:} Switching budget $S$, number of actions $K$, horizon $T$. \\
{\bf Initialization:} Compute $q(S,K)=\left\lfloor\frac{S-1}{K-1}\right\rfloor$. Divide the entire time horizon $T$ into $q(S,K)+1$ epochs: $(t_0:t_1],(t_1:t_2],\dots,(t_{q(S,K)}:t_{q(S,K)+1}]$, where the endpoints are defined by $t_0=0$ and
$$
t_j=\left\lfloor K^{1-\frac{j}{{q(S,K)+1}}}{T}^{\frac{j}{{q(S,K)+1}}}\right\rfloor,~~\forall j=1,\dots,q(S,K)+1.
$$%
Let $A_1=[K]$. Let $a_0$ be a random action in $[K]$.\vspace{0.1in}\\
{\bf Policy:} The same as Lines 1 to 10 of \pref{alg:lsse}.
\end{algorithm}

\begin{algorithm}[htbp]\linespread{0.7}\selectfont{}
\caption{Adaptive Limited-Switch Policy 2 (\textsf{AdaLS2})}
\label{alg:ddnew}
{\bf Input:} Switching budget $S$, number of actions $K$, horizon $T$. \\
{\bf Initialization:} Compute $q(S,K)=\left\lfloor\frac{S-1}{K-1}\right\rfloor$ and $r(S,K)=(S-1)\%(K-1)$. Define $\wh{r}(S,K)=\max\crl{r(S,K)+1-q(S,K),0}$. Define $t_0^{(1)}=t_0^{(2)}=0$ and 
\[
{t}_{j}^{(1)}=\floor*{\prn*{K-r(S,K)}^{1-\frac{j}{q(S,K)+1}}{{T}}^{\frac{j}{{q(S,K)+1}}}},~~\forall j=1,\dots,q(S,K)+1,
\]
\[
t_j^{(2)}=\floor*{K^{1-\frac{j}{{q(S,K)+2}}}{{T}}^{\frac{j}{{q(S,K)+2}}}},~~\forall j=1,\dots,q(S,K)+2.
\]
Let $A_1=[K]$. Let $A_1^{(2)}$ be a subset of $A_1$ obtained by uniformly sampling $\wh{r}(S,K)$ actions from $A_1$ \emph{without replacement} (thus $\abs{A_1^{(2)}}=\wh{r}(S,K)$). Let $A_1^{(1)}=A_1\setminus A_1^{(2)}$. Let $a_0$ be a random action in $A_1^{(1)}$.\vspace{0.1in}\\
{\bf Policy:} The same as Lines 1 to 18 of \pref{alg:new}.
\end{algorithm}

For any environment $\cD$, let $k^*=\arg\max_{k\in[K] \mu_i}$ denote the optimal action, and $\Delta=\Delta(\cD)=\min_{k\ne k^*}|\mu_{k^*}-\mu_k|>0$ denote the gap between the rewards of the optimal action and the best sub-optimal action. We have the following upper and lower bounds on regret.

\begin{proposition}\label{thm:up-unit-dd}
Let $\pi$ be the \textsf{LS-SE2} policy. There exists an absolute constant $C\ge0$ such that for all $\cD$, for all $K\ge1$, $S\ge0$ and $T\ge K$,
$$
R_{\cD}^\pi(T)\le C\left(K^{1-\frac{1}{q(S,K)+1}}\log K\right) \frac{T^{\frac{1}{q(S,K)+1}}\log T}{\Delta},
$$
where $q(S,K)=\left\lfloor\frac{S-1}{K-1}\right\rfloor$.
\end{proposition}

\begin{theorem}\label{thm:new-dd}
Let ${\pi}$ be the \textsf{AdaLS2} policy. There exists an absolute constant $C\ge0$ such that for all $\cD$, for all $K\ge1$, $S\ge0$ and $T\ge K$,
\[
R^\pi_{\cD}(K,T)\le C(\log K\log T)\cdot \max\crl*{\frac{\prn*{K-{r}(S,K)}^{2-\frac{1}{q(S,K)+1}}}{K}\frac{T^{\frac{1}{q(S,K)+1}}}{\Delta},\ {{K}^{1-\frac{1}{q(S,K)+1}}}\frac{T^{\frac{1}{q(S,K)+1}}}{\Delta}},
\]
where $q(S,K)=\left\lfloor\frac{S-1}{K-1}\right\rfloor$ and $r(S,K)=(S-1)\%(K-1)$.
\end{theorem}

\begin{theorem}\label{thm:lb-unit-dd}
There exists an absolute constant $C>0$ such that for all $K> 1,S\ge0, T\ge 2K$ and for all policy $\pi\in\Pi_S$,
\begin{equation*}
    \sup_{\Delta\in[0,1]}\Delta R^\pi_{\cD}(K,T)\ge\frac{C}{\log T}\cdot \max\crl*{\frac{\prn*{K-{r}(S,K)}^{2-\frac{1}{{q(S,K)+1}}}}{K}T^{\frac{1}{{q(S,K)+1}}},\ {{K}^{1-\frac{1}{{q(S,K)+2}}}}T^{\frac{1}{{q(S,K)+2}}}},
\end{equation*}
where $q(S,K)=\left\lfloor\frac{S-1}{K-1}\right\rfloor$ and $r(S,K)=(S-1)\%(K-1)$.
\end{theorem}

Note that the upper bound in \pref{thm:new-dd} and the lower bound in \pref{thm:lb-unit-dd} match in the minimax sense (up to logarithmic factors), which implies that the \textsf{AdaLS2} algorithm can be considered as near-optimal. We thus characterize the distribution-dependent complexity of the \UB problem. 
We also note that when $S=\Omega(K\log T)$, both \textsf{LS-SE2} and \textsf{AdaLS2} algorithms recover the well-known $\cO\prn*{\frac{K\log T}{\Delta}}$ distribution-dependent regret bound of the classical \MAB (up to a $\log K$ factor), which is shown to be rate-optimal (\citealt{lai1985asymptotically}).

We omit the proofs of above results: the proof of \pref{thm:up-unit-dd} resembles the proof of \pref{thm:up-unit} in \pref{app:proof-thm1}, the proof of \pref{thm:new-dd} resembles the proof of \pref{thm:ub-n} in \pref{app:upper}, 
and the proof of \pref{thm:lb-unit-dd} resembles the proof of \pref{thm:lb-unit} in \pref{app:lower}. The difference is mainly on the partition of epochs.

Besides results on regret upper and lower bounds, we also establish \pref{cor:sc-dd}, which can be viewed as a counterpart of \pref{cor:sc} in \pref{sec:nec} of the main article.

\begin{corollary}\label{cor:sc-dd}
For any $K\ge1$, for any environment $\cD$, let $\Delta=\min\limits_{k\in[K],k\ne k^*}|\mu_{k^*}-\mu_k|$ denote the gap between the mean rewards of the optimal action and the best sub-optimal action.
\begin{enumerate}
    \item $N(K-1)+1$ switches are necessary and sufficient for uniformly achieving $\wt{\cO}(KT^\frac{1}{N+1}/\Delta)$ distribution-dependent regret  for all $\cD$ in the $K$-armed \MAB ($N\in\mathbb{Z}_{>0}$).
    \item $\Omega(\frac{K\log T}{\log\log T})$ switches are necessary for uniformly achieving $\wt{\cO}(K\log{T}/\Delta)$ distribution-dependent regret for all $\cD$ in the $K$-armed \MAB.
\end{enumerate}
\end{corollary}

\section{Rounding Issues of Algorithms}\label{app:rounding}
We present a more rigorous version of \pref{alg:lsse}, which takes care of the rounding issues in \pref{line:3} and \pref{line:5} of \pref{alg:lsse}. The key idea is to maintain a ``after-rounding''  epoch schedule $(T_j)_{j=0}^{q(S,K)+1}$ which is slightly different from the original epoch schedule $(t_j)_{0}^{q(S,K)+1}$; see \pref{alg:lsse-r}. In the proof of \pref{thm:ub-n}, we will directly analyze \pref{alg:lsse-r}.

We remark that the main algorithm of our article, the \algon algorithm (\pref{alg:new}), has \emph{already} taken care of the rounding issues (using a similar idea).The other two algorithms \algog and \algoa omit the rounding issues --- they can be easily modified to incorporate the rounding issues, using exactly the same idea as \pref{alg:lsse-r}.

\begin{algorithm}[htbp]\linespread{0.7}\selectfont{}
\caption{Limited-Switch Successive Elimination (\algoo)}
\label{alg:lsse-r}
{\bf Input:} Switching budget $S$, number of actions $K$, horizon $T$. \\
{\bf Initialization:} Compute $q(S,K)=\left\lfloor\frac{S-1}{K-1}\right\rfloor$. Define $t_0=0$ and
$$
t_j=\left\lfloor K^{1-\frac{2-2^{-(j-1)}}{2-2^{-q(S,K)}}}{T}^{\frac{2-2^{-(j-1)}}{2-2^{-q(S,K)}}}\right\rfloor,~~\forall j=1,\dots,q(S,K)+1.
$$
Let $A_1=[K]$. Let $a_0$ be a random action in $[K]$. Let $T_l=0$.\\
{\bf Policy:}
\begin{algorithmic}[1]
\For{$l=1,\dots,q(S,K)$}
\If{$a_{T_{l-1}}\in A_{l}$}
\For{$i=a_{T_{l-1}}$ and {then} $i\in A_l\setminus\crl{a_{T_{l-1}}}$} {\color{blue}\Comment{starting from $i=a_{T_{l-1}}$ is critical}}
\State{Choose action $i$ for ${\floor*{\frac{t_{l}-T_{l-1}}{|A_l|}}}$ \emph{consecutive} rounds.}
\EndFor
\Else 
\For{$i\in A_l$}
\State{Choose action $i$ for $\floor*{{\frac{t_{l}-T_{l-1}}{|A_l|}}}$ \emph{consecutive} rounds.}
\EndFor
\EndIf
\State{Mark the last round as $T_l$, and mark the last chosen  action as $a_{T_l}$.{\color{blue}\Comment record $T_l$
}}
\State{Elimination:}
compute $\texttt{UCB}_{i}(T_l)$ and $\texttt{LCB}_{i}(T_l)$ for all $i\in A_l$ and let {\color{blue}\Comment learn from data
}
\[A_{l+1}=\crl*{i\in A_l \mid \texttt{UCB}_{i}(T_l)\ge \max_{j\in A_l} \texttt{LCB}_{j}(T_l)}.\]
\EndFor
\State{For $l=q(S,K)+1$, compute an action in $A_l$ that maximizes $\wb{\mu}_i(T_l)$. Keep choosing this action until round $T$. Let $T_{q(S,K)+1}\ldef T$.}
\end{algorithmic}
\end{algorithm}

\section{Illustration of AdaLS}\label{app:illustration}
We use the example of $S=2K-2$ to illustrate how \algon utilizes the switching budget more efficiently than \algoo. In this case, $q(S,K)=1$ and $r(S,K)=K-2$. The \algoo algorithm will observe the data only once throughout the entire horizon, and makes at most $K$ switches.

How does \algon behave in this case? At initialization, \algon computes $\wh{r}(S,K)=\max\crl{r(S,K)+1-q(S,K),0}=K-2+1-1=K-2$. The algorithm then randomly splits $[K]$ into two subsets $A_1^{(1)}$ and $A_1^{(2)}$, with $\abs{A_1^{(1)}}=K-\wh{r}(S,K)=2$ and $\abs{A_1^{(2)}}=\wh{r}(S,K)=K-2$. Then, in the execution of the policy, \algon treats the actions in $A_1^{(1)}$ and $A_1^{(2)}$ differently, allowing  the actions in $A_1^{(2)}$ to be explored more frequently than the actions in $A_1^{(1)}$. Specifically, in the first epoch, \algon explores all actions in $[K]$ and makes $K-1$ switches; then, in the second epoch, \algon first explores all uneliminated actions in $A_1^{(2)}$ (which incurs at most $(K-2)-1=K-3$ switches), and finally commits to the best action (which incurs at most 1 switch). Note that \algon may also incur a switch between the first and second epochs, so its total number of switches is at most $(K-1)+1+(K-3)+1=2K-2$ (assuming no action is eliminated). Clearly, compared with \algoo which only makes $K$ switches, \algon makes much better use of the switching budget in this case.

\section{Explanations for \pref{sec:sym}}
This section contains some additional explanations for our results in \pref{sec:sym}.

\subsection{Relaxing the Triangle Inequality Assumption in \pref{sec:sym}}\label{app:relax}
Consider an arbitrary switching graph $G$ with $K=|G|> 1$. In the following, we show that, even without the triangle inequality assumption, a modified version of the results in \pref{sec:sym} still hold.

\subsubsection{Construction of a New Switching Graph that Satisfies the Triangle Inequality}\label{app:relax1}

Assume that the switching costs associated with $G$ do not satisfy the triangle inequality. We then run the Floyd-Warshall algorithm (see \citealt{cormen2009introduction}) on $G$ to efficiently find the shortest paths between all pairs of vertices. For any $i,j\in[K]$ such that $i\ne j$, let $p_{i,j}=i\rightarrow\dots\rightarrow j$ denote the shortest path between $i$ and $j$, and $c_{i,j}'$ denote the total weight of the shortest path between $i$ and $j$. We construct a new switching graph $G'=(V,E')$ --- the vertices in $G'$ are the same as $G$, while the edge between $i$ and $j$  in $G'$ is assigned a weight $c_{i,j}'$, which is the total weight of the shortest path between $i$ and $j$ in $G$. Obviously, $G'$ is a switching graph whose switching costs satisfy the triangle inequality. Therefore, for \BwSC problems defined with $G'$, we can apply the \algog policy, and the regret upper and lower bounds in \pref{thm:gub} and \pref{thm:glb} in Section \ref{sec:sym} hold.

\subsubsection{Modification of the HS-SE policy}\label{app:relax2}

In this part we assume that $K=\cO(1)$. %

For any \GB problem defined with switching graph $G$ (whose switching costs do not satisfy the triangle inequality) and switching budget $S$, we construct a new switching graph $G'$ according to Appendix \ref{app:relax1}, and construct a new \GB problem defined with switching graph $G'$ and switching budget $S$. Let $\pi'$ denote the \algog policy running on the new \GB problem. Obviously $\pi'$ is a $S$-switching budget policy for the new problem. We construct $\pi$ by modifying $\pi'$, aiming to obtain an $S$-switching-budget policy for the original \GB problem. Let $\pi$ switch (on $G$) following $\pi'$ (on $G'$): every time $\pi'$ switches from $i$ to $j$ on $G'$, let $\pi$ switch according to the path $p_{i,j}=i\rightarrow\dots\rightarrow j$ on $G$, visiting each vertex in $p_{i,j}$ once (since in the HS-SE policy, every active action is chosen for at least $\Omega(T^{1/2})$ consecutive rounds in each epoch, while $p_{i,j}$ contains at most $K=o(\sqrt{T})$ vertices, we know that $\pi'$ is a valid policy). Since the total weight of $p_{i,j}$ is $c'_{i,j}$ and $\pi'$ is an $S$-switching-budget policy for $G'$, we know that $\pi$ is an $S$-switching-budget policy for $G$.

\subsection{Computation of the Offline Step in the HS-SE Policy}\label{app:comp}

The \algog policy is practical --- for any given switching graph $G$, the policy only involves solving the shortest Hamiltonian path problem once, which can be finished \textit{offline}. Thus, the computational complexity of the shortest Hamiltonian path problem does not affect the online decision-making process of the \algog policy. %

Moreover, under the condition that the switching costs satisfy the triangle inequality, the shortest Hamiltonian path problem can be reduced to the celebrated \textit{metric traveling salesman problem} (metric TSP), see \cite{lawler1985traveling}. This means that we can directly apply many commercial solvers for the TSP to solve (or approximately solve) the shortest Hamiltonian path problem efficiently. The reduction also indicates that any approximation algorithm designed for the metric TSP can be adapted to be an approximation algorithm for the shortest Hamiltonian path problem. In particular, the celebrated Christofides algorithm for the metric TSP  (\citealt{christofides1976worst}) can be used to compute a good approximation of $H$ in polynomial time.

\vspace{2em}
\noindent{\large \bf Part II. Proof of Upper Bounds}

\section{Proof of \pref{thm:up-unit}}\label{app:proof-thm1}

Note that our proof is based on the more rigorous version of \algoo{} --- \pref{alg:lsse-r}.

\subsection{LS-SE is Indeed an $S$-Switch Policy}

From round 1 to round $t_1$, \algoo makes $K-1$ switches. For $1\le l\le q(S,K)-1$, from round $T_l$ to round $T_{l+1}$:
\begin{itemize}
    \item If the last action in epoch $l$ remains active in epoch $l+1$, then it will be the first action in epoch $l+1$, and no switch occurs between round $T_l$ and round $T_l+1$. Since \algoo makes at most $K-1$ switches within epoch $l+1$, i.e., from round $T_l+1$ to round $T_{l+1}$, it makes at most $0+(K-1)=K-1$ switches from round $T_l$ to round $T_{l+1}$.
    \item If the last action in epoch $l$ is eliminated before the start of epoch $l+1$, then epoch $l+1$ starts from another active action, and one switch occurs between round $T_l$ and round $T_l+1$. The elimination implies that $|A_{l+1}|\le K-1$, thus \algoo makes $|A_{l+1}|-1\le (K-1)-1=K-2$ switches within epoch $l+1$, i.e., from round $T_l+1$ to round $T_{l+1}$. Therefore, the SS-SE policy makes at most $1+(K-2)=K-1$ switches from round $T_l$ to round $T_{l+1}$.
\end{itemize}
From round $T_{q(S,K)}$ to round $T$, since \algoo does not switch within epoch $q(S,K)+1$, i.e., from round $T_{q(S,K)}+1$ to round $T$, the only possible switch is between round $T_{q(S,K)}$ and $T_{q(S,K)}+1$. Thus \algoo makes at most 1 switch from round $T_{q(S,K)}$ to round $T$.

Summarizing the above arguments, we find that the SS-SE policy makes at most $q(S,K)(K-1)+1\le S$ switches from round 1 to round $T$. Thus it is indeed an $S$-switching-budget policy.

\subsection{Proof of Upper Bound}\label{app:pub}
Since \algoo is a $(q(S,K)+1)$-batch policy, existing upper bound analysis for batched \MAB (\citealt{gao2019batched}) applies here. Still, we present our upper bound proof for completeness. A difference is that we obtain  better dependence on $K$ under the condition $\sup_{i,j\in[K]}\abs{\mu_i-\mu_j}\in[0,1]$.

If $K>T/4$, then the upper bound in \pref{thm:up-unit} becomes $\cO(T)$ and is trivial. Therefore, without loss of generality, we assume that $T\ge 4K$.

We start the proof of the upper bound on regret with some definitions.  Define the confidence radius as
$$
r_i(t)=\sqrt{\frac{6\log T}{N_i(t)}},~~\forall i\in[K], t\in[T].
$$
The $\texttt{UCB}_{i}(t)$ and $\texttt{LCB}_{i}(t)$ confidence bounds defined in \pref{eq:confbounds} can be expressed as
$$
\texttt{UCB}_{i}(t)=\wb{\mu}_{i}(t)+r_{i}(t),~~\forall i\in[K], t\in[T],
$$
$$
\texttt{LCB}_{i}(t)=\wb{\mu}_{i}(t)-r_{i}(t),~~\forall i\in[K], t\in[T].
$$
Define the \textit{clean event} as $$\mathcal{E}:=\{\forall i\in[K], \forall t\in[T],~~|\bar{\mu}_i(t)-\mu_i|\le r_t(i)\}.$$ By Hoeffding's inequality for sub-Gaussian variables and a standard union bound argument (see, e.g., 
Lemma 1.5 in \citealt{slivkins2019introduction}), since $T\ge K$, for any policy $\pi$ and any environment $\cD$, we always have $\mathbb{P}_{\cD}^{\pi}(\mathcal{E})\ge 1-\frac{2}{T^3}\cdot T\cdot K=1-\frac{2}{T}$. Define the \textit{bad event} $\wb{\mathcal{E}}$ as the complement of the clean event.

Let $\pi$ denote the \algoo policy. First, observe that for any environment $\cD$,
\begin{align}\label{eq:cle}
R_{\cD}^{\pi}(T)&=\mathbb{E}_{\cD}^{\pi}\left[T\mu^*-\sum_{t=1}^{T}\mu_{a_t}\mid \mathcal{E}\right]\mathbb{P}_{\cD}^{\pi}(\mathcal{E})+\mathbb{E}_{\cD}^{\pi}\left[T\mu^*-\sum_{t=1}^{T}\mu_{a_t}\mid \wb{\mathcal{E}}\right]\mathbb{P}_{\cD}^{\pi}(\wb{\mathcal{E}})\notag\\
&\le\mathbb{E}_{\cD}^{\pi}\left[T\mu^*-\sum_{t=1}^{T}\mu_{a_t}\mid \mathcal{E}\right]+ T\cdot\frac{2}{T}\notag\\&=\mathbb{E}_{\cD}^{\pi}\left[T\mu^*-\sum_{t=1}^{T}\mu_{a_t}\mid \mathcal{E}\right]+2,
\end{align}
so in order to bound $R^{\pi}(T)=\sup_{\cD}R_{\cD}^{\pi}(T)$, we only need to focus on the clean event.

Consider an arbitrary environment $\cD$ and assume the occurrence of the clean event. By the specification of \pref{alg:lsse-r}, we know that the optimal action $k^*\in A_l$ for all $l\in[q(S,K)+1]$. 
For any $k\in[K]$, define $\eta_k\ldef\max\crl{j\in[q(S,K)+1]\mid \text{action }k\text{ is chosen in epoch }j}$, i.e., $\eta_k$ the index of the last epoch where the learner chooses action $k$. Consider any action $k$ such that $\mu_{k}<\mu_{k^*}$. By the specification of \pref{alg:lsse-r}, if $\eta_k>1$, then the confidence intervals of the two actions $k^*$ and $k$ at the end of round $T_{\eta_k-1}$ must overlap, i.e., $\texttt{UCB}_{k}(T_{\eta_k-1})\ge\texttt{LCB}_{k^*}(T_{\eta_k-1})$. Therefore,
\begin{equation}\label{eq:upinq}
  \Delta(k):=\mu_{k^*}-\mu_k\le 2 r_{k^*}({T_{\eta_k-1}})+2r_{k}({T_{\eta_{k}-1}})= 4r_{k}({T_{\eta_{k}-1}}),  
\end{equation}
where the last equality is because $k^*$ and $k$ are chosen for equal times in each epoch until epoch $\eta_k$, which implies that $N_{k^*}({T_{\eta_i-1}})=N_{k}({T_{\eta_{i}-1}})$. %
Since $k$ is never chosen after the $\eta_k$-th epoch, we have $N_{k}(T_{\eta_k})=N_k(T)$, and therefore $r_{i}(T_{\eta_k})=r_k(T)$.

For any $j\in[q(S,K)]$, since $K\le\frac{1}{{2}}\floor{2K}\le\frac{1}{2}\floor{K\sqrt{T/K}}\le\frac{1}{2}t_j$, we have
\begin{align}\label{eq:compare}
    T_{j}&=T_{j-1}+\abs{A_{j}}\floor*{\frac{t_{j}-T_{j-1}}{\abs{A_{j}}}}\notag\\
    &\ge T_{j-1}+{{t_{j}-T_{j-1}}-\prn{{\abs{A_{j}}}-1}}\notag\\
    & \ge t_{j} - (K-1)\ge\frac{1}{2}t_j.
\end{align}

For any $k\in[K]$, define $R(T;k)\ldef\sum_{t=1}^{T}\prn*{\mu^*-\mu_{k}}\Ind\crl{a_t=k}=\Delta(k)N_k(T)$. For any $k$ such that $\eta_k\in[2:q(S,K)]$, by \pref{eq:upinq},  \pref{eq:compare}, and the specification of \pref{alg:lsse-r}, conditional on the clean event $\cE$,
\begin{align*}
R(T;k)&=N_k(T)\Delta(k)\\&\le 4 N_k({T_{\eta_k}})\sqrt{\frac{6\log T}{N_k(T_{\eta_k-1})}}\\&\le 4\frac{T_{\eta_k}}{|A_{\eta_k}|}\frac{\sqrt{6\log T}}{\sqrt{T_{\eta_k-1}/K}}\\
&\le 4\sqrt{6K\log T}\frac{1}{\abs{A_{\eta_k}}}\frac{T_{\eta_k}}{\sqrt{T_{\eta_k-1}}}\\
&\le 8\sqrt{3K\log T}\frac{1}{\abs{A_{\eta_k}}}\frac{t_{\eta_k}}{\sqrt{t_{\eta_k-1}}}\\
&=8\sqrt{6\log T}\frac{1}{\abs{A_{\eta_k}}}K(T/K)^{\frac{1}{2-2^{-q(S,K)}}}.
\end{align*}
For any $k$ such that $\eta_k=1$, then we have $R(T;k)=N_k(T)\Delta(k)\le N_K(T_1)\le \frac{1}{|A_1|}K(T/K)^{\frac{1}{2-2^{-q(S,K)}}}$. Moreover, we have  $\sum_{k:\eta_k=q(S,K)+1}R(T;k)\le T\Delta(k)\le 4T\sqrt{\frac{6\log T}{N_k(T_{q(S,K)})}}\le 8\sqrt{6\log T} K(T/K)^{\frac{1}{2-2^{-q(S,K)}}}$.

Therefore, for any environment $\cD$, conditional on the clean event $\cE$, we have
\begin{align*}
T\mu^*-\sum_{t=1}^{T}\mu_{a_t}&=\sum_{k\in [K]}R(T;k)\\&=\sum_{k:\eta_k=1}R(T;k)+\sum_{k:\eta_k\in[2:q(S,K)]}R(T;k)+\sum_{k:\eta_k=q(S,K)+1}R(T;k)\\
&\le 8\sqrt{6\log T}K(T/K)^{\frac{1}{2-2^{-q(S,K)}}}\prn*{\sum_{k=1}^{K}\frac{1}{|A_{\eta_{k}}|}+1}\\
&\le 8\sqrt{6\log T}K(T/K)^{\frac{1}{2-2^{-q(S,K)}}}\prn*{\sum_{j=1}^{K}\frac{1}{j}+1}\\
&\le 16\sqrt{6}(\log K\log T)K^{1-\frac{1}{2-2^{-q(S,K)}}}T^{\frac{1}{2-2^{-q(S,K)}}}.
\end{align*}
 Thus by \pref{eq:cle} and $R^{\pi}(T)=\sup_{\cD}R_{\cD}^{\pi}(T)$, we have
\[
R^{\pi}(T)\le 16\sqrt{6}(\log K\log T)K^{1-\frac{1}{2-2^{-q(S,K)}}}T^{\frac{1}{2-2^{-q(S,K)}}}+2.\qed
\]

\section{Proof of \pref{thm:ub-n}}\label{app:upper}
In this proof, we let the tuning parameter $\lambda=1/2$ (see \pref{alg:new}). Our proof essentially holds for all $\lambda\in(0,1)$ being a constant.

\subsection{The AdaLS Policy is Indeed an $S$-Switch Policy}
According to \pref{sec:improved}, before the last switch (where we commit to the empirical best action; see \cref{line:commit}), we allow \algon to switch to each action in $A_1^{(1)}$ for at most $q(S,K)$ times, while allowing it to switch to each action in $A_1^{(2)}$ for at most $q(S,K)+1$ times. Therefore, 
if $\wh{r}(S,K)>0$, then \algon will  make up to \[\underbrace{q(S,K)(K-\wh{r}(S,K))}_{\text{switching to an action in }A_1^{(1)}}+\underbrace{(q(S,K)+1)\wh{r}(S,K)}_{\text{switching to an action in }A_1^{(2)}}-\underbrace{1}_{\text{the first round}}+\underbrace{1}_{\text{the last switch}}=S\] switches. See \pref{app:illustration} for a concrete example when $K=2S-2$.

On the other hand, if $\wh{r}(S,K)=0$, then \algon will behave similar to \algoo and  make up to $q(S,K)(K-1)+1<S$ switches.

To sum up, \algon is indeed an $S$-switch policy.

\subsection{Proof of Upper Bound}
If $K>T/16$, then the upper bound in \pref{thm:ub-n} becomes $\cO(T)$ and is trivial. Therefore, without loss of generality, we assume that $T\ge 16K$.

We start the proof of the upper bound on regret with some definitions.  Define the confidence radius as
$$
r_i(t)=\sqrt{\frac{6\log T}{N_i(t)}},~~\forall i\in[K], t\in[T].
$$
The $\texttt{UCB}_{i}(t)$ and $\texttt{LCB}_{i}(t)$ confidence bounds defined in \pref{eq:confbounds} can be expressed as
$$
\texttt{UCB}_{i}(t)=\wb{\mu}_{i}(t)+r_{i}(t),~~\forall i\in[K], t\in[T],
$$
$$
\texttt{LCB}_{i}(t)=\wb{\mu}_{i}(t)-r_{i}(t),~~\forall i\in[K], t\in[T].
$$
Define the \textit{clean event} as $$\mathcal{E}:=\{\forall i\in[K], \forall t\in[T],~~|\bar{\mu}_i(t)-\mu_i|\le r_t(i)\}.$$ By Hoeffding's inequality for sub-Gaussian variables and a standard union bound argument (see, e.g., 
Lemma 1.5 in \citealt{slivkins2019introduction}), since $T\ge K$, for any policy $\pi$ and any environment $\cD$, we always have $\mathbb{P}_{\cD}^{\pi}(\mathcal{E})\ge 1-\frac{2}{T^3}\cdot T\cdot K=1-\frac{2}{T}$. Define the \textit{bad event} $\wb{\mathcal{E}}$ as the complement of the clean event.

Let $\pi$ denote the \algon policy. First, observe that for any environment $\cD$,
\begin{align}\label{eq:clean}
R_{\cD}^{\pi}(T)&=\mathbb{E}_{\cD}^{\pi}\left[T\mu^*-\sum_{t=1}^{T}\mu_{a_t}\mid \mathcal{E}\right]\mathbb{P}_{\cD}^{\pi}(\mathcal{E})+\mathbb{E}_{\cD}^{\pi}\left[T\mu^*-\sum_{t=1}^{T}\mu_{a_t}\mid \wb{\mathcal{E}}\right]\mathbb{P}_{\cD}^{\pi}(\wb{\mathcal{E}})\notag\\
&\le\mathbb{E}_{\cD}^{\pi}\left[T\mu^*-\sum_{t=1}^{T}\mu_{a_t}\mid \mathcal{E}\right]+ T\cdot\frac{2}{T}\notag\\&=\mathbb{E}_{\cD}^{\pi}\left[T\mu^*-\sum_{t=1}^{T}\mu_{a_t}\mid \mathcal{E}\right]+2,
\end{align}
so in order to bound $R^{\pi}(T)=\sup_{\cD}R_{\cD}^{\pi}(T)$, we only need to focus on the clean event.

Define $\cE_1\ldef\crl*{k^*\in A_1^{(1)}}$ and $\cE_2\ldef\crl*{k^*\in A_1^{(2)}}$. Since $A_1^{(1)}$ and $A_1^{(2)}$ are determined by random sampling independent of the clean event $\cE$, we have
\[
\bbP_{\cD}^{\pi}(\cE_1\mid \cE)=\bbP_{\cD}^{\pi}(\cE_1)=\frac{K-\wh{r}(S,K)}{K},
\]
\[
\bbP_{\cD}^{\pi}(\cE_2\mid \cE)=\bbP_{\cD}^{\pi}(\cE_2)=\frac{\wh{r}(S,K)}{K}.
\]
Thus
\begin{equation}\label{eq:decomp}
    \bbE_{\cD}^{\pi}\brk*{T\mu^*-\sum_{t=1}^{T}\mu_{a_t}\mid \mathcal{E}}=\frac{K-\wh{r}(S,K)}{K}\bbE_{\cD}^{\pi}\brk*{T\mu^*-\sum_{t=1}^{T}\mu_{a_t}\mid \mathcal{E},\cE_1}+\frac{\wh{r}(S,K)}{K}\bbE_{\cD}^{\pi}\brk*{T\mu^*-\sum_{t=1}^{T}\mu_{a_t}\mid \mathcal{E},\cE_2}.
\end{equation}
Note that we define $\bbP_{\cD}^{\pi}(\cdot\mid \cE,\cE_1)\equiv0$ if $\wh{r}(S,K)=0$.

Define $T_{q(S,K)+1}^{(1)}\ldef T$. For all $j\in[q(S,K)+1]$, ``epoch $j$'' corresponds to ``period $[T_{l-1}^{(1)}+1:T_l^{(1)}]$'', and we define 
\[\wt{A}_j\ldef\crl*{k\in A_j\mid \text{action }k\text{ is chosen in epoch }j},\]
\[\wt{A}_j^{(1)}\ldef\crl*{k\in A_j^{(1)}\mid \text{action }k\text{ is chosen in epoch }j},\]
\[
\wt{A}_j^{(2)}\ldef\crl*{k\in A_j^{(2)}\mid \text{action }k\text{ is chosen in epoch }j}.
\] %

For any $j\in[q(S,K)]$, since $K\le\frac{1}{{4}}\floor{4K}\le\frac{1}{4}\floor{K\sqrt{T/K}}\le\frac{1}{4}t_j^{(2)}$, we have
\begin{align}\label{eq:comparen}
    T_{j}^{(1)}&=T_{j-1}^{(1)}+\abs{A_{j}}\floor*{\frac{t_{j}^{(2)}-T_{j-1}^{(1)}}{\abs{A_{j}}}}\notag\\
    &\ge T_{j-1}^{(1)}+{{t_{j}^{(2)}-T_{j-1}^{(1)}}-\prn{{\abs{A_{j}}}-1}}\notag\\
    & \ge t_{j}^{(2)} - (K-1)\ge\frac{1}{2}t_j^{(2)}.
\end{align}
For any $j\in[q(S,K)+2]$, we have
\begin{align}\label{eq:comparen1}
    t_j^{(2)}&=\floor*{K^{1-\frac{2-2^{1-j}}{2-2^{-q(S,K)-1}}}{{T}}^{\frac{2-2^{1-j}}{2-2^{-q(S,K)-1}}}}\notag\\
    &\ge \floor*{{K}^{1-\frac{2-2^{2-j}}{2-2^{-q(S,K)}}}{{T}}^{\frac{2-2^{2-j}}{2-2^{-q(S,K)}}}}\notag\\
    &\ge\floor*{\prn*{K-r(S,K)}^{1-\frac{2-2^{2-j}}{2-2^{-q(S,K)}}}{{T}}^{\frac{2-2^{2-j}}{2-2^{-q(S,K)}}}}=t_{j-1}^{(1)}.
\end{align}

Define $\cR_1\ldef{\prn*{K-{r}(S,K)}^{1-\frac{1}{2-2^{-q(S,K)}}}}T^{\frac{1}{2-2^{-q(S,K)}}}$ and $\cR_2\ldef{{K}^{1-\frac{1}{2-2^{-q(S,K)-1}}}}T^{\frac{1}{2-2^{-q(S,K)-1}}}$.  

Consider an arbitrary environment $\cD$ and assume the occurrence of the clean event $\cE$. In what follows, we discuss two cases: $\cE_1$ occurs, and $\cE_2$ occurs.

\textbf{Case 1: Both $\cE$ and $\cE_1$ occur.} By the specification of \pref{alg:new}, we know that the optimal action $k^*\in A_l^{(1)}\subset A_l$ for all $l\in[q(S,K)+1]$, and $k^*\in A_{q(S,K)+2}$. Also, we have $k^*\in \wt{A}_l^{(1)}\subset \wt{A}_l$ for all $l\in[q(S,K)]$. Moreover, by \pref{line:start} to \pref{line:end} of \pref{alg:new}, we know that action $k^*$ is chosen for
\[
\max\crl*{\floor*{\frac{{t}_l^{(1)}/2-T_l^{(2)}}{\abs{A_l^{(1)}}}},n_l^{(2)}}
\]
rounds in each epoch $l\in[q(S,K)]$, which is no less than the number of plays of any other action in epoch $l$. Therefore, for all $k\in[K]$ and $l\in[q(S,K)]$ we have
\begin{equation}\label{eq:choice}
    N_{k^*}(T_l^{(1)})\ge N_{k}(T_l^{(1)}),~~r_{k^*}(T_l^{(1)})\le r_{k}(T_l^{(1)}).
\end{equation}

For any $k\in[K]$, define $\eta_k\ldef\max\crl*{j\in[q(S,K)+1]\mid k\in\wt{A}_j}$.

Consider any action $k$ such that $\eta_k>1$.  By \pref{line:elimination2} of \pref{alg:new},  the confidence intervals of the two actions $k^*$ and $k$ at the end of round $T_{\eta_k-1}^{(1)}$ must overlap, i.e., $\texttt{UCB}_{k}(T_{\eta_k-1}^{(1)})\ge\texttt{LCB}_{k^*}(T_{\eta_k-1}^{(1)})$. Therefore,
\begin{equation}\label{eq:upinqn}
  \Delta(k):=\mu_{k^*}-\mu_k\le 2 r_{k^*}({T_{\eta_k-1}^{(1)}})+2r_{k}({T_{\eta_{k}-1}^{(1)}})\le 4r_{k}({T_{\eta_{k}-1}^{(1)}}),  
\end{equation}
where the last equality follows from \pref{eq:choice}. 
Since $k$ is never chosen after the $\eta_k$-th epoch, we have $N_{k}(T_{\eta_k}^{(1)})=N_k(T)$.

We now try to prove the following key lemma.
\begin{lemma}\label{lm:bounding} Assume both $\cE$ and $\cE_1$ hold. For any action $k$ such that $\eta_{k}\in[2:q(S,K)]$, we have
\begin{equation}\frac{N_k({T_{\eta_k}^{(1)}})}{\sqrt{N_{k}(T_{\eta_k-1}^{(1)})}}\le\begin{cases}
 \sqrt{N_{k}(T_{\eta_k-1}^{(1)})}+\frac{2}{|A_{\eta_k}^{(1)}|}{\max\crl*{\sqrt{\frac{K-\wh{r}(S,K)}{K-r(S,K)}}\cR_1,\frac{32K}{K-r(S,K)}\cR_2}},&\text{if }k\in A_1^{(1)};\\
  \sqrt{N_{k}(T_{\eta_k-1}^{(1)})}+\frac{3}{|A_{\eta_k}|}\cR_2, &\text{if }k\in A_1^{(2)}.
\end{cases}\label{eq:adals0}
\end{equation}
Moreover, considering all $k$ such that $\eta_k=q(S,K)+1$ (i.e., $k\in\wt{A}_{q(S,K)+1}$), we have
\begin{equation}{\sum_{k\in\wt{A}_{q(S,K)+1}^{(1)}}} \frac{N_k(T)}{\sqrt{N_k(T_{q(S,K)}^{(1)})}}\le {4}{\max\crl*{\sqrt{\frac{K-\wh{r}(S,K)}{K-r(S,K)}}\cR_1,\frac{32K}{K-r(S,K)}\cR_2}}, \label{eq:inq1}\end{equation}
\begin{equation}\sum_{k\in\wt{A}_{q(S,K)+1}^{(2)}}\frac{N_k(T_{q(S,K)+1}^{(2)})}{\sqrt{N_{k}(T_{q(S,K)}^{(1)})}}+\frac{N_k(T)}{\sqrt{N_{k}(T_{q(S,K)+1}^{(2)})}}\le 5\cR_2.\label{eq:inq2}\end{equation}
\end{lemma}

\proof{Proof of \pref{lm:bounding}.} We first show \pref{eq:adals0}. Fix any action $k$ such that $\eta_k\in[2:q(S,K)]$. %

We first consider the case of $k\in{A}_{1}^{(1)}$. By \pref{line:start} to \pref{line:end} of \pref{alg:new}, we know that action $k$ is chosen for
\[
\max\crl*{\floor*{\frac{{t}_l^{(1)}/2-T_l^{(2)}}{\abs{A_l^{(1)}}}},n_l^{(2)}}
\]
rounds in each epoch $l\in[\eta_k-1]$, which is no less than the number of plays of any other action in epoch $l$. This implies \begin{equation}\label{eq:zz}
N_k(T_l^{(1)})\ge T_l^{(1)}/K,~~~\forall l\in[\eta_k-1].
\end{equation} Since
\begin{align}\label{eq:adals1}
    T_l^{(1)}\le T_l^{(2)}+|A_l^{(1)}| \floor*{\frac{{t}_l^{(1)}/2-T_l^{(2)}}{\abs{A_l^{(1)}}}}\le\frac{t_l^{(1)}}{2}
\end{align}
for all $l\in[\eta_k-1]$, we have
\begin{align}\label{eq:adals2}
    n_l^{(2)}=\floor*{\frac{t_{l}^{(2)}-T_{l-1}^{(1)}}{\abs{A_l}}}\ge \floor*{\frac{t_{l}^{(2)}-t_{l-1}^{(1)}/2}{\abs{A_l}}}\overset{\rm(i)}{\ge}\floor*{\frac{t_{l}^{(2)}/2}{\abs{A_l}}}\overset{\rm(ii)}{\ge}\frac{t_{l}^{(2)}}{4|A_l|}
\end{align}
for all $l\in[\eta_k-1]$, where (i) follows from \pref{eq:comparen1} and (ii) follows from $t_l^{(2)}\ge 4K$.

If $t_{\eta_k-1}^{(1)}/4\ge t_{\eta_k-1}^{(2)}$, then by $T_{\eta_k-1}^{(2)}= T_{\eta_k-2}^{(1)}+|A_{\eta_k-1}^{(2)}|n_{\eta_k-1}^{(2)}\le t_{\eta_k-1}^{(2)}$, we have
\begin{align}\label{eq:zzz}
  \floor*{\frac{{t}_{\eta_k-1}^{(1)}/2-T_{\eta_k-1}^{(2)}}{\abs{A_{\eta_k-1}^{(1)}}}} \ge \floor*{\frac{t_{\eta_k-1}^{(1)}/4}{|A_{\eta_k-1}^{(1)}|}}\ge\floor*{\frac{t_{\eta_k-1}^{(1)}}{4(K-\wh{r}(S,K))}}\ge\frac{t_{\eta_k-1}^{(2)}}{8(K-\wh{r}(S,K))},
\end{align}
where the last inequality follows from \[K-\wh{r}(S,K)\le \frac{1}{4}\floor{(K-\wh{r}(S,K))\sqrt{T/(K-\wh{r}(S,K))}}\le \frac{t_j^{(1)}}{4}, ~~~\forall j\in[q(S,K)+1].\]
By \pref{eq:comparen,eq:zz,eq:zzz}, we have
\begin{align*}
    \frac{N_k({T_{\eta_k}^{(1)}})}{\sqrt{N_{k}(T_{\eta_k-1}^{(1)})}}&\le\frac{N_{k}(T_{\eta_k-1}^{(1)})+\max\crl*{\floor*{\frac{{t}_{\eta_k}^{(1)}/2-T_{\eta_k}^{(2)}}{\abs{A_{\eta_k}^{(1)}}}},n_{\eta_k}^{(2)}}}{\sqrt{N_{k}(T_{\eta_k-1}^{(1)})}}\\
    &\le \sqrt{N_{k}(T_{\eta_k-1}^{(1)})}+\frac{1}{|A_{\eta_k^{(1)}}|}\frac{\max\crl*{{t_{\eta_k}^{(1)}}/2,{t_{\eta_k}^{(2)}}}}{\sqrt{N_{k}(T_{\eta_k-1}^{(1)})}}\\
    &\le \sqrt{N_{k}(T_{\eta_k-1}^{(1)})}+\frac{1}{|A_{\eta_k^{(1)}}|}{\max\crl*{\frac{t_{\eta_k}^{(1)}}{2}\sqrt{\frac{8(K-\wh{r}(S,K))}{t_{\eta_k-1}^{(1)}}},{t_{\eta_k}^{(2)}}\sqrt{\frac{K}{T_{\eta_k-1}^{(1)}}}}}\\
    &\le \sqrt{N_{k}(T_{\eta_k-1}^{(1)})}+\frac{1}{|A_{\eta_k}^{(1)}|}{\max\crl*{{t_{\eta_k}^{(1)}}\sqrt{\frac{2(K-\wh{r}(S,K))}{t_{\eta_k-1}^{(1)}}},{t_{\eta_k}^{(2)}}\sqrt{\frac{2K}{t_{\eta_k-1}^{(2)}}}}}\\
    &\le \sqrt{N_{k}(T_{\eta_k-1}^{(1)})}+\frac{2}{|A_{\eta_k}^{(1)}|}\max\crl*{\sqrt{\frac{K-\wh{r}(S,K)}{K-r(S,K)}}\cR_1,\cR_2}.
\end{align*}

If $t_{\eta_k-1}^{(1)}/4< t_{\eta_k-1}^{(2)}$, then $\prn*{\frac{T}{K-r(S,K)}}^{\frac{2-2^{2-\eta_k}}{2-2^{-q(S,K)}}}\le \frac{2t_{\eta_k-1}^{(1)}}{K-r(S,K)}\le\frac{8 t_{\eta_k-1}^{(2)}}{K-r(S,K)}\le\frac{8 K}{K-r(S,K)}\prn*{\frac{T}{K}}^{\frac{2-2^{2-\eta_k}}{2-2^{-q(S,K)-1}}}$, which implies $\prn*{\frac{T}{K-r(S,K)}}^{\frac{2-2^{1-\eta_k}}{2-2^{-q(S,K)}}}\le \prn{\frac{8K}{K-r(S,K)}}^{2}\prn*{\frac{T}{K}}^{\frac{2-2^{1-\eta_k}}{2-2^{-q(S,K)-1}}}$. Thus
\begin{equation}\label{eq:inq3}
    t_{\eta_k}^{(1)}\le (K-r(S,K))\prn*{\frac{T}{K-r(S,K)}}^{\frac{2-2^{1-\eta_k}}{2-2^{-q(S,K)}}}\le {\frac{64K}{K-r(S,k)}}K\prn*{\frac{T}{K}}^{\frac{2-2^{1-\eta_k}}{2-2^{-q(S,K)-1}}}.
\end{equation}
By \pref{eq:comparen,eq:zz}, we have $N_k(T_{\eta_k-1}^{(1)})\ge t_{\eta_k-1}^{(2)}/(2K)$, thus
\begin{align*}
    \frac{N_k({T_{\eta_k}^{(1)}})}{\sqrt{N_{k}(T_{\eta_k-1}^{(1)})}}&\le\frac{N_{k}(T_{\eta_k-1}^{(1)})+\max\crl*{\floor*{\frac{{t}_{\eta_k}^{(1)}/2-T_{\eta_k}^{(2)}}{\abs{A_{\eta_k}^{(1)}}}},n_{\eta_k}^{(2)}}}{\sqrt{N_{k}(T_{\eta_k-1}^{(1)})}}\\
    &\le \sqrt{N_{k}(T_{\eta_k-1}^{(1)})}+\frac{1}{|A_{\eta_k}^{(1)}|}\frac{\max\crl{t_{\eta_k}^{(1)}/2,{t_{\eta_k}^{(2)}}}}{\sqrt{N_{k}(T_{\eta_k-1}^{(1)})}}\\
    &\le \sqrt{N_{k}(T_{\eta_k-1}^{(1)})}+\frac{1}{|A_{\eta_k}^{(1)}|}{\max\crl{t_{\eta_k}^{(1)}/2,{t_{\eta_k}^{(2)}}}\sqrt{\frac{2K}{t_{\eta_k-1}^{(2)}}}}\\
    &\le \sqrt{N_{k}(T_{\eta_k-1}^{(1)})}+\frac{1}{|A_{\eta_k}^{(1)}|}\frac{64K}{K-{r}(S,K)}\cR_2.\tag{by \pref{eq:inq3}}
\end{align*}

We then consider the case of $k\in{A}_{1}^{(2)}$. By \pref{line:start} to \pref{line:end} of \pref{alg:new}, we know that action $k$ is chosen for $n_l^{(2)}$ rounds in each epoch $l\in[\eta_k]$. Thus
\begin{align*}
    \frac{N_k({T_{\eta_k}^{(1)}})}{\sqrt{N_{k}(T_{\eta_k-1}^{(1)})}}&\le\frac{N_{k}(T_{\eta_k-1}^{(1)})+n_{\eta_k}^{(2)}}{\sqrt{N_{k}(T_{\eta_k-1}^{(1)})}}\\
    &\le \sqrt{N_{k}(T_{\eta_k-1}^{(1)})}+\frac{n_{\eta_k}^{(2)}}{\sqrt{n_{\eta_k-1}^{(2)}}}\\
    &\le \sqrt{N_{k}(T_{\eta_k-1}^{(1)})}+\frac{t_{\eta_k}^{(2)}/|A_{\eta_k}|}{{\sqrt{n_{\eta_k-1}^{(2)}}}}\\
    &{\le} \sqrt{N_{k}(T_{\eta_k-1}^{(1)})}+\frac{t_{\eta_k}^{(2)}/|A_{\eta_k}|}{{\sqrt{{t_{\eta_k-1}^{(2)}/(4|A_{\eta_k-1}|)}}}}\tag{by \pref{eq:adals2}}\\
    &{\le} \sqrt{N_{k}(T_{\eta_k-1}^{(1)})}+\frac{2t_{\eta_k}^{(2)}/|A_{\eta_k}|}{{{\sqrt{t_{\eta_k-1}^{(2)}/K}}}}\\
    &\le \sqrt{N_{k}(T_{\eta_k-1}^{(1)})}+\frac{2\sqrt{2}}{\abs{A_{\eta_k}}}\cR_2.
\end{align*}

Combing the above three paragraphs, we prove \pref{eq:adals0}.

We then show \pref{eq:inq1} and \pref{eq:inq2}. Consider $k\in \wt{A}_{q(S,K)+1}^{(1)}$ If $t_{q(S,K)}^{(1)}/4\ge t_{q(S,K)}^{(2)}$, then \pref{eq:zzz} holds for $\eta_k=q(S,K)+1$; if $t_{q(S,K)}^{(1)}/4< t_{q(S,K)}^{(2)}$, then \pref{eq:inq3} holds for $\eta_k=q(S,K)+1$. Thus
\begin{align*}
{\sum_{k\in\wt{A}_{q(S,K)+1}^{(1)}}} \frac{N_k(T)}{\sqrt{N_k(T_{q(S,K)}^{(1)})}}&\le \max\crl*{T{\sqrt{\frac{8(K-\wh{r}(S,K))}{t_{q(S,K)}^{(1)}}}},{\frac{64K}{K-r(S,k)}}K\prn*{\frac{T}{K}}^{\frac{2-2^{-q(S,K)}}{2-2^{-q(S,K)-1}}}\sqrt{\frac{2K}{t_{q(S,K)}^{(2)}}}}\\
&\le {4}{\max\crl*{\sqrt{\frac{K-\wh{r}(S,K)}{K-r(S,K)}}\cR_1,\frac{32K}{K-r(S,K)}\cR_2}}.
\end{align*}
Consider $k\in \wt{A}_{q(S,K)+1}^{(2)}$. Since $N_k(T_{q(S,K)}^{(1)})\ge \frac{t_{q(S,K)}^{(2)}}{2K}$ and $N_k(T_{q(S,K)+1}^{(2)})\ge\frac{ t_{q(S,K)+1}^{(2)}}{4K}$, we have
\begin{align*}
    {\sum_{k\in\wt{A}_{q(S,K)+1}^{(2)}}}\frac{N_k(T_{q(S,K)+1}^{(2)})}{\sqrt{N_{k}(T_{q(S,K)}^{(1)})}}+\frac{N_k(T)}{\sqrt{N_{k}(T_{q(S,K)+1}^{(2)})}}&\le t_{q(S,K)+1}^{(2)}\sqrt{\frac{2K}{t_{q(S,K)}^{(2)
    }}}+T\sqrt{\frac{4K}{t_{q(S,K)+1}^{(2)}}}\\
    &\le 5\cR_2. \qed
\end{align*}

For any $k\in[K]$, define $R(T;k)\ldef\sum_{t=1}^{T}\prn*{\mu^*-\mu_{k}}\Ind\crl{a_t=k}=\Delta(k)N_k(T)$. Consider any $k$ such that $\eta_k\in[2:q(S,K)]$, by \pref{eq:upinqn}, conditional on the events $\cE$ and $\cE_1$, we always have
\begin{align*}
R(T;k)&=N_k(T)\Delta(k)\le 4 N_k({T_{\eta_k}^{(1)}})\sqrt{\frac{6\log T}{N_{k}(T_{\eta_k-1}^{(1)})}}\le 4\sqrt{6\log T}\frac{N_k({T_{\eta_k}^{(1)}})}{\sqrt{N_{k}(T_{\eta_k-1}^{(1)})}}
\end{align*}
Moreover, we have 
\[\scalebox{1}{$\underset{k:\eta_k=q(S,K)+1}{\sum} R(T;k)\le 4\sqrt{6\log T}\prn*{\underset{k\in\wt{A}_{q(S,K)+1}^{(1)}}{\sum} \frac{N_k(T)}{\sqrt{N_k(T_{q(S,K)}^{(1)})}}+\underset{k\in \wt{A}_{q(S,K)+1}^{(2)}}{\sum}\frac{N_k(T_{q(S,K)+1}^{(2)})}{\sqrt{N_{k}(T_{q(S,K)}^{(1)})}}+\frac{N_k(T)}{\sqrt{N_{k}(T_{q(S,K)+1}^{(2)})}}}$}\]
and $\sum_{k:\eta_k=1}R(T;k)\le T_1^{(1)}\le \max\crl{\cR_1/2,\cR_2}$. 

Therefore, for any $\cD$, conditional on the events $\cE$ and $\cE_1$, we have
\begin{align*}
T\mu^*-\sum_{t=1}^{T}\mu_{a_t}&=\sum_{k\in [K]}R(T;k)=\sum_{k:\eta_k=1}R(T;k) + \sum_{k:\eta_k\in[2:q(S,K)]}R(T;k) +\sum_{k:\eta_k=q(S,K)+1}R(T;k) \\
&\le \max\crl{\cR_1/2,\cR_2}+4\sqrt{6\log T}\prn*{\sum_{k:\eta_k\in[2:q(S,K)]}\frac{N_k({T_{\eta_k}^{(1)}})}{\sqrt{N_{k}(T_{\eta_k-1}^{(1)})}}}\\
&+4\sqrt{6\log T}\prn*{\sum_{k\in \wt{A}_{q(S,K)+1}^{(1)}} \frac{N_k(T)}{\sqrt{N_k(T_{q(S,K)}^{(1)})}}+\sum_{k\in \wt{A}_{q(S,K)+1}^{(2)}}\frac{N_k(T_{q(S,K)+1}^{(2)})}{\sqrt{N_{k}(T_{q(S,K)}^{(1)})}}+\frac{N_k(T)}{\sqrt{N_{k}(T_{q(S,K)+1}^{(2)})}}}.
\end{align*}
Combining the above inequality with \pref{lm:bounding} and  $\sum_{k:\eta_k>1}\sqrt{N_k(T_{\eta_k-1}^{(1)})}\le\sqrt{KT}$, we have
{\small
\begin{align}
\mathbb{E}_{\cD}^{\pi}\left[T\mu^*-\sum_{t=1}^{T}\mu_{a_t}\mid \mathcal{E},\cE_1\right]\le\cO(\log K\sqrt{ \log T})\cdot\max\crl*{\sqrt{\frac{K-\wh{r}(S,K)}{K-r(S,K)}}\cR_1,\ \frac{K}{K-{r}(S,K)}\cR_2}.\label{eq:case1c}
\end{align}
}

\textbf{Case 2: Both $\cE$ and $\cE_2$ occur.}  By the specification of \pref{alg:new}, we know that the optimal action $k^*\in A_l^{(2)}\subset A_l$ for all $l\in[q(S,K)+1]$, and $k^*\in A_{q(S,K)+2}$. Also, we have $k^*\in \wt{A}_l^{(2)}\subset \wt{A}_l$ for all $l\in[q(S,K)+1]$. Moreover, by \pref{line:start} to \pref{line:end} of \pref{alg:new}, we know that action $k^*$ is chosen for $n_l^{(2)}$
rounds in each epoch $l\in[q(S,K)]$, which is no greater than the number of plays of any other action chosen in epoch $l$. Therefore, for all $k\in[K]$ and $l\in[q(S,K)]$ we have
\begin{equation}\label{eq:choice2}
    N_{k^*}(T_l^{(1)})\le N_{k}(T_l^{(1)}),~~r_{k^*}(T_l^{(1)})\ge r_{k}(T_l^{(1)}).
\end{equation}

For any $k\in[K]$, define $\eta_k\ldef\max\crl*{j\in[q(S,K)+1]\mid k\in\wt{A}_j}$.

Consider any action $k$ such that $\eta_k>1$.  By \pref{line:elimination2} of \pref{alg:new},  the confidence intervals of the two actions $k^*$ and $k$ at the end of round $T_{\eta_k-1}^{(1)}$ must overlap, i.e., $\texttt{UCB}_{k}(T_{\eta_k-1}^{(1)})\ge\texttt{LCB}_{k^*}(T_{\eta_k-1}^{(1)})$. Therefore,
\begin{equation}\label{eq:upinqn3}
  \Delta(k):=\mu_{k^*}-\mu_k\le 2 r_{k^*}({T_{\eta_k-1}^{(1)}})+2r_{k}({T_{\eta_{k}-1}^{(1)}})\le 4r_{k^*}({T_{\eta_{k}-1}^{(1)}}),  
\end{equation}
where the last inequality follows from \pref{eq:choice2}. %
Since $k$ is never chosen after the $\eta_k$-th epoch, we have $N_{k}(T_{\eta_k}^{(1)})=N_k(T)$.

Consider any action $k$ such that action $k$ is chosen for more than $n_{\eta_k}^{(2)}$ rounds in epoch $\eta_k$. If $\eta_k<q(S,K)+1$,  by \pref{line:learn1,line:learn2} of \pref{alg:new},  the confidence intervals of the two actions $k^*$ and $k$ at the end of round $T_{\eta_k,i}^{(1)}$ must overlap, i.e., $\texttt{UCB}_{k}(T_{\eta_k,i}^{(1)})\ge\texttt{LCB}_{k^*}(T_{\eta_k,i}^{(1)})$. Therefore,
\begin{equation}\label{eq:upinqn2}
  \Delta(k):=\mu_{k^*}-\mu_k\le 2 r_{k^*}({T_{\eta_k,i}^{(1)}})+2r_{k}({T_{\eta_{k},i}^{(1)}})\le 4r_{k^*}({T_{\eta_{k}}^{(2)}}),  
\end{equation}
where the last equality follows from \pref{eq:choice2} and the specification of \pref{alg:new}. 
If $\eta_k=q(S,K)+1$, by \pref{line:last-elim} of \pref{alg:new}, the confidence intervals of the two actions $k^*$ and $k$ at the end of round $T_{q(S,K)+1}^{(2)}$ must overlap, i.e., $\texttt{UCB}_{k}(T_{q(S,K)+1}^{(2)})\ge\texttt{LCB}_{k^*}(T_{q(S,K)+1}^{(2)})$. Therefore,
\begin{equation}\label{eq:upinqn4}
  \Delta(k):=\mu_{k^*}-\mu_k\le 2 r_{k^*}({T_{q(S,K)+1}^{(2)}})+2r_{k}({T_{q(S,K)+1}^{(2)}})\le 4r_{k^*}({T_{q(S,K)+1}^{(2)}}),  
\end{equation}
where the last equality follows from \pref{eq:choice2} and the specification of \pref{alg:new}.

Using arguments similar to Case 1, we can establish the following lemma.

\begin{lemma}\label{lm:bounding2} Assume both $\cE$ and $\cE_2$ hold. For any action $k$ such that $\eta_k\in[2:q(S,K)]$, we have
\[\frac{N_k({T_{\eta_k}^{(2)}})}{\sqrt{N_{k^*}(T_{\eta_k-1}^{(1)})}}+\frac{N_k({T_{\eta_k}^{(1)}})}{\sqrt{N_{k^*}(T_{\eta_k}^{(2)})}}\le\begin{cases}
 \frac{({t_{\eta_k-1}^{(1)}}/2)/{\abs{A_{\eta_k}^{(1)}}}}{\sqrt{n_{\eta_k-1}^{(2)}}}+\frac{({t_{\eta_k}^{(1)}}/2)/{\abs{A_{\eta_k}^{(1)}}}}{\sqrt{n_{\eta_k}^{(2)}}}\le\frac{2\sqrt{2}}{|A_{\eta_k}^{(1)}|}\cR_2,&\text{if }k\in A_1^{(1)};\\
  \frac{{t_{\eta_k}^{(2)}}/{\abs{A_{\eta_k}}}}{\sqrt{n_{\eta_k-1}^{(2)}}}+\frac{{t_{\eta_k}^{(2)}}/{\abs{A_{\eta_k}}}}{\sqrt{n_{\eta_k}^{(2)}}}\le\frac{4\sqrt{2}}{|A_{\eta_k}|}\cR_2,&\text{if } k\in A_1^{(2)}.
\end{cases}
\]
Moreover, considering all $k$ such that $\eta_k=q(S,K)+1$ (i.e., $k\in\wt{A}_{q(S,K)+1}$), we have
\[\sum_{k\in\wt{A}_{q(S,K)+1}}\frac{N_k(T_{q(S,K)+1}^{(2)})}{\sqrt{N_{k^*}(T_{q(S,K)}^{(1)})}}+\frac{N_k(T)}{\sqrt{N_{k^*}(T_{q(S,K)+1}^{(2)})}}\le 4\sqrt{2}\cR_2.\]
\end{lemma}

For any $k\in[K]$, define $\cR(T;k)\ldef\sum_{t=1}^{T}\prn*{\mu^*-\mu_{k}}\Ind\crl{a_t=k}=\Delta(k)N_k(T)$. If $\eta_k>1$, by \pref{eq:upinqn3,eq:upinqn2,eq:upinqn4}, conditional on the events $\cE$ and $\cE_2$, we always have
\begin{align*}
R(T;k)&=N_k(T)\Delta(k)\le N_k(T_{\eta_k}^{(2)}) r_{k^*}(T_{\eta_k-1}^{(1)})+ \prn{N_k(T_{\eta_k}^{(1)})-N_k(T_{\eta_k^{(2)}})}r_{k^*}(T_{\eta_k}^{(2)})\\
&\le N_k({T_{\eta_k}^{(2)}})4\sqrt{\frac{6\log T}{N_{k^*}(T_{\eta_k-1}^{(1)})}}+  N_k({T_{\eta_k}^{(1)}})4\sqrt{\frac{6\log T}{N_{k^*}(T_{\eta_k}^{(2)})}}\\
&\le 4\sqrt{6\log T}\prn*{\frac{N_k({T_{\eta_k}^{(2)}})}{\sqrt{N_{k^*}(T_{\eta_k-1}^{(1)})}}+\frac{N_k({T_{\eta_k}^{(1)}})}{\sqrt{N_{k^*}(T_{\eta_k}^{(2)})}}}.
\end{align*}
Moreover, we have $\sum_{k:\eta_k=1}R(T;k)\le T_1^{(1)}\le \max\crl{\cR_1/2,\cR_2}$. 

Therefore, for any $\cD$, conditional on the events $\cE$ and $\cE_2$, we have
\begin{align*}
T\mu^*-\sum_{t=1}^{T}\mu_{a_t}&=\sum_{k\in [K]}R(T;k)=\sum_{k:\eta_k=1}R(T;k)+\sum_{k:\eta_k>1}R(T;k)\\&\le \max\crl{\cR_1/2,\cR_2}+  4\sqrt{6\log T}\prn*{\sum_{k:\eta_k>1} \frac{N_k({T_{\eta_k}^{(2)}})}{\sqrt{N_{k^*}(T_{\eta_k-1}^{(1)})}}+\frac{N_k({T_{\eta_k}^{(1)}})}{\sqrt{N_{k^*}(T_{\eta_k}^{(2)})}}}.
\end{align*}
Combining the above inequality with \pref{lm:bounding2}, we have
\begin{align}
\mathbb{E}_{\cD}^{\pi}\left[T\mu^*-\sum_{t=1}^{T}\mu_{a_t}\mid \mathcal{E},\cE_2\right]\le\cO(\log K \sqrt{\log T})\cR_2.\label{eq:case2c}
\end{align}

Combining \pref{eq:decomp,eq:case1c,eq:case2c},  we have
\begin{align}\label{eq:adals3}
R^{\pi}(T)\le\cO({\log K\sqrt{\log T}})\cdot\max\crl*{{\sqrt{\frac{K-\wh{r}(S,K)}{K-r(S,K)}}}^3\frac{K-r(S,K)}{K}\cR_1,\ \cR_2}.
\end{align}
Note that
\[
\frac{K-\wh{r}(S,K)}{K-r(S,K)}\le\frac{K-r(S,K)-1+q(S,K)}{K-r(S,K)}=1+\frac{q(S,K)-1}{K-r(S,K)}
\]
is $\cO(\log\log T)$ when $q(S,K)\le K\log_2\log_2 (T/K))$; and \pref{eq:adals3} is $\sqrt{KT}\cdot\cO(\sqrt{\log K \log T})$ when $q(S,K)\ge K\log_2\log_2 (T/K))$. This implies that \pref{eq:adals3} is always
\[
\cO((\log T)^2)\cdot\max\crl*{\frac{K-r(S,K)}{K}\cR_1,\ \cR_2}.
\]
Therefore, we finish the proof of \pref{thm:ub-n}. \hfill$\Box$

\section{Proof of \pref{thm:gub}}\label{app:proof-thm3}

Consider an arbitrary switching graph $G$ whose switching costs satisfy the triangle inequality. Recall that $H$ is the total weight of the shortest Hamiltonian path in $G$.

\subsection{The HS-SE Policy is Indeed an $S$-Switching-Budget Policy}

From round 1 to round $t_1$, \algog incurs $H$ switching cost.

For $1\le l\le q'(S,G)-1$, from round $t_l$ to round $t_{l+1}$, no matter whether $l$ is odd or even, no matter whether the last action in epoch $l$ is eliminated before the start of epoch $l+1$ or not, by the switching order (determined by the shortest Hamiltonian path of $G$) and the triangle inequality, \algog always incurs at most $H$ switching cost.

From round $t_{q'(S,G)}$ to round $T$, since \algog does not switch within epoch $q'(S,G)+1$, i.e., from round $t_{q'(S,G)}+1$ to round $T$, the only possible switch is between round $t_{q'(S,G)}$ and $t_{q'(S,G)}+1$. Thus \algog incurs at most $\max_{i,j\in[k]}c_{i,j}$ switching cost from round $t_{q'(S,G)}$ to round $T$.

Summarizing the above arguments, we find that \algog incurs at most $q'(S,G)H+\max_{i,j\in[k]}c_{i,j}\le S$ switching cost from round 1 to round $T$. Thus it is indeed an $S$-switching-budget policy.

\subsection{Proof of Upper Bound}
The proof is essentially the same as \pref{app:pub}, with $q(S,K)$ replaced by $q'(S,G)$. $\hfill\Box$

\section{Proof of the Upper Bound in \pref{thm:asym}}\label{app:ub-dbwsc}

Consider an arbitrary $\bm{c}\in\bbR_{\ge0}^K$. Recall that $i_K\in\arg\max_{i\in[K]}c_i$, $i_1\in\arg\max_{i\in[K]\setminus\crl{i_1}}c_i$, $c^{(1)}=\max_{i\in[K]}=c_{i_K}$, $c^{(2)}=\max_{i\ne i_K}c_i=c_{i_1}$, and $\Sigma=\sum_{i=1}^K c_i$.

\subsection{The AS-SE Policy is Indeed an $S$-Switching-Budget Policy}

From round 1 to round $T$, by the switching order specified in \pref{alg:asse}, \algoa departs from action $i_K$ for at most $\ceil*{\frac{q(S,\bm{c})}{2}}$ times, departs from action $i_1$ for at most $\floor*{\frac{q(S,\bm{c})}{2}}+1$ times, and departs from every other action for at most $q'(S,K)$ times. The total switching cost is no larger than
\begin{equation}\label{eq:magic}
\ceil*{\frac{q(S,\bm{c})}{2}}c^{(1)}+\prn*{\floor*{\frac{q(S,\bm{c})}{2}}+1}c^{(2)}+q(S,\bm{c})\sum_{i\in[K]\setminus\crl{i_1,i_K}}c_i.
\end{equation}
If $q(S,\bm{c})=\max\crl*{1+2\left\lfloor\frac{S-\Sigma}{2\Sigma-c^{(1)-c^{
(2)}}}\right\rfloor,2\floor*{\frac{S-c^{(2)}}{2\Sigma-c^{(1)}-c^{(2)}}}}=1+2\left\lfloor\frac{S-\Sigma}{2\Sigma-c^{(1)-c^{
(2)}}}\right\rfloor$, then \pref{eq:magic} equals to
\begin{align*}
    &~~~\prn*{1+\left\lfloor\frac{S-\Sigma}{2\Sigma-c^{(1)-c^{
(2)}}}\right\rfloor}\prn*{c^{(1)}+c^{(2)}}+\prn*{1+2\left\lfloor\frac{S-\Sigma}{2\Sigma-c^{(1)-c^{
(2)}}}\right\rfloor}\prn*{\Sigma-c^{(1)}-c^{(2)}}\\
&=\Sigma+\left\lfloor\frac{S-\Sigma}{2\Sigma-c^{(1)-c^{
(2)}}}\right\rfloor\prn*{2\Sigma-c^{(1)}-c^{(2)}}\\
&\le \Sigma+S-\Sigma=S.
\end{align*}
If $q(S,\bm{c})=\max\crl*{1+2\left\lfloor\frac{S-\Sigma}{2\Sigma-c^{(1)-c^{
(2)}}}\right\rfloor,2\floor*{\frac{S-c^{(2)}}{2\Sigma-c^{(1)}-c^{(2)}}}}=2\floor*{\frac{S-c^{(2)}}{2\Sigma-c^{(1)}-c^{(2)}}}$, then \pref{eq:magic} equals to
\begin{align*}
    &~~~\floor*{\frac{S-c^{(2)}}{2\Sigma-c^{(1)}-c^{(2)}}}\prn*{c^{(1)}+c^{(2)}}+c^{(2)}+2\floor*{\frac{S-c^{(2)}}{2\Sigma-c^{(1)}-c^{(2)}}}\prn*{\Sigma-c^{(1)}-c^{(2)}}\\
&=c^{(2)}+\left\lfloor\frac{S-c^{(2)}}{2\Sigma-c^{(1)-c^{
(2)}}}\right\rfloor\prn*{2\Sigma-c^{(1)}-c^{(2)}}\\
&\le c^{(2)}+S-c^{(2)}=S.
\end{align*}
Therefore, \algoa is indeed an $S$-switching-budget policy.

\subsection{Proof of Upper Bound}
The proof is essentially the same as \pref{app:pub}, with $q(S,K)$ replaced by $q(S,\bm{c})$. $\hfill\Box$

\vspace{2em}
\noindent{\large \bf Part III. Proof of Lower Bounds}

\section{Information-Theoretic Tools}\label{app:tools}
In this section, we introduce our information-theoretic tools.

For any two probability measures $\mathbb{P}$ and $\mathbb{Q}$ defined on the same measurable space $(\Omega,\cF)$, let $D_{\rm{TV}}(\mathbb{P}\|\mathbb{Q})\ldef\sup_{E\in\cF}\abs{\bbP(E)-\bbQ(E)}$ denote the total variation distance between $\mathbb{P}$ and $\mathbb{Q}$. We write $\bbP<<\bbQ$ to indicate that $\bbP$ is absolutely continuous with respect to $\bbQ$, and
\[D_{\rm{KL}}(\mathbb{P}\|\mathbb{Q})\ldef\begin{cases}
\int_{\Omega}\log\prn*{\frac{d\bbP}{d\bbQ}}d\bbP,&\text{if }\bbP<<\bbQ,\\
+\infty,&\text{otherwise.}
\end{cases}
\] 
be the \emph{Kullback-Leibler (KL) divergence} between $\mathbb{P}$ and $\mathbb{Q}$. Furthermore, let
\[
D_{\rm re}(\bbP\parallel \bbQ)\ldef D_{\rm KL}(\bbQ\parallel \bbP)
\]
be the \emph{reverse KL divergence} between $\bbP$ and $\bbQ$. For any $p,q\in[0,1]$, let
\[d_{\rm TV}(p\parallel q)\ldef D_{\rm TV}(\Ber (p)\parallel \Ber (q)),\]
\[d_{\rm KL}(p\parallel q)\ldef D_{\rm TV}(\Ber (p)\parallel \Ber (q))=p\log(\frac{p}{q})+(1-p)\log(\frac{1-p}{1-q}),\]
\[
d_{\rm re}(p\parallel q)\ldef D_{\rm re}(\Ber (p)\parallel \Ber (q))=q\log(\frac{q}{p})+(1-q)\log(\frac{1-q}{1-p}),
\]
where $\Ber(p)$ stands for the Bernoulli distribution with mean $p$. 
More generally, for any divergence denoted by $D(\cdot \parallel\cdot)$, let
\[
d(p\parallel q)\ldef D(\Ber (p)\parallel \Ber (q)).
\]

\subsection{Reverse Fano-type Inequalities}\label{app:fanoexp}
We first introduce a basic version of the reverse Fano-type inequality below.

\begin{proposition}[Reverse Fano-type Inequality]\label{prop:rf} Let $D$ be  the KL divergence or the reverse KL divergence. Let $\Prob_1,\dots,\Prob_N$ and $\bbQ$ be arbitrary probability measures on a common measurable space $(\Omega,\cF)$. For any measurable function $\psi:\Omega\mapsto[N]$, we have
\begin{equation}\label{eq:rf1}
     \frac{1}{N}\sum_{i=1}^N \Prob_i(\psi=i)\ge \frac{1}{N}-\frac{1}{N}\sqrt{2\prn*{1-\frac{1}{N}}\sum_{i=1}^N D(\Prob_i \parallel\bbQ)}.
\end{equation}
More generally, let $\bbQ_1,\dots,\bbQ_N$ be arbitrary probability measures on $(\Omega,\cF)$. For any sequence of events $E_1,\dots,E_N\in\cF$ (not necessarily disjoint), if $\wb{q}\ldef \frac{1}{N}\sum_{i=1}^N \bbQ_i(E_i)\in[0,\frac{1}{2}]$, then
\begin{equation}\label{eq:rf2}
    \frac{1}{N}\sum_{i=1}^N \Prob_i(E_i)\ge \wb{q}-\sqrt{2\wb{q}(1-\wb{q})\frac{1}{N}\sum_{i=1}^N D(\Prob_i\parallel \bbQ_i)}.
\end{equation}
\end{proposition}

To the best of our knowledge, both \pref{eq:rf1} and \pref{eq:rf2} are new. Note that the classical Fano's inequality \pref{eq:fano} provides a lower bound on the minimum ``average error probability''
\[\inf_{\psi}\frac{1}{N}\sum_{i=1}^N \Prob_i(\psi\ne i).\]
Our inequality \pref{eq:rf1} provides a sharp upper bound on the maximum ``average error probability''
\[
\sup_{\psi}\frac{1}{N}\sum_{i=1}^N \Prob_i(\psi\ne i),
\]
thus can be viewed as a reverse version of the classical Fano's inequality. Our inequality \pref{eq:rf2} further generalizes \pref{eq:rf1} to arbitrary events.

\textbf{Remark 1.} While there are some existing inequalities sometimes referred to as ``reverse Fano's inequalities'' in the literature (e.g., \citealt{chu1966inequalities,tebbe1968uncertainty}), they are very different from \pref{eq:rf1}, as all of them only provide an upper bound on 
$
\inf_{\psi}\frac{1}{N}\sum_{i=1}^N \Prob_i(\psi\ne i)
$
rather than
$
\sup_{\psi}\frac{1}{N}\sum_{i=1}^N \Prob_i(\psi\ne i)
$, i.e., their upper bound only holds for the minimax test  and does not hold for an \emph{arbitrary} test $\phi$. For this reason, we call the inequalities in \pref{prop:rf} ``reverse \textbf{Fano-type} inequalities'' to distinguish them from the existing ``reverse \textbf{Fano's} inequalities'' in the literature.

\textbf{Remark 2.} \cite{gerchinovitz2020fano} provide a very general framework to derive Fano-type inequalities, and their results imply two related inequalities. In the setting of \pref{eq:rf1}, their results in their Section 4.2 imply
\[
 \frac{1}{N}\sum_{i=1}^N \Prob_i(\psi=i)\ge \frac{1}{N}-\sqrt{{\frac{1}{N\log N}}\sum_{i=1}^N D(\Prob_i \parallel\bbQ)},
\]
which is worse than our \pref{eq:rf1} by a $\frac{1}{\sqrt{N}}$ factor --- importantly, this worse result cannot help us to obtain a tight lower bound for \UB whenever $K-r(S,K)=o(K)$. In the setting of \pref{eq:rf2}, \cite{gerchinovitz2020fano} show that
\[
 \frac{1}{N}\sum_{i=1}^N \Prob_i(E_i)\ge 1-{\frac{\frac{1}{N}\sum_{i=1}^N D_{\rm KL}(\Prob_i\parallel \bbQ)+\log 2}{\log\prn*{\frac{1}{1-\wb{q}}}}}.
\]
However, this bound becomes meaningless under our condition $\wb{q}\in[0,\frac{1}{2}]$. In other words, this bound is only useful for proving lower bounds for ``high-probability events'' rather than ``low-probability events'' --- the latter is required in the proof of \UB lower bounds.

\subsection{Generalized Reverse Fano-type Inequalities}

In this section, we introduce a more general version of the inequalities in \pref{prop:rf}, which enjoy several advantages as described in \pref{sec:methodology}. We then give the proof.

\begin{proposition}[Generalized Reverse Fano-type Inequality]\label{prop:grf}
Let $D$ be  the KL divergence or the reverse KL divergence.  Let $(\Omega_1,\cF_1), \dots, (\Omega_N,\cF_N)$ be an arbitrary sequence of measurable spaces. For any $i\in[N]$, let 
$\Prob_i$ and $\bbQ_i$ be arbitrary probability measures on  $(\Omega_i,\cF_i)$, and $E_i\in\cF_i$ be an arbitrary event. We have
\begin{equation}\label{eq:grf}
    \frac{1}{N}\sum_{i=1}^N \Prob_i(E_i)\ge \frac{1}{N}\sum_{i=1}^N \bbQ_i(E_i)-\sqrt{2\cdot{\frac{1}{N}\sum_{i=1}^N \bbQ_i(E_i)}\cdot\frac{1}{N}\sum_{i=1}^N D(\Prob_i\parallel \bbQ_i)}.
\end{equation}
Moreover, if $\wb{q}\ldef \frac{1}{N}\sum_{i=1}^N \bbQ_i(E_i)\in[0,\frac{1}{2}]$, then  we have a slightly tighter bound
\begin{equation}\label{eq:grf2}
    \frac{1}{N}\sum_{i=1}^N \Prob_i(E_i)\ge \wb{q}-\sqrt{2\wb{q}(1-\wb{q})\cdot{}\cdot\frac{1}{N}\sum_{i=1}^N D(\Prob_i\parallel \bbQ_i)}.
\end{equation}
\end{proposition}

\proof{Proof of \pref{prop:grf}.} 
Our proof builds on a two-step procedure established by \cite{gerchinovitz2020fano}, with a few key modifications in the second step to obtain sharper one-sided inequalities.

Our first step is a reduction to Bernoulli distributions. By the joint convexity of general $f$-divergences, we have
\[
d\prn*{ \frac{1}{N}\sum_{i=1}^N \Prob_i(E_i) \parallel \frac{1}{N}\sum_{i=1}^N \bbQ_i(E_i)}\le \frac{1}{N}\sum_{i=1}^Nd\prn*{\bbP_i(E_i)\parallel\bbQ_i(E_i)}=\frac{1}{N}\sum_{i=1}^Nd\prn*{\bbP_i(E_i)\parallel\bbQ_i(E_i)}.
\]
Note that the above inequality holds even if $\bbP_1,\dots,\bbP_N$ are on different measurable spaces. 
For all $i\in[N]$, since $\Ber(\bbP_i(E_i))$ (resp., $\Ber(\bbQ_i(E_i))$) is the law of $\Ind_{E_i}$ under $\bbP_i$ (resp., $\bbQ_i$), using the data-processing inequality for $f$-divergences (see, e.g., Lemma 1 in \citealt{gerchinovitz2020fano}), we have
\[
d\prn*{\bbP_i(E_i)\parallel\bbQ_i(E_i)}=D\prn*{\Ber(\bbP_i(E_i))\parallel\Ber(\bbQ_i(E_i))}\le D\prn*{\bbP_i'\parallel\bbQ_i'}.
\]
Thus we have
\[
d\prn*{ \frac{1}{N}\sum_{i=1}^N \Prob_i(E_i) \parallel \frac{1}{N}\sum_{i=1}^N \bbQ_i(E_i)}\le\frac{1}{N}\sum_{i=1}^ND\prn*{\bbP_i\parallel\bbQ_i}.
\]
Let $\wb{p}\ldef \frac{1}{N}\sum_{i=1}^N \Prob_i(A_i)$, we have
\begin{equation}\label{eq:bern}
d_f\prn*{ \wb{p} \parallel \wb{q}}\le\frac{1}{N}\sum_{i=1}^ND\prn*{\bbP_i\parallel\bbQ_i}.
\end{equation}

In the second step, we lower bound $d\prn*{ \wb{p} \parallel \wb{q}}$ to extract a lower bound on $\wb{p}$. When $D$ is restricted to be the KL divergence or the reverse KL divergence,  \pref{lm:pinsker} and \pref{lm:pinsker2},  we have
\[
d\prn*{ \wb{p} \parallel \wb{q}}\ge \frac{\prn*{\wb{p}-\wb{q}}^2}{2\wb{q}}
\]
for all $q\in[0,1)$ and 
\[
d\prn*{ \wb{p} \parallel \wb{q}}\ge \frac{\prn*{\wb{p}-\wb{q}}^2}{2\wb{q}(1-\wb{q})}\]
for all $q\in[0,\frac{1}{2}]$. Note that  \pref{lm:pinsker} and \pref{lm:pinsker2} provide ``localized'' versions of the Pinsker's inequality  that substantially improves over existing ``global'' variants of the Pinsker's inequality by exploiting the one-sided condition $\wb{p}\le\wb{q}$ (see the remarks after \pref{lm:pinsker} and \pref{lm:pinsker2}). Such improvement is critical for our second step and enables us to obtain tight one-sided inequalities about $\wb{p}$ (with improved dependence on $\wb{q}$), which we describe below:
\begin{itemize}
    \item If $\wb{p}\notin[0,\wb{q}]$, then $\wb{p}\ge\wb{q}$.
    \item If $\wb{p}\in[0,\wb{q}]$, then we have
    \[
    d\prn*{ \wb{p} \parallel \wb{q}}\ge \frac{\prn*{\wb{p}-\wb{q}}^2}{2\wb{q}},
    \]
    which implies $ \wb{p}\ge\wb{q}-\sqrt{2\wb{q}d(\wb{p}\parallel \wb{q})}$.
\end{itemize}
Therefore, no matter $\wb{p}\in [0,\wb{q}]$ or not, we always have
\begin{equation}\label{eq:one-sided}
        \wb{p}\ge\wb{q}-\sqrt{2\wb{q}d_f(\wb{p}\parallel \wb{q})}.
    \end{equation}
The above inequality can be improved to $\wb{p}\ge\wb{q}-\sqrt{2\wb{q}(1-\wb{q})d(\wb{p}\parallel \wb{q})}$ when $\wb{q}\in[0,\frac{1}{2}]$.
    
 By \pref{eq:bern}, we prove \pref{eq:grf}. $\hfill\Box$

\subsection{Localized Pinsker's Inequalities}

\begin{lemma}[Localized Pinsker's Inequality]\label{lm:pinsker}
If $0\le p\le q\le\frac{1}{2}$ or $\frac{1}{2}\le q\le p\le 1$, then
\[
d_{\rm KL}(p\parallel q)\ge\frac{(p-q)^2}{2q(1-q)}~~~~  \text{and}~~~~
d_{\rm KL}(q\parallel p)\ge\frac{(p-q)^2}{2q(1-q)}.
\]
\end{lemma}
\proof{Proof of \pref{lm:pinsker}.} If $p=q=0$ or $p=q=1$, then $d_{\rm KL}(p\parallel q)=d_{\rm KL}(q\parallel p)=0=\frac{(p-q)^2}{2q(1-q)}$. In the rest of the proof, we fix $q\in(0,1)$.

We first define
\[g(x)\ldef{\rm kl}(x,q)-\frac{(x-q)^2}{2q(1-q)}= x\log\frac{x}{q}+(1-x)\log\frac{1-x}{1-q}-\frac{(x-q)^2}{2q(1-q)}\]
for all $x\in[0,1]$. 
We have 
\[g'(x)=\log\prn*{\frac{x}{1-x}\frac{1-q}{q}}-\frac{x-q}{q(1-q)},
\]
\[
g''(x)=\frac{1}{x(1-x)}-\frac{1}{q(1-q)}.
\]
We discuss two cases:
\begin{itemize}
    \item If $q\le\frac{1}{2}$, then $g''(x)\ge0$ for all $x\in(0,q]$. Furthermore, since $g'(q)=0$, we have $g'(x)\le0$ for all $x\in(0,q)$, which implies that $g(x)\ge g(q)=0$ for all $x\in[0,q]$. Thus $g(p)={\rm kl}(p,q)-\frac{(p-q)^2}{2q(1-q)}\ge0$ when $0\le p\le q\le \frac{1}{2}$.
    \item  If $q\ge\frac{1}{2}$, then $g''(x)\ge0$ for all $x\in[q,1)$. Furthermore, since $g'(q)=0$, we have $g'(x)\ge0$ for all $x\in(q,1)$, which implies that $g(x)\ge g(q)=0$ for all $x\in[q,1]$. Thus $g(p)={\rm kl}(p,q)-\frac{(p-q)^2}{2q(1-q)}\ge0$ when $\frac{1}{2}\le q\le p\le 1$.
\end{itemize}

We then define
\[g(x)\ldef{\rm kl}(q,x)-\frac{(x-q)^2}{2q(1-q)}= q\log\frac{q}{x}+(1-q)\log\frac{1-q}{1-x}-\frac{(x-q)^2}{2q(1-q)}\]
for all $x\in(0,1)$. We have
\[
g'(x)=\frac{x-q}{x(1-x)}-\frac{x-q}{q(1-q)}=\frac{(x+q-1)(x-q)^2}{x(1-x)q(1-q)}.
\]
We discuss two cases:
\begin{itemize}
\item If $q\le\frac{1}{2}$, then $x+q-1\le0$ and $g'(x)\le0$ for all $x\in(0,q]$, which implies that $g(x)\ge g(q)=0$ for all $x\in(0,q]$. Thus $g(p)={\rm kl}(q,p)-\frac{(p-q)^2}{q(1-q)}\ge0$ when $0\le p\le q\le\frac{1}{2}$.
\item If $q\ge\frac{1}{2}$, then $x+q-1\ge0$ and $g'(x)\ge0$ for all $x\in[q,1)$, which implies that $g(x)\ge g(q)=0$ for all $x\in(q,1]$. Thus $g(p)={\rm kl}(q,p)-\frac{(p-q)^2}{q(1-q)}\ge0$ when $\frac{1}{2}\le q\le p\le 1$.
\end{itemize}

To sum up, if $0\le p\le q\le\frac{1}{2}$ or $\frac{1}{2}\le q\le p\le 1$, then ${\rm kl}(p,q)\ge\frac{(p-q)^2}{2q(1-q)}$ and ${\rm kl}(q,p)\ge\frac{(p-q)^2}{2q(1-q)}$. $\hfill\Box$

\begin{lemma}[Localized Pinsker's Inequality, version 2]\label{lm:pinsker2}
If $0\le p\le q\le1$, then
\[
d_{\rm KL}(p\parallel q)\ge\frac{(p-q)^2}{2q}~~~~  \text{and}~~~~
d_{\rm KL}(q\parallel p)\ge\frac{(p-q)^2}{2q}.
\]
\end{lemma}

\pref{lm:pinsker2} is a corollary of Lemma A.2 in \cite{talebi2017minimizing}. It has a slightly worse constant compared with \pref{lm:pinsker}, but holds for a more general range of $\wb{q}$.

\textbf{Remark.} The classical Pinsker's inequality (see, e.g., Lemma 2.5 in \citealt{tsybakov2008introduction}) for Bernoulli distributions states
that 
\[
d_{\rm KL}(p \parallel q)\ge 2 \prn*{d_{\rm TV}(p \parallel q)}^2=2(p-q)^2
\]
for all $p,q\in[0,1]$. Many improvements and generalizations of the Pinsker's inequality have been obtained in the literature, including the ``refined Pinsker's inequality'' by \cite{ordentlich2005distribution}, which states that
\[
d_{\rm KL}(p \parallel q)\ge \frac{\log((1-q)/q)}{1-2q}(p-q)^2
\]
for  all $p,q\in[0,1]$. The above bounds become substantially weaker than the $\frac{(p-q)^2}{2q(1-q)}$ bound in \pref{lm:pinsker} and the $\frac{(p-q)^2}{2q}$ bound in \pref{lm:pinsker2} as $q$ gets closer to 0, i.e., they lose an $\wt{O}(1/q)$ factor when $q\rightarrow0$. In fact, all variants of the Pinsker's inequality that seek to establish a \emph{global} bound which holds for all $p,q\in[0,1]$ must lose such a huge
factor compared with \pref{lm:pinsker,lm:pinsker2}, thus are loose for our purpose (note that \pref{lm:pinsker,lm:pinsker2} critically utilizes the \emph{one-sided} condition: $p\le q$).  It is also worth noting that \pref{lm:pinsker,lm:pinsker2} hold for not only the KL divergence $d_{\rm KL}(p\parallel q)$ but also the \emph{reverse} KL divergence $d_{\rm re}(p\parallel q)$. This feasture is crucial for us to prove \pref{prop:grf} and  establish tight lower bounds on the regret of the \BwSC problem.

\section{Proof of \pref{thm:lb-unit}}\label{app:lower}

For an overview of the proof, see \pref{sec:sketch}.

Given any $K>1$, $S\ge0$ and $T\ge 2K$, we focus on the setting of $\cD_k=\mathcal{N}(\mu_k,1)$ ($\forall k\in[K]$), as this is sufficient for us to prove the desired lower bound. Note that now the underlying environment (i.e., latent distributions) $\cD$ can be completely determined by a vector $\bmu=(\mu_1,\cdots,\mu_K)\in\mathbb{R}^{K}$. For simplicity, in this proof we will directly use the vector $\bmu$ to represent the environment.

For any environment $\bmu$, let $X_{\bmu}^{t}(k)\sim\mathcal{N}(\mu_k,1)$ denote the i.i.d. random reward of each action $k$ at round $t$ ($k\in[K],t\in[T]$). 
For any policy $\pi\in\Pi_S$, for any environment $\bmu$, for any $t\in[T]$, we use $a_t$ to denote the random action selected by policy $\pi$ at round $t$ under environment $\bmu$, and use $X^t_{\bmu}(a_t)$ to denote the random reward observed by policy $\pi$ at round $t$ under environment $\bmu$. Let $\cH_t\ldef\prn*{\prn*{a_1,X_{\bmu}^1(a_1)},\dots,\prn*{a_t,X_{\bmu}^t(a_t)}}$ be the history (of actions and observations) up to round $t$ (inclusive), whose value lies in $\Omega_t\ldef\prn*{[K]\times\bbR}^t$. Let $\cF_t\ldef\cB(\Omega_t)$ be the Borel $\sigma$-algebra on $\Omega_t$. Let $\bbP_{\bmu}^{\pi}$ be the probability measure induced by (i.e., the joint distribution of) $\cH_T$, and $\bbE_{\bmu}^\pi$ be the associated expectation operator. Let $R_{\bmu}^{\pi}(T)\ldef T\mu^*-\bbE_{\bmu}^{\pi}\brk*{\sum_{t=1}^T{\mu}_{a_t}}$ be policy $\pi$'s distribution-dependent regret under environment $\bmu$.

We argue that in our proof,  we only need to consider the case of $q(S,K)+2\le \log_2\log_2(T/K)$. 
Suppose $q(S,K)+2>\log_2\log_2(T/K)$, then we have 
\begin{align*}
K^{1-\frac{1}{2-2^{-q(S,K)-1}}}T^{\frac{1}{2-2^{-q(S,K)-1}}}&=K(T/K)^{\frac{1}{2-2^{-q(S,K)-1}}}\\
&=K(T/K)^\frac{1}{2}(T/K)^\frac{2^{-q(S,K)-2}}{2-2^{-q(S,K)-1}}\\
&\le \sqrt{KT}(T/K)^{2^{-q(S,K)-2}}\\
&<\sqrt{KT}(T/K)^{2^{-\log_2\log_2(T/K)}}\\
&=\sqrt{KT}(T/K)^{{\log_{T/K}(2)}}=\sqrt{2KT},
\end{align*}
thus the lower bound in \pref{thm:lb-unit} becomes $\Omega(\sqrt{KT}/\log T)$ and can be directly obtained by applying the well-known $\Omega(\sqrt{KT})$ lower bound of the classical \MAB (see, e.g., \citealt{lattimore2020bandit}). Therefore, the really non-trivial case of \pref{thm:lb-unit} is the case of $q(S,K)+2\le \log_2\log_2(T/K)$, and we focus on this case in the rest of our proof.

Our goal is to explicitly construct a family of environments $\Phi$, such that for any $S$-switching-budget policy $\pi\in\Pi_S$, the ``average-case regret'' $\frac{1}{|\Phi|}\sum_{\bmu\in\Phi}R_{\bmu}^{\pi}(T)$ is lower bounded by both 
\begin{equation}\label{eq:term1}
    \wt{\Omega}\prn*{\frac{(K-r(S,K))^{2-\frac{1}{2-2^{-q(S,K)}}}}{K}T^{\frac{1}{2-2^{-q(S,K)}}}}
\end{equation}
and
\begin{equation}\label{eq:term2}
    \wt{\Omega}\prn*{K^{1-\frac{1}{2-2^{-q(S,K)-1}}}T^{\frac{1}{2-2^{-q(S,K)-1}}}}.
\end{equation}
Since the worst-case regret $R^{\pi}(T)$ is no less than the ``average-case regret'' $\frac{1}{|\Phi|}\sum_{\bmu\in\Phi}R_{\bmu}^{\pi}(T)$, the above goal directly implies \pref{thm:lb-unit}. 
In our proof, we construct two classes of environments $\Phi_{1}$ and $\Phi_2$ to show the lower bounds \pref{eq:term1} and \pref{eq:term2} respectively.

Our lower bound proof program consists of five steps:
\begin{enumerate}
    \item \textbf{R}isky \textbf{E}vents
    \item \textbf{C}ombinatorial arguments and lower bounds under a single environment
    \item \textbf{A}lternative environments, bad events, and lower bound reductions
    \item \textbf{P}robability space changing tricks
    \item Applying the \GRF inequality %
\end{enumerate}
Based on the initials of the first four steps, we call the program \lbargu.  We present the five steps in the five subsections below.

\subsection{Definitions of Risky Events}\label{app:REdef}

For any policy $\pi\in\Pi_S$, for any environment ${\bmu}$, we make some key definitions below. 

1. For any $n_1,n_2\in[T]$, we define a random variable $S(n_1,n_2)$ to be the  total switching cost incurred in period $[n_1:n_2]$ (note that if there is a switch happening between round $n_1-1$ and round $n_1$, or between round $n_2$ and round $n_2+1$, we do not count its cost in $S(n_1,n_2)$).

2. Second, we define a stopping time 
\[
\tau\ldef\min\crl*{t\in[T]: S\prn*{1:t}=S}
\]
if the set is non-empty and $\tau=\infty$ otherwise. That is, $\tau$ is the first round that the learner's total switching cost reaches $S$.

3. We define a class of \emph{risky} events as follows: for any $k\in[K]$, let

$E_{1,k}^{(1)}\ldef\crl*{\text{action }k \text{ is not chosen in period }\brk*{1:t_{1}^{(1)}}},$

$E_{j,k}^{(1)}\ldef\crl*{\text{action }k \text{ is not chosen in period }\brk*{t_{j-1}^{(1)}:t_{j}^{(1)}}},~~\forall j\in[2:q(S,K)],$

$E_{q(S,K)+1,k}^{(1)}\ldef\crl*{\text{action }k \text{ is not chosen in period }\brk*{t_{q(S,K)}^{(1)}:\floor{\prn{t_{q(S,K)}^{(1)}+T}/2}}},$

$E_{q(S,K)+2,k}^{(1)}\ldef\crl*{\tau\le \floor{\prn{t_{q(S,K)}^{(1)}+T}/2},\, a_\tau=k, \text{ action }k\text{ is not chosen in period }\brk*{t_{q(S,K)}^{(1)}:\tau-1}}.$\\
By doing so, we get $\prn*{q(S,K)+2}K$ risky events (of the form $E_{j,k}^{(1)}$) in total. Note that the time points $\prn{t_j^{(1)}}_{j=1}^{q(S,K)+1}$ are fixed and given in \pref{alg:new}, and the events $\prn{E_{q(S,K)+2,k}^{(1)}}_{k\in[K]}$ are defined based on the stopping time $\tau$. We refer to the above class of risky events as ``{the first class of risky events},'' and will use them to prove the lower bound \pref{eq:term1}.

4. We  then define another class of risky events: for any $k\in[K]$, let

$E_{1,k}^{(2)}\ldef\crl*{\text{action }k \text{ is not chosen in period }\brk*{1:t_{1}^{(2)}}},$

$E_{j,k}^{(2)}\ldef\crl*{\text{action }k \text{ is not chosen in period }\brk*{t_{j-1}^{(2)}:t_{j}^{(2)}}},~~\forall j\in[2:q(S,K)+2].$
\\
By doing so, we get $\prn*{q(S,K)+2}K$ risky events (of the form $E_{j,k}^{(2)}$) in total. Note that the time points $\prn{t_j^{(2)}}_{j=1}^{q(S,K)+1}$ are fixed and given in \pref{alg:new}. We refer to the above class of risky events as ``{the second class of risky events},'' and will use them to prove the lower bound \pref{eq:term2}.

\textbf{Remark.} Note that in the first class of risky events, the events $\prn{E_{q(S,K)+2,k}^{(1)}}_{k\in[K]}$ are defined based on the stopping time $\tau$. Such delicate design is crucial for our analysis --- the importance should become clear quite soon. In particular, if we do not define the events $\prn{E_{q(S,K)+2,k}^{(1)}}_{k\in[K]}$ in this step, but only define the events  $\prn{E_{q(S,K)+2,k}^{(1)}}_{j\in[q(S,K)+1],k\in[K]}$ (which do not involve $\tau$), then in the next step, we can only show
\[
\sum_{j\in[q(S,K)+1]}\sum_{k\in[K]}\bbP_{\bmu}^{\pi}\prn*{E_{j,k}^{(1)}}\ge K-1-\floor*{\frac{S}{q(S,K)+1}}
\]
for \pref{lm:first-class}, which is unfortunately a meaningless result when $S\%(K-1)=0$ (i.e., when $S$ is at the end of a ``phase'' defined in \pref{sec:phase1}), as the right-hand side $K-1-\floor*{\frac{S}{q(S,K)+1}}$ becomes 0 when $S\%(K-1)=0$ --- this issue will eventually prevent us from making the floor function $\floor*{\frac{S-1}{K-1}}$ appear in the lower bound. By contrast, by defining the events $\prn{E_{q(S,K)+2,k}^{(1)}}_{k\in[K]}$ based on $\tau$, we are able to show
\[
\sum_{j\in[q(S,K)+2]}\sum_{k\in[K]}\bbP_{\bmu}^{\pi}\prn*{E_{j,k}^{(1)}}\ge K-\ceil*{\frac{S}{q(S,K)+1}}
\]
for \pref{lm:first-class}. Importantly, the right-hand side is always no less than 1 --- this property will play a fundamental role in  subsequent analysis. %

\subsection{Combinatorial Arguments and Lower Bounds for Risky Events (under a Single Environment)}
The main purpose of this subsection is to prove the following two  lemmas (\cref{lm:first-class} and \cref{lm:second-class}) using (non-trivial) combinatorial (and probabilistic) arguments. The arguments extensively exploit the properties of the switching constraint.

\begin{lemma}\label{lm:first-class}For any policy $\pi\in\Pi_S$, for any environment $\bmu$, we have
\[
\sum_{j\in[q(S,K)+2]}\sum_{k\in[K]}\bbP_{\bmu}^{\pi}\prn*{E_{j,k}^{(1)}}\ge K-\ceil*{\frac{S}{q(S,K)+1}}.
\]
\end{lemma}

\begin{lemma}\label{lm:second-class}For any policy $\pi\in\Pi_S$, for any environment $\bmu$, we have
\[
\sum_{j\in[q(S,K)+2]}\sum_{k\in[K]}\bbP_{\bmu}^{\pi}\prn*{E_{j,k}^{(2)}}\ge K-1-\floor*{\frac{S}{q(S,K)+2}}.
\]
\end{lemma}

\cref{lm:first-class} and \cref{lm:second-class} lead to the following corollary, which will be utilized in subsequent subsections.

\begin{corollary}
\label{cor:first-second-class}For any policy $\pi\in\Pi_S$, for any environment $\bmu$, we have
\[
\frac{1}{(q(S,K)+2)K}\sum_{j\in[q(S,K)+2]}\sum_{k\in[K]}\bbP_{\bmu}^{\pi}\prn*{E_{j,k}^{(1)}}\ge \frac{K-r(S,K)}{2(q(S,K)+2)^2K},
\]
\[
\frac{1}{(q(S,K)+2)K}\sum_{j\in[q(S,K)+2]}\sum_{k\in[K]}\bbP_{\bmu}^{\pi}\prn*{E_{j,k}^{(2)}}\ge \frac{1}{2(q(S,K)+2)^2}.
\]
\end{corollary}

\cref{cor:first-second-class} tells us the following fact: under any \emph{single} environment $\bmu$, the average probability of the first class of risky events is $\wt{\Omega}\prn*{\frac{K-r(S,K)}{K}}$, and the average probability of the second class of risky events is $\wt{\Omega}(1)$.

In the rest of this subsection, we provide proofs for \cref{lm:first-class}, \cref{lm:second-class} and \cref{cor:first-second-class}.

\proof{Proof of \pref{lm:first-class}.} For any $j\in[q(S,K)]$, we have
\begin{align*}
\sum_{k\in[K]}\Ind\crl*{E_{j,k}^{(1)}}&=\text{number of actions that are not chosen in period }\brk*{t_{j-1}^{(1)}:t_{j}^{(1)}}\\
&\ge K-1-S\prn*{t_{j-1}^{(1)}:t_j^{(1)}}\\
&\ge \prn*{K-\ceil*{\frac{S}{q(S,K)+1}}}\Ind\crl*{S\prn*{t_{j-1}^{(1)}:t_j^{(1)}}<\frac{S}{q(S,K)+1}}
\end{align*}
almost surely. Thus for any $j\in[q(S,K)]$, we have
\begin{equation}
\begin{aligned}[b]\label{eq:lm6}
\sum_{k\in[K]}\bbP_{\bmu}^{\pi}\prn*{E_{j,k}^{(1)}}& = \sum_{k\in[K]}\bbE_{\bmu}^{\pi}\brk*{\Ind\crl*{E_{j,k}^{(1)}}}\\
& =  \bbE_{\bmu}^{\pi}\brk*{\sum_{k\in[K]}\Ind\crl*{E_{j,k}^{(1)}}}\\
& \ge \bbE_{\bmu}^{\pi}\brk*{\prn*{K-\ceil*{\frac{S}{q(S,K)+1}}}\Ind\crl*{S\prn*{t_{j-1}^{(1)}:t_j^{(1)}}<\frac{S}{q(S,K)+1}}}\\
& =\prn*{K-\ceil*{\frac{S}{q(S,K)+1}}}\bbE_{\bmu}^{\pi}\brk*{\Ind\crl*{S\prn*{t_{j-1}^{(1)}:t_j^{(1)}}<\frac{S}{q(S,K)+1}}}.
\end{aligned}
\end{equation}
Summing \pref{eq:lm6} over $j\in[q(S,K)]$, we have
\begin{align*}
\sum_{j\in[q(S,K)]}\sum_{k\in[K]}\bbP_{\bmu}^{\pi}\prn*{E_{j,k}^{(1)}}& \ge \prn*{K-\ceil*{\frac{S}{q(S,K)+1}}}\bbE_{\bmu}^\pi\brk*{ \sum_{j\in[q(S,K)]}\Ind\crl*{S\prn*{t_{j-1}^{(1)}:t_j^{(1)}}<\frac{S}{q(S,K)+1}}}\\
&\overset{\rm(i)}{\ge}\prn*{K-\ceil*{\frac{S}{q(S,K)+1}}}\bbE_{\bmu}^\pi\brk*{\prn*{1- \Ind\crl*{S\prn*{1:t_{q(S,K)}^{(1)}}\ge \frac{q(S,K)}{q(S,K)+1}S}}}.
\end{align*}
Note that (i) follows from
\begin{align*}
\sum_{j\in[q(S,K)]}\Ind\crl*{S\prn*{t_{j-1}^{(1)}:t_j^{(1)}}<\frac{S}{q(S,K)+1}}&\ge\Ind\crl*{\bigcup_{j\in[q(S,K)]}\crl*{S\prn*{t_{j-1}^{(1)}:t_j^{(1)}}<\frac{S}{q(S,K)+1}}}\\
&\overset{\rm(ii)}{\ge}\Ind\crl*{S\prn*{1:t_{q(S,K)}^{(1)}}< \frac{q(S,K)}{q(S,K)+1}S}\\
&=1-\Ind\crl*{S\prn*{1:t_{q(S,K)}^{(1)}}\ge \frac{q(S,K)}{q(S,K)+1}S},
\end{align*}
where (ii) follows from the pigeonhole principle.

Now we define
\begin{align*}
E_{\sim,k}^{(1)} \ldef   \crl*{\text{action }k\text{ is not among the first }\ceil*{\frac{S}{q(S,K)+1}}\text{ (different) actions chosen in period }\brk*{t_{q(S,K)}^{(1)}:T}}.
\end{align*}
If $\crl*{S\prn*{1:t_{q(S,K)}^{(1)}}\ge \frac{q(S,K)}{q(S,K)+1}S}$ happens, then by the switching constraint $S(1:T)\le S$, we have $S\prn*{t_{q(S,K)}^{(1)}:T}\le\frac{S}{q(S,K)+1}$. If we further assume $ E_{\sim,k}^{(1)} $ happens, then by $S\prn*{t_{q(S,K)}^{(1)}:T}\le\frac{S}{q(S,K)+1}$, 
either $E_{q(S,K)+1,k}^{(1)}=\crl*{\text{action }k \text{ is not chosen in period }\brk*{t_{q(S,K)}^{(1)}:\floor{\prn{t_{q(S,K)}^{(1)}+T}/2}}}$ happens, or $E_{q(S,K)+2,k}^{(1)}=\crl*{\tau\le \floor{\prn{t_{q(S,K)}^{(1)}+T}/2},\, a_\tau=k, \text{ action }k\text{ is not chosen in period }\brk*{t_{q(S,K)}^{(1)}:\tau-1}}$ happens.
Therefore, we know that
\[
E_{q(S,K)+1,k}^{(1)}\cup E_{q(S,K)+2,k}^{(1)} \ \supset\   E_{\sim,k}^{(1)} \cap \crl*{S\prn*{1:t_{q(S,K)}^{(1)}}\ge \frac{q(S,K)}{q(S,K)+1}S}
\]
This implies that
\begin{align*}
    \sum_{k\in[K]}\bbP_{\bmu}^{\pi}\prn*{E_{q(S,K)+1,k}^{(1)}\cup E_{q(S,K)+2,k}^{(1)} }& \ge\sum_{k\in[K]}\bbP_{\bmu}^{\pi}\prn*{E_{\sim,k}^{(1)} \cap \crl*{S\prn*{1:t_{q(S,K)}^{(1)}}\ge \frac{q(S,K)}{q(S,K)+1}S}}\\
    &= \sum_{k\in[K]}\bbE_{\bmu}^\pi\brk*{\Ind\crl*{E_{\sim,k}^{(1)}}\Ind{\crl*{S\prn*{1:t_{q(S,K)}^{(1)}}\ge \frac{q(S,K)}{q(S,K)+1}S}}} \\
    & =\bbE_{\bmu}^{\pi}\brk*{\sum_{k\in[K]}\Ind\crl*{E_{\sim,k}^{(1)}}\Ind{\crl*{S\prn*{1:t_{q(S,K)}^{(1)}}\ge \frac{q(S,K)}{q(S,K)+1}S}}}\\
    &\overset{\rm (iii)}{\ge} \bbE_{\bmu}^{\pi}\brk*{\prn*{K-\ceil*{\frac{S}{q(S,K)+1}}}\Ind{\crl*{S\prn*{1:t_{q(S,K)}^{(1)}}\ge \frac{q(S,K)}{q(S,K)+1}S}}}\\
    &=\prn*{K-\ceil*{\frac{S}{q(S,K)+1}}}\bbE_{\bmu}^{\pi}\brk*{\Ind{\crl*{S\prn*{1:t_{q(S,K)}^{(1)}}\ge \frac{q(S,K)}{q(S,K)+1}S}}},
\end{align*}
where (iii) follow from the  definition of $E_{\sim,k}^{(1)}$.

Combining the above two paragraphs, we have
\begin{align*}
    \sum_{j\in[q(S,K)+2]}\sum_{k\in[K]}\bbP_{\bmu}^{\pi}\prn*{E_{j,k}^{(1)}}
    &\ge\sum_{j\in[q(S,K)]}\sum_{k\in[K]}\bbP_{\bmu}^{\pi}\prn*{E_{j,k}^{(1)}}+   \sum_{k\in[K]}\bbP_{\bmu}^{\pi}\prn*{E_{q(S,K)+1,k}^{(1)}\cup E_{q(S,K)+2,k}^{(1)} }\\
    &\ge\prn*{K-\ceil*{\frac{S}{q(S,K)+1}}}\bbE_{\bmu}^\pi\brk*{\prn*{1- \Ind\crl*{S\prn*{1:t_{q(S,K)}^{(1)}}\ge \frac{q(S,K)}{q(S,K)+1}S}}}\\
    &~~~+\prn*{K-\ceil*{\frac{S}{q(S,K)+1}}}\bbE_{\bmu}^{\pi}\brk*{\Ind{\crl*{S\prn*{1:t_{q(S,K)}^{(1)}}\ge \frac{q(S,K)}{q(S,K)+1}S}}}\\
    &=K-\ceil*{\frac{S}{q(S,K)+1}}.
\end{align*}
$\hfill\Box$

\proof{Proof of \pref{lm:second-class}.}For any $j\in[q(S,K)]$, we have
\begin{align*}
\sum_{k\in[K]}\Ind\crl*{E_{j,k}^{(2)}}&=\text{number of actions that are not chosen in period }\brk*{t_{j-1}^{(2)}:t_{j}^{(2)}}\\
&\ge K-1-S\prn*{t_{j-1}^{(2)}:t_j^{(2)}}\\
&\ge \prn*{K-1-\floor*{\frac{S}{q(S,K)+2}}}\Ind\crl*{S\prn*{t_{j-1}^{(2)}:t_j^{(2)}}\le\frac{S}{q(S,K)+2}}
\end{align*}
almost surely. Thus for any $j\in[q(S,K)+2]$, we have
\begin{equation}\label{eq:lm7}
\begin{aligned}[b]
    \sum_{k\in[K]}\bbP_{\bmu}^{\pi}\prn*{E_{j,k}^{(2)}}& = \sum_{k\in[K]}\bbE_{\bmu}^{\pi}\brk*{\Ind\crl*{E_{j,k}^{(2)}}}\\
& =  \bbE_{\bmu}^{\pi}\brk*{\sum_{k\in[K]}\Ind\crl*{E_{j,k}^{(2)}}}\\
& \ge \bbE_{\bmu}^{\pi}\brk*{\prn*{K-1-\floor*{\frac{S}{q(S,K)+2}}}\Ind\crl*{S\prn*{t_{j-1}^{(2)}:t_j^{(2)}}\le\frac{S}{q(S,K)+2}}}\\
& =\prn*{K-1-\floor*{\frac{S}{q(S,K)+2}}}\bbE_{\bmu}^{\pi}\brk*{\Ind\crl*{S\prn*{t_{j-1}^{(2)}:t_j^{(2)}}\le\frac{S}{q(S,K)+2}}}
\end{aligned}
\end{equation}
Summing \pref{eq:lm7} over $j\in[q(S,K)+2]$, we have
\begin{align*}
    \sum_{j\in[q(S,K)+2]}  \sum_{k\in[K]}\bbP_{\bmu}^{\pi}\prn*{E_{j,k}^{(2)}}& \ge \prn*{K-1-\floor*{\frac{S}{q(S,K)+2}}}\bbE_{\bmu}^{\pi}\brk*{\sum_{j\in[q(S,K)+2]}\Ind\crl*{S\prn*{t_{j-1}^{(2)}:t_j^{(2)}}\le\frac{S}{q(S,K)+2}}}\\
    &\overset{\rm(i)}{\ge} \prn*{K-1-\floor*{\frac{S}{q(S,K)+2}}}\bbE_{\bmu}^{\pi}\brk*{\Ind\crl*{S\prn*{1:T}\le S}}\\
    &\overset{\rm(ii)}{=}K-1-\floor*{\frac{S}{q(S,K)+2}},
\end{align*}
where (i) follows from the pigeonhole principle and (ii) follows from the switching constraint.
$\hfill\Box$

\proof{Proof of \cref{cor:first-second-class}.}Since $K\ge 2$ and $0\le r(S,K)\le K-2$, we have
\begin{align*}
    {K-\ceil*{\frac{S}{q(S,K)+1}}}&=K-\ceil*{(K-1)-\frac{K-2-r(S,K)}{q(S,K)+1}}\\
    &\ge K-\min\crl*{K-1,\brk*{(K-1)-\frac{K-2-r(S,K)}{q(S,K)+1}+1}}\\
    &= \max\crl*{1,\frac{K-2-r(S,K)}{q(S,K)+1}}\\
    &\ge\frac{K-r(S,K)}{2(q(S,K)+2)}
\end{align*}
and
\begin{align*}
    K-1-\floor*{\frac{S}{q(S,K)+2}}&=K-1-\floor*{\frac{(K-1)q(S,K)+r(S,K)+1}{q(S,K)+2}}\\
    &\ge K-1-\brk*{\frac{(K-1)q(S,K)+r(S,K)+1}{q(S,K)+2}}\\
    &\ge \frac{2(K-1)-r(S,K)-1}{q(S,K)+2}\\
    &\ge\frac{K-1}{q(S,K)+2}\\
    &\ge\frac{K}{2(q(S,K)+2)}.
\end{align*}
Combining the above inequalities with \pref{lm:first-class} and \pref{lm:second-class}, we obtain \pref{cor:first-second-class}.
$\hfill\Box$

\subsection{Alternative Environments, Bad Events, and Lower Bound Reductions}
\underline{In the rest of the proof, we fix an arbitrary policy $\pi\in\Pi_S$.}

In this subsection, we define the following concepts: (i) reference environment \& reference measure, (ii) alternative environments \& alternative measures, and (iii) bad events. Based on these definitions, we explicitly construct two classes of environments $\Phi_1$ and $\Phi_2$, and reduce the task of proving lower bounds on the ``average-case regret''  $\frac{1}{|\Phi_1|}\sum_{\bmu\in\Phi}R_{\bmu}^{\pi}(T)$ and $\frac{1}{|\Phi_2|}\sum_{\bmu\in\Phi}R_{\bmu}^{\pi}(T)$ to the task of proving lower bounds on the ``average-case bad event probability''  $\frac{1}{(q(S,K)+2)K}\sum_{j\in[q(S,K)+2]}\sum_{k\in[K]}\bbP_{j,k}^{(1)}\prn*{E_{j,k}^{(1)}}$ and $\frac{1}{(q(S,K)+2)K}\sum_{j\in[q(S,K)+2]}\sum_{k\in[K]}\bbP_{j,k}^{(2)}\prn*{E_{j,k}^{(2)}}$, respectively.

Let $\bm{0}=(0,\dots,0)\in\bbR^K$ be the \emph{reference} environment. Let $\bbQ\ldef\bbP_{\bm{0}}^\pi$ denote the \emph{reference} measure.

\subsubsection{Results Associated with the First Class of Risky Events}\label{app:first-class}
~\\
\indent For any $j\in[q(S,K+2)]$, define a reward gap
\[
\Delta_{j}^{(1)}\ldef\begin{cases}
1, &\text{if }j=1,\\
\frac{1}{{2(q(S,K)+2)}}\sqrt{\frac{{K-r(s,k)}}{{t_{j-1}^{(1)}}}}, &\text{if }j\in[2:q(S,K)+1],\\
-\frac{1}{{2(q(S,K)+2)}}\sqrt{\frac{{K-r(s,k)}}{{t_{q(S,K)}^{(1)}}}}, &\text{if }j=q(S,K)+2.
\end{cases}
\]
Note that $\abs*{\Delta_j^{(1)}}\in[0,1]$ for all $j\in[q(S,K)+2]$.

For any $j\in[q(S,K)+2], k\in[K]$, define an \emph{alternative} environment $\bmu^{(1)}_{j,k}\ldef \prn*{\mu^{(1)}_{j,k;1},\dots,\mu^{(1)}_{j,k;K}}\in\bbR^K$ where
\[
\mu^{(1)}_{j,k;i}\ldef\begin{cases}\Delta_j^{(1)}, &\text{if }i=k,\\
0, &\text{otherwise.}
\end{cases}
\]
Note that each alternative environment $\bmu^{(1)}_{j,k}$ only differs from the reference environment in terms of the mean reward of action $k$.

For any  $j\in[q(S,K)+2], k\in[K]$, let $\bbP_{j,k}^{(1)}\ldef\bbP^{\pi}_{\bmu_{j,k}^{(1)}}$ denote the \emph{alternative} measure associated with the alternative environment $\bmu_{j,k}^{(1)}$. 

We explicitly construct a class of environments $\Phi_1\ldef\crl*{\bmu_{j,k}^{(1)}\mid j\in[q(S,K)+2], k\in[K]}$.

For any  $j\in[q(S,K)+2], k\in[K]$, under environment $\bmu_{j,k}^{(1)}$, the risky event $E_{j,k}^{(1)}$ becomes a \emph{bad event}\footnote{In our language, we call $E_{j,k}^{(1)}$ a \emph{risky event} for any environment, but a \emph{bad event} only for environment $\bmu_{j,k}^{(1)}$.} whose occurrence would lead to large regret. Specifically:
\begin{itemize}
    \item Suppose $j=1$. Since action $k$ is the unique optimal action under environment $\bmu_{1,k}^{(1)}$, choosing any action other than $k$ for one round incurs at least a $\Delta_1^{(1)}$ term in the policy's  regret, and the occurrence of $E_{1,k}^{(1)}=\crl*{\text{action }k \text{ is not chosen in period }\brk*{1:t_{1}^{(1)}}}$ incurs at least a ${t_1^{(1)}}\Delta_1^{(1)}$ term in the policy's  regret.
    \item Suppose $j\in[2:q(S,K)]$. Since action $k$ is the unique optimal action under environment $\bmu_{j,k}^{(1)}$, choosing any action other than $k$ for one round incurs at least a $\Delta_j^{(1)}$ term in the policy's  regret, and the occurrence of $E_{j,k}^{(1)}=\crl*{\text{action }k \text{ is not chosen in period }\brk*{t_{j-1}^{(1)}:t_{j}^{(1)}}}$ incurs at least a $\prn*{t_j^{(1)}-t_{j-1}^{(1)}+1}\Delta_j^{(1)}$ term in the policy's  regret.
    \item Suppose $j=q(S,K)+1$. Since action $k$ is the unique optimal action under environment $\bmu_{q(S,K)+1,k}^{(1)}$, choosing any action other than $k$ for one round incurs at least a $\Delta_{q(S,K)+1}^{(1)}$ term in the policy's  regret, and the occurrence of $E_{q(S,K)+1,k}^{(1)}=\crl*{\text{action }k \text{ is not chosen in period }\brk*{t_{q(S,K)}^{(1)}:\floor{\prn{t_{q(S,K)}^{(1)}+T}/2}}}$ incurs at least a $\prn*{\floor{\prn{t_{q(S,K)}^{(1)}+T}/2}-t_{q(S,K)}^{(1)}+1}\Delta_{q(S,K)+1}^{(1)}$ term in the policy's  regret.
    \item Suppose $j=q(S,K)+2$. Since action $k$ is the worst action under environment $\bmu_{q(S,K)+2,k}^{(1)}$, choosing action $k$ for one round incurs at least a $-\Delta_{q(S,K)+2}^{(1)}$ term in the policy's regret. Furthermore, since the occurrence of $E_{q(S,K)+2,k}^{(1)}$ implies the occurrence of $\crl*{\text{action }k\text{ is chosen in every round in }[\floor{\prn{t_{q(S,K)}^{(1)}+T}/2}:T]}$ (because of the switching constraint), it incurs at least a $-\prn*{T-\floor{\prn{t_{q(S,K)}^{(1)}+T}/2}+1}\Delta_{q(S,K)+2}^{(1)}$ term in the policy's  regret.
\end{itemize}
The above arguments lead to \cref{lm:bad}.

\begin{lemma}[From risky events to bad events]\label{lm:bad} For any $j\in[q(S,K)+2], k\in[K]$, under environment $\bmu_{j,k}^{(1)}$, the risky event $E_{j,k}^{(1)}$ becomes a \emph{bad event} in the sense that
\[\bbE_{\bmu_{j,k}^{(1)}}^{\pi}\brk*{T\mu_{j,k;k}^{(1)}-\sum_{t=1}^T\mu_{j,k;a_t}^{(1)} \mid E_{j,k}^{(1)}}\ge \cR_{\rm bad}\prn{S,K,T},\]
where \[\cR_{\rm bad}(S,K,T)\ldef\frac{\prn*{K-r(S,K)}}{8\prn{q(S,K)+2}}{\prn*{\frac{T}{{K-r(S,K)}}}^{\frac{1}{2-2^{-q(S,K)}}}}\]
is a universal lower bound on the ``distribution-dependent regret conditional on the bad event.''
\end{lemma}

\proof{Proof of \cref{lm:bad}.} By the arguments in the previous paragraph, we have
\begin{align*}
\bbE_{\bmu_{j,k}^{(1)}}^{\pi}\brk*{T\mu_{j,k;k}^{(1)}-\sum_{t=1}^T\mu_{j,k;a_t}^{(1)} \mid E_{j,k}^{(1)}}
&\ge\bbE_{\bmu_{j,k}^{(1)}}^{\pi}\brk*{\prn*{t_j^{(1)}-t_{j-1}^{(2)}+1}\mu_{j,k;k}^{(1)}-\sum_{t\in \brk*{t_{j-1}^{(1)}:t_{j}^{(1)}}} \mu_{j,k;a_t}^{(1)} \mid E_{j,k}^{(1)}}\\
&\ge    \begin{cases}
        t_1^{(1)}\Delta_{1}^{(1)}, &\text{if }j=1,\\
        \prn*{t_j^{(1)}-t_{j-1}^{(1)}+1}\Delta_{j}^{(1)}, &\text{if }j\in[2:q(S,K)],\\
        \prn*{\floor{\prn{t_{q(S,K)}^{(1)}+T}/2}-t_{q(S,K)}^{(1)}+1}\Delta_{q(S,K)+1}^{(1)}, &\text{if }j=q(S,K)+1,\\
        -\prn*{T-\floor{\prn{t_{q(S,K)}^{(1)}+T}/2}+1}\Delta_{q(S,K)+2}^{(1)}, &\text{if }j=q(S,K)+2,
        \end{cases}\\
&\ge \frac{\prn*{K-r(S,K)}^{1-\frac{1}{2-2^{-q(S,K)}}}}{8(q(S,K)+2)}{{{T}}^{\frac{1}{2-2^{-q(S,K)}}}},
\end{align*}
where the last inequality follows from the following inequalities:
\[t_1^{(1)}\Delta_{1}^{(1)}=t_1^{(1)}\ge\frac{\prn*{K-r(S,K)}^{1-\frac{1}{2-2^{-q(S,K)}}}}{(q(S,K)+2)}{{{T}}^{\frac{1}{2-2^{-q(S,K)}}}},
\]
\begin{align*}
   & ~~\prn*{t_j^{(1)}-t_{j-1}^{(1)}+1}\Delta_{j}^{(1)}\\
   &\ge \prn*{K-r(S,K)}\prn*{{\prn*{\frac{T}{{K-r(S,K)}}}^{\frac{2-2^{1-j}}{2-2^{-q(S,K)}}}}-{\prn*{\frac{T}{{K-r(S,K)}}}^{\frac{2-2^{2-j}}{2-2^{-q(S,K)}}}}}\Delta_{j}^{(1)}\\
   &\ge \frac{\prn*{K-r(S,K)}}{2\prn{q(S,K)+2}}\prn*{{\prn*{\frac{T}{{K-r(S,K)}}}^{\frac{2-2^{1-j}}{2-2^{-q(S,K)}}}}-{\prn*{\frac{T}{{K-r(S,K)}}}^{\frac{2-2^{2-j}}{2-2^{-q(S,K)}}}}}{\prn*{\frac{T}{{K-r(S,K)}}}^{-\frac{1-2^{1-j}}{2-2^{-q(S,K)}}}}\\
   &=\frac{\prn*{K-r(S,K)}}{2\prn{q(S,K)+2}}\prn*{{\prn*{\frac{T}{{K-r(S,K)}}}^{\frac{1}{2-2^{-q(S,K)}}}}-{\prn*{\frac{T}{{K-r(S,K)}}}^{\frac{1-2^{1-j}}{2-2^{-q(S,K)}}}}}\\
   &=\frac{\prn*{K-r(S,K)}}{2\prn{q(S,K)+2}}{\prn*{\frac{T}{{K-r(S,K)}}}^{\frac{1}{2-2^{-q(S,K)}}}}\prn*{1-{\prn*{\frac{T}{{K-r(S,K)}}}^{\frac{-2^{1-j}}{2-2^{-q(S,K)}}}}}\\
   &\ge\frac{\prn*{K-r(S,K)}}{2\prn{q(S,K)+2}}{\prn*{\frac{T}{{K-r(S,K)}}}^{\frac{1}{2-2^{-q(S,K)}}}}\prn*{1-{\prn*{\frac{T}{{K-r(S,K)}}}^{\frac{-2^{-q(S,K)}}{2-2^{-q(S,K)}}}}}\\
   &\ge \frac{\prn*{K-r(S,K)}}{2\prn{q(S,K)+2}}{\prn*{\frac{T}{{K-r(S,K)}}}^{\frac{1}{2-2^{-q(S,K)}}}}\prn*{1-{\prn*{\frac{T}{{K-r(S,K)}}}^{{-2^{-q(S,K)-1}}}}}\\
   &\overset{\rm(i)}{\ge}\frac{\prn*{K-r(S,K)}}{2\prn{q(S,K)+2}}{\prn*{\frac{T}{{K-r(S,K)}}}^{\frac{1}{2-2^{-q(S,K)}}}}\prn*{1-{\prn*{\frac{T}{{K-r(S,K)}}}^{-\frac{1}{\log_2(T/K)}}}}\\
    &\ge\frac{\prn*{K-r(S,K)}}{2\prn{q(S,K)+2}}{\prn*{\frac{T}{{K-r(S,K)}}}^{\frac{1}{2-2^{-q(S,K)}}}}\prn*{1-{\prn*{T/K}^{-\frac{1}{\log_2(T/K)}}}}\\
    &=\frac{\prn*{K-r(S,K)}}{4\prn{q(S,K)+2}}{\prn*{\frac{T}{{K-r(S,K)}}}^{\frac{1}{2-2^{-q(S,K)}}}},~~~~\forall j\in[2:q(S,K)+1],
\end{align*}
\begin{align*}
\prn*{\floor{\prn{t_{q(S,K)}^{(1)}+T}/2}-t_{q(S,K)}^{(1)}+1}\Delta_{q(S,K)+1}^{(1)}&\ge\frac{1}{2}\prn*{t_{q(S,K)+1}^{(1)}-t_{q(S,K)}^{(1)}+1}\Delta_{q(S,K)+1}^{(1)}\\
&\ge\frac{\prn*{K-r(S,K)}}{8\prn{q(S,K)+2}}{\prn*{\frac{T}{{K-r(S,K)}}}^{\frac{1}{2-2^{-q(S,K)}}}},
\end{align*}
\begin{align*}
    -\prn*{T-\floor{\prn{t_{q(S,K)}^{(1)}+T}/2}+1}\Delta_{q(S,K)+2}^{(1)}&\ge\prn*{\floor{\prn{t_{q(S,K)}^{(1)}+T}/2}-t_{q(S,K)}^{(1)}+1}\Delta_{q(S,K)+1}^{(1)}\\
    &\ge\frac{\prn*{K-r(S,K)}}{8\prn{q(S,K)+2}}{\prn*{\frac{T}{{K-r(S,K)}}}^{\frac{1}{2-2^{-q(S,K)}}}}.
\end{align*}
Note that in (i) we utilize the fact that $q(S,K)+1\le\log_2\log_2(T/K)$.
$\hfill\Box$

Based on \cref{lm:bad}, we can reduce the task of proving a lower bound on the policy's (distribution-dependent) regret $R_{\bmu_{j,k}^{(1)}}^{\pi}(T)$ to the task of proving a lower bound on the bad event probability $\bbP_{j,k}^{(1)}\prn*{E_{j,k}^{(1)}}$. Consequently, we can reduce the task of proving a lower bound on the policy's ``average-case regret''  $\frac{1}{\abs{\Phi_1}}\sum_{\bmu\in\Phi_1}R^{\pi}_{\bmu}(T)$ to the task of proving a lower bound on the ``average-case bad event probability'' $\frac{1}{(q(S,K)+2)K}\sum_{j\in[q(S,K)+2]}\sum_{k\in[K]}\bbP_{j,k}^{(1)}\prn*{E_{j,k}^{(1)}}$.
\begin{lemma}[Reducing regret lower bounds to bad event probability lower bounds]\label{lm:reduction} For any $j\in[q(S,K)+2], k\in[K]$, we have
\[
R_{\bmu_{j,k}^{(1)}}^{\pi}(T)\ge \cR_{\rm bad}(S,K,T)\cdot \bbP_{j,k}^{(1)}\prn*{E_{j,k}^{(1)}}.
\]
As a result, we have
\begin{align*}
    R^{\pi}(T)\ge \frac{1}{\abs{\Phi_1}}\sum_{\bmu\in\Phi_1}R^{\pi}_{\bmu}(T)\ge\cR_{\rm bad}(S,K,T)\cdot\frac{1}{(q(S,K)+2)K}\sum_{j\in[q(S,K)+2]}\sum_{k\in[K]}\bbP_{j,k}^{(1)}\prn*{E_{j,k}^{(1)}}.
\end{align*}
\end{lemma}
\proof{Proof of \cref{lm:reduction}.}
For any $j\in[q(S,K)+2], k\in[K]$, by \cref{lm:bad}, we have
\begin{align*}
R_{\bmu_{j,k}^{(1)}}^{\pi}(T)&= \bbE_{\bmu_{j,k}^{(1)}}^{\pi}\brk*{T\mu_{j,k;k}^{(1)}-\sum_{t=1}^T\mu_{j,k;a_t}^{(1)}}\\
&\ge \bbE_{\bmu_{j,k}^{(1)}}^{\pi}\brk*{T\mu_{j,k;k}^{(1)}-\sum_{t=1}^T\mu_{j,k;a_t}^{(1)} \mid E_{j,k}^{(1)}}\cdot \bbP_{j,k}^{(1)}\prn*{E_{j,k}^{(1)}}\\
&\ge \cR_{\rm bad}(S,K,T)\cdot \bbP_{j,k}^{(1)}\prn*{E_{j,k}^{(1)}},
\end{align*}
and hence
\begin{align*}
    R^{\pi}(T)&=\sup_{\cD}R_{\cD}^{\pi}(T)\\
    &\ge \sup_{\bmu\in\Phi_1} R_{\bmu}^{\pi}(T)\\
    &\ge \frac{1}{\abs{\Phi_1}}\sum_{\bmu\in\Phi_1}R^{\pi}_{\bmu}(T)\\
    &\ge \frac{1}{(q(S,K)+2)K}\sum_{j\in[q(S,K)+2]}\sum_{k\in[K]}R_{\bmu_{j,k}^{(1)}}^{\pi}(T)\\
    &\ge \cR_{\rm bad}(S,K,T)\cdot\frac{1}{(q(S,K)+2)K}\sum_{j\in[q(S,K)+2]}\sum_{k\in[K]}\bbP_{j,k}^{(1)}\prn*{E_{j,k}^{(1)}}.
\end{align*}
$\hfill\Box$

\subsubsection{Results Associated with the Second Class of Risky Events}
~\\
\indent For any $j\in[q(S,K+2)]$, define a reward gap
\[
\Delta_{j}^{(2)}\ldef\begin{cases}
1, &\text{if }j=1,\\
\frac{1}{{2(q(S,K)+2)}}\sqrt{\frac{{K}}{{t_{j-1}^{(2)}}}}, &\text{if }j\in[2:q(S,K)+2].
\end{cases}
\]
Note that $\abs*{\Delta_j^{(2)}}\in[0,1]$ for all $j\in[q(S,K)+2]$.

For any $j\in[q(S,K)+2], k\in[K]$, define an \emph{alternative} environment $\bmu^{(2)}_{j,k}\ldef \prn*{\mu^{(2)}_{j,k;1},\dots,\mu^{(2)}_{j,k;K}}\in\bbR^K$ where
\[
\mu^{(2)}_{j,k;i}\ldef\begin{cases}\Delta_j^{(2)}, &\text{if }i=k,\\
0, &\text{otherwise.}
\end{cases}
\]
Note that each alternative environment $\bmu^{(1)}_{j,k}$ only differs from the reference environment in terms of the mean reward of action $k$.

For any  $j\in[q(S,K)+2], k\in[K]$, let $\bbP_{j,k}^{(2)}\ldef\bbP^{\pi}_{\bmu_{j,k}^{(2)}}$ denote the \emph{alternative} measure associated with the alternative environment $\bmu_{j,k}^{(2)}$. 

We explicitly construct a class of environments $\Phi_2\ldef\crl*{\bmu_{j,k}^{(2)}\mid j\in[q(S,K)+2], k\in[K]}$.

For any  $j\in[q(S,K)+2], k\in[K]$, under environment $\bmu_{j,k}^{(2)}$, the risky event $E_{j,k}^{(2)}$ becomes a \emph{bad event} whose occurrence would lead to large regret. Similar to our analysis in \pref{app:first-class}, we have the following two lemmas.

\begin{lemma}[From risky events to bad events]\label{lm:bad2} For any $j\in[q(S,K)+2], k\in[K]$, under environment $\bmu_{j,k}^{(2)}$, the risky event $E_{j,k}^{(2)}$ becomes a \emph{bad event} in the sense that
\[\bbE_{\bmu_{j,k}^{(2)}}^{\pi}\brk*{T\mu_{j,k;k}^{(2)}-\sum_{t=1}^T\mu_{j,k;a_t}^{(2)} \mid E_{j,k}^{(2)}}\ge \cR_{\rm bad2}\prn{S,K,T},\]
where \[\cR_{\rm bad2}(S,K,T)\ldef\frac{{K}}{4\prn{q(S,K)+2}}{\prn*{\frac{T}{{K}}}^{\frac{1}{2-2^{-q(S,K)-1}}}}\]
is a universal lower bound on the ``distribution-dependent regret conditional on the bad event.''
\end{lemma}

\begin{lemma}[Reducing regret lower bounds to bad event probability lower bounds]\label{lm:reduction2} For any $j\in[q(S,K)+2], k\in[K]$, we have
\[
R_{\bmu_{j,k}^{(2)}}^{\pi}(T)\ge \cR_{\rm bad2}(S,K,T)\cdot \bbP_{j,k}^{(2)}\prn*{E_{j,k}^{(2)}}.
\]
As a result, we have
\begin{align*}
    R^{\pi}(T)\ge\frac{1}{\abs{\Phi_2}}\sum_{\bmu\in\Phi_2}R_{\bmu}^\pi(T)\ge\cR_{\rm bad2}(S,K,T)\cdot\frac{1}{(q(S,K)+2)K}\sum_{j\in[q(S,K)+2]}\sum_{k\in[K]}\bbP_{j,k}^{(2)}\prn*{E_{j,k}^{(2)}}.
\end{align*}
\end{lemma}

\subsection{Probability Space Changing Tricks}\label{sec:changing}
\cref{lm:reduction} and \cref{lm:reduction2} indicate that, in order to prove the desired lower bound on the regret $R^\pi(T)$, it suffices to prove the following two statements:
\begin{equation}\label{eq:obj1}
    \wb{p^{(1)}}\ldef\frac{1}{(q(S,K)+2)K}\sum_{j\in[q(S,K)+2]}\sum_{k\in[K]}\bbP_{j,k}^{(1)}\prn*{E_{j,k}^{(1)}}=\wt{\Omega}\prn*{\frac{K-r(S,K)}{K}},
\end{equation}
\begin{equation}\label{eq:obj2}
    \wb{p^{(2)}}\ldef\frac{1}{(q(S,K)+2)K}\sum_{j\in[q(S,K)+2]}\sum_{k\in[K]}\bbP_{j,k}^{(2)}\prn*{E_{j,k}^{(2)}}=\wt{\Omega}\prn*{1}.
\end{equation}
That is, we only need to establish tight lower bounds on the average probability $  \wb{p^{(1)}}$ and the average probability $\wb{p^{(2)}}$.

By \cref{cor:first-second-class}, we have
\[
\wb{q^{(1)}}\ldef\frac{1}{(q(S,K)+2)K}\sum_{j\in[q(S,K)+2]}\sum_{k\in[K]}\bbQ\prn*{E_{j,k}^{(1)}}=\wt{\Omega}\prn*{\frac{K-r(S,K)}{K}},
\]
\[
\wb{q^{(2)}}\ldef\frac{1}{(q(S,K)+2)K}\sum_{j\in[q(S,K)+2]}\sum_{k\in[K]}\bbQ\prn*{E_{j,k}^{(2)}}=\wt{\Omega}\prn*{1}.
\]
Therefore, in order to show \cref{eq:obj1}, it suffices to show that $\wb{p^{(1)}}$ is close to $\wb{q^{(1)}}$; in order to show \cref{eq:obj2}, it suffices to show that  $\wb{p^{(2)}}$ is close to $\wb{q^{(2)}}$.

Let us first focus on the relationship between $\wb{p^{(1)}}$ and $\wb{q^{(1)}}$. Note that $\wb{p^{(1)}}$ is the average of the sequence $\crl*{\bbP_{j,k}^{(1)}\prn*{E_{j,k}^{(1)}}}$\footnote{We use the notation $\crl*{\bbP_{j,k}^{(1)}\prn*{E_{j,k}^{(1)}}}$ to represent the sequence $\prn*{\bbP_{j,k}^{(1)}\prn*{E_{j,k}^{(1)}}}_{j\in[q(S,K)+2],k\in[K]}$. In general, we use  $\crl*{a_{j,k}}$ to represent a sequence $\prn*{a_{j,k}}_{j\in[q(S,K)+2],k\in[K]}$.} (where a sequence of events $\crl*{E_{j,k}^{(1)}}$ are evaluated by a sequence of varying alternative measures $\crl*{\bbP_{j,k}^{(1)}}$), while $\wb{q^{(1)}}$ is the average of the sequence $\crl*{\bbQ\prn*{E_{j,k}^{(1)}}}$ (where the same sequence of events $\crl*{E_{j,k}^{(1)}}$  are evaluated by a single and fixed reference measure $\bbQ$). Intuitively, we just need a ``change of  measure'' / information-theoretic argument --- if the alternative measures $\crl*{\bbP_{j,k}^{(1)}}$ are ``close enough'' to the reference measure $\bbQ$, then $\wb{p^{(1)}}$ is close to $\wb{q^{(1)}}$.

Unfortunately, it turns out that the divergence between $\crl*{\bbP_{j,k}^{(1)}}$ and  $\bbQ$ is too large to make the above argument work. An important reason is that such an argument directly deals with the underlying measures $\crl*{\bbP_{j,k}^{(1)}}$ and  $\bbQ$, thus 
completely overlooks the special structures of the risky event sequence $\crl*{E_{j,k}^{(1)}}$.  Therefore, if we want to show that $\wb{p^{(1)}}$ is close to $\wb{q^{(1)}}$, we need to integrate the structural properties of risky events into our argument. 

The same challenge exists when we want to show that $\wb{p^{(2)}}$ is close to $\wb{q^{(2)}}$.

We develop \emph{probability space changing tricks} to address this challenge. See below.

\subsubsection{Results Associated with the First Class of Risky Events}\label{app:pct1}
~\\
\indent We start with some key structural properties of  the risky event sequence $\crl*{E_{j,k}^{(1)}}$.
\begin{itemize}
    \item For any $k\in[K]$, the occurrence of the event $E_{1,k}^{(1)}=\crl*{\text{action }k \text{ is not chosen in period }\brk*{1:t_{1}^{(1)}}}$ is independent of the random variables $\prn*{X_{\bmu}^t(k)}_{t\in\brk{1:t_{1}^{(1)}}}$ and the random variables $\prn*{X_{\bmu}^t(i)}_{t\in\brk{t_{1}^{(1)}+1:T},i\in[K]}$.
    \item For any $j\in[2:q(S,K)]$, $k\in[K]$, the occurrence of the event $E_{j,k}^{(1)}=\crl*{\text{action }k \text{ is not chosen in period }\brk*{t_{j-1}^{(1)}:t_{j}^{(1)}}}$ is independent of the random variables  $\prn*{X_{\bmu}^t(k)}_{t\in\brk{t_{j-1}^{(1)}:t_j^{(1)}}}$ and the random variables $\prn*{X_{\bmu}^t(i)}_{t\in\brk{t_{j}^{(1)}+1:T},i\in[K]}$.
    \item For any $k\in[K]$, the occurrence of the event $E_{q(S,K)+1,k}^{(1)}=\crl*{\text{action }k \text{ is not chosen in period }\brk*{t_{q(S,K)}^{(1)}:\floor{\prn{t_{q(S,K)}^{(1)}+T}/2}}}$ is independent of the random variables  $\prn*{X_{\bmu}^t(k)}_{t\in\brk{t_{q(S,K)}^{(1)}:\floor{\prn{t_{q(S,K)}^{(1)}+T}/2}}}$ and the random variables $\prn*{X_{\bmu}^t(i)}_{t\in\brk{\floor{\prn{t_{q(S,K)}^{(1)}+T}/2}+1:T},i\in[K]}$.
    \item For any $k\in[K]$, the occurrence of the event $E_{q(S,K)+2,k}^{(1)}\ldef\crl*{\tau\le \floor{\prn{t_{q(S,K)}^{(1)}+T}/2},\, a_\tau=k, \text{ action }k\text{ is not chosen in period }\brk*{t_{q(S,K)}^{(1)}:\tau-1}}$ is independent of the random variables $\prn*{X_{\bmu}^t(k)}_{t\in\brk{t_{q(S,K)}^{(1)}:\floor{\prn{t_{q(S,K)}^{(1)}+T}/2}}}$ and the random variables $\prn*{X_{\bmu}^t(i)}_{t\in\brk{\floor{\prn{t_{q(S,K)}^{(1)}+T}/2}+1:T},i\in[K]}$. (Note that this property crucially relies on the delicate design of $E_{q(S,K)+2,k}^{(1)}$.)
\end{itemize}
The above properties indicate that, when we want to represent the probability of a risky event $E_{j,k}^{(1)}$ under an environment $\bmu$, we do not need to use the ``full'' measure induced by $\cH_T$ (i.e., $\bbP_{\bmu}^{\pi}$) or the ``natural'' measure induced by $\cH_{t_j^{(1)}}$. Instead, we can use a ``ristricted'' measure which deliberately ``ignores'' certain reward information associated with action $k$ --- thanks to the structural property of  $E_{j,k}^{(1)}$, such ignorance would not affect the measure's well-definedness and value on $E_{j,k}^{(1)}$. Moreover, since the alternative environment $\bmu_{j,k}^{(1)}$ only differs from the reference environment in terms of the mean reward of action $k$, such ignorance can help make the measures associated with the two environments ``closer.'' We can then establish tighter bounds on the distance between $\wb{p^{(1)}}$ and $\wb{q^{(1)}}$.

Motivated by the above idea, we design two sequences of \emph{artificial} measures $\crl*{\bbP_{j,k}'}$ and  $\crl*{\bbQ_{j,k}'}$ as follows.
 
\textbf{Artificial measures $\crl*{\bbP_{j,k}'}$.} For any $j\in[2:q(S,K)]$, $k\in[K]$, let $\bbP'_{j,k}$ be the probability measure induced by the joint random variable
\begin{equation}\label{eq:censor}
\prn*{\prn*{a_t,X_{\bmu_{j,k}^{(1)}}^t(a_t)}_{t\in[1:t_{j-1}^{(1)}-1]},\prn*{a_t,X_{\bmu_{j,k}^{(1)}}^t(a_t)\Ind\crl*{a_t\ne k} }_{t\in[t_{j-1}^{(1)}:t_j^{(1)}]}}.
\end{equation}
For $j=1$, for any $k\in[K]$, let $\bbP'_{j,k}$ be the probability measure induced by the joint random variable
\[
\prn*{\prn*{a_t,X_{\bmu_{j,k}^{(1)}}^t(a_t)\Ind\crl*{a_t\ne k} }_{t\in[1:t_j^{(1)}]}}.
\]
For $j\in\crl{q(S,K)+1,q(S,K)+2}$, let  $\bbP'_{j,k}$ be the probability measure induced by the joint random variable
\[
\prn*{\prn*{a_t,X_{\bmu_{j,k}^{(1)}}^t(a_t)}_{t\in[1:t_{q(S,K)}^{(1)}-1]},\prn*{a_t,X_{\bmu_{j,k}^{(1)}}^t(a_t)\Ind\crl*{a_t\ne k} }_{t\in[t_{q(S,K)}^{(1)}:\floor{\prn{t_{q(S,K)}^{(1)}+T}/2}]}}.
\]

\textbf{Artificial reference measures $\crl*{\bbQ_{j,k}'}$.} For any $j\in[2:q(S,K)]$, $k\in[K]$, let $\bbQ'_{j,k}$ be the probability measure induced by the joint random variable
\begin{equation*}
\prn*{\prn*{a_t,X_{\bm{0}}^t(a_t)}_{t\in[1:t_{j-1}^{(1)}-1]},\prn*{a_t,X_{\bm{0}}^t(a_t)\Ind\crl*{a_t\ne k} }_{t\in[t_{j-1}^{(1)}:t_j^{(1)}]}}.
\end{equation*}
For $j=1$, for any $k\in[K]$, let $\bbQ'_{j,k}$ be the probability measure induced by the joint random variable
\[
\prn*{\prn*{a_t,X_{\bm{0}}^t(a_t)\Ind\crl*{a_t\ne k} }_{t\in[1:t_j^{(1)}]}}.
\]
For $j\in\crl{q(S,K)+1,q(S,K)+2}$, let  $\bbQ'_{j,k}$ be the probability measure induced by the joint random variable
\[
\prn*{\prn*{a_t,X_{\bm{0}}^t(a_t)}_{t\in[1:t_{q(S,K)}^{(1)}-1]},\prn*{a_t,X_{\bm{0}}^t(a_t)\Ind\crl*{a_t\ne k} }_{t\in[t_{q(S,K)}^{(1)}:\floor{\prn{t_{q(S,K)}^{(1)}+T}/2}]}}.
\]

Let us provide some explanations on the above definitions of artificial measures. For brevity, we focus on the case of $j\in[2:q(S,K)]$. Note that \pref{eq:censor} is \emph{not} the total data collected by $\pi$ under environment $\bmu_{j,k}^{(1)}$ during period $\brk*{1:t_j^{(1)}}$, which should be
\[\cH_{t_j^{(1)}}=\prn*{\prn*{a_t,X_{\bmu_{j,k}^{(1)}}^t(a_t)}_{t\in[1:t_{j}^{(1)}]}};\]
instead, \pref{eq:censor} is a censored variant of $\cH_{t_j^{(1)}}$, with all the reward observations associated with action $k$ during period  $\brk*{1:t_j^{(1)}}$ being ``ignored.'' Consequently, $\bbP'_{j,k}$ is \emph{neither} a measure on $\prn*{\Omega_{T},\cF_{T}}$ \emph{nor} a measure on $\prn*{\Omega_{t_j^{(1)}},\cF_{t_j^{(1)}}}$; instead, it is a measure on a ``restricted'' measurable space  $\prn*{\Omega_{j,k}',\cF'_{j,k}}$, where \[\Omega_{j,k}'\ldef \Omega_{t_{j}^{(1)}}\setminus\prn*{\Omega_{t_{j-1}^{(1)}-1}\times\prn*{\crl{k}\times \prn{\bbR\setminus{\crl{0}}}}^{t_j^{(1)}-t_{j-1}^{(1)}+1}},~~\cF_{j,k}'\ldef\cB\prn*{\Omega_{j,k}'}.\] 
Since $\prn*{\Omega_{j,k}',\cF'_{j,k}}\subset\prn*{\Omega_{t_j^{(1)}},\cF_{t_j^{(1)}}}\subset\prn*{\Omega_{T},\cF_{T}}$, the artificial measure $\bbP'_{j,k}$ can be seen as the restriction of $\bbP^{(1)}_{j,k}$ to a ``much smaller'' measurable space which still keeps $E_{j,k}^{(1)}$  measurable.\footnote{In measure theory, $\cF'_{j,k}$ is called a sub-$\sigma$-algebra, and $\bbP'_{j,k}$ is called a \emph{restricted measure}.} Similarly, the artificial reference measure $\bbQ'_{j,k}$ is the restriction of the reference measure $\bbQ$ to the same measurable space $\prn*{\Omega_{j,k}',\cF'_{j,k}}$. Such restrictions guarantee two nice properties:
\begin{enumerate}
    \item $\bbP_{j,k}'(E_{j,k}^{(1)})=\bbP_{j,k}^{(1)}(E_{j,k}^{(1)})$ and $\bbQ_{j,k}'(E_{j,k}^{(1)})=\bbQ(E_{j,k}^{(1)})$ for all $j,k$.
    \item Since the alternative environment $\bmu_{j,k}^{(1)}$ only differs from the reference environment $\bm{0}$ in terms of the mean reward of action $k$, the divergence between $\bbP'_{j,k}$ and $\bbQ'_{j,k}$ becomes much smaller on the new measurable space $\prn*{\Omega_{j,k}',\cF'_{j,k}}$, compared with the divergence between $\bbP^{(1)}_{j,k}$ and $\bbQ$ on the original measurable space $\prn*{\Omega_T,\cF_T}$.
\end{enumerate}

We now use the policy's behavior under the fixed reference environment $\bm{0}$ to bound the \emph{reverse} KL divergence between $\bbP'_{j,k}$ and $\bbQ'_{j,k}$. Using a standard ``divergence decomposition'' lemma (see, e.g., Lemma 15.1 of \citealt{lattimore2020bandit}), we have
\[
D_{\rm re}\prn*{ \bbP'_{j,k} \parallel\bbQ'_{j,k}}=D_{\rm KL}\prn*{\bbQ'_{j,k} \parallel \bbP'_{j,k}}=\bbE_{\bm{0}}^\pi\brk*{\sum_{t=1}^{t_{j-1}^{(1)}-1}\frac{\brk*{\Delta_{j}^{(1)}}^2}{2}\Ind\crl*{a_t=k}}=\frac{\brk*{\Delta_{j}^{(1)}}^2}{2}\bbE_{\bm{0}}^\pi\brk*{\sum_{t=1}^{t_{j-1}^{(1)}-1}\Ind\crl*{a_t=k}}
\]
for all $j\in[2:q(S,K)]$ and $k\in[K]$. 
This implies that
\[
\sum_{k=1}^K D_{\rm re}\prn*{ \bbP'_{j,k} \parallel\bbQ'_{j,k}}=\sum_{k=1}^K\frac{\brk*{\Delta_{j}^{(1)}}^2}{2}\bbE_{\bm{0}}^\pi\brk*{\sum_{t=1}^{t_{j-1}^{(1)}-1}\Ind\crl*{a_t=k}}=\frac{\brk*{\Delta_{j}^{(1)}}^2}{2}\prn*{t_{j-1}^{(1)}-1}
\]
for all $j\in[2:q(S,K)]$. Similarly, we have $\sum_{k=1}^K D_{\rm re}\prn*{ \bbP'_{j,k} \parallel\bbQ'_{j,k}}=0$ for $j=0$ and 
\[
\sum_{k=1}^K D_{\rm re}\prn*{ \bbP'_{j,k} \parallel\bbQ'_{j,k}}=\sum_{k=1}^K\frac{\brk*{\Delta_{j}^{(1)}}^2}{2}\bbE_{\bm{0}}^\pi\brk*{\sum_{t=1}^{t_{q(S,K)}^{(1)}-1}\Ind\crl*{a_t=k}}=\frac{\brk*{\Delta_{j}^{(1)}}^2}{2}\prn*{t_{q(S,K)}^{(1)}-1}
\]
for $j\in\crl{q(S,K)+1,q(S,K)+2}$. Combining with the definitions of $\Delta_{j}^{(1)}$ in \pref{app:first-class}, we have 
\[
\sum_{k=1}^K D_{\rm re}\prn*{ \bbP'_{j,k} \parallel\bbQ'_{j,k}}\le\frac{K-r(S,K)}{8(q(S,K)+2)^2}
\]
for all $j\in[q(S,K)+2]$. Therefore, we have the following lemma.

\begin{lemma}\label{lm:close1}
It holds that
\[
\sum_{j\in[q(S,K)+2]}\sum_{k\in[K]}D_{\rm re}\prn*{ \bbP'_{j,k} \parallel\bbQ'_{j,k}} \le \frac{K-r(S,K)}{8(q(S,K)+2)}.
\]
Combined with \pref{cor:first-second-class}, this implies
\[
\frac{1}{(q(S,K)+2)K}\sum_{j\in[q(S,K)+2]}\sum_{k\in[K]}D_{\rm re}\prn*{ \bbP'_{j,k} \parallel\bbQ'_{j,k}} \le \frac{K-r(S,K)}{8(q(S,K)+2)^2K}\le\frac{\wb{q^{(1)}}}{4}.
\]
\end{lemma}

\subsubsection{Results Associated with the Second Class of Risky Events}
~\\
\indent Similar to \pref{app:pct1}, we design two sequences of \emph{artificial} measures $\crl*{\bbP_{j,k}''}$ and  $\crl*{\bbQ_{j,k}''}$ as follows.
 
\textbf{Artificial measures $\crl*{\bbP_{j,k}''}$.} For any $j\in[2:q(S,K)+1]$, $k\in[K]$, let $\bbP''_{j,k}$ be the probability measure induced by the joint random variable
\begin{equation*}
\prn*{\prn*{a_t,X_{\bmu_{j,k}^{(1)}}^t(a_t)}_{t\in[1:t_{j-1}^{(2)}-1]},\prn*{a_t,X_{\bmu_{j,k}^{(2)}}^t(a_t)\Ind\crl*{a_t\ne k} }_{t\in[t_{j-1}^{(2)}:t_j^{(2)}]}}.
\end{equation*}
For $j=1$, for any $k\in[K]$, let $\bbP''_{j,k}$ be the probability measure induced by the joint random variable
\[
\prn*{\prn*{a_t,X_{\bmu_{j,k}^{(2)}}^t(a_t)\Ind\crl*{a_t\ne k} }_{t\in[1:t_j^{(2)}]}}.
\]

\textbf{Artificial reference measures $\crl*{\bbQ_{j,k}''}$.} For any $j\in[2:q(S,K)]$, $k\in[K]$, let $\bbQ''_{j,k}$ be the probability measure induced by the joint random variable
\begin{equation*}
\prn*{\prn*{a_t,X_{\bm{0}}^t(a_t)}_{t\in[1:t_{j-1}^{(2)}-1]},\prn*{a_t,X_{\bm{0}}^t(a_t)\Ind\crl*{a_t\ne k} }_{t\in[t_{j-1}^{(2)}:t_j^{(2)}]}}.
\end{equation*}
For $j=1$, for any $k\in[K]$, let $\bbQ''_{j,k}$ be the probability measure induced by the joint random variable
\[
\prn*{\prn*{a_t,X_{\bm{0}}^t(a_t)\Ind\crl*{a_t\ne k} }_{t\in[1:t_j^{(2)}]}}.
\]

Similar to \pref{lm:close1}, we have the following lemma.
\begin{lemma}\label{lm:close2}
It holds that
\[
\sum_{j\in[q(S,K)+2]}\sum_{k\in[K]}D_{\rm re}\prn*{ \bbP''_{j,k} \parallel\bbQ'_{j,k}} \le \frac{K}{8(q(S,K)+2)}.
\]
Combined with \pref{cor:first-second-class}, this implies
\[
\frac{1}{(q(S,K)+2)K}\sum_{j\in[q(S,K)+2]}\sum_{k\in[K]}D_{\rm re}\prn*{ \bbP''_{j,k} \parallel\bbQ''_{j,k}} \le \frac{1}{8(q(S,K)+2)^2}\le\frac{\wb{q^{(2)}}}{4}.
\]
\end{lemma}

\subsection{Applying the GRF Inequality}\label{app:apply}

In this part, we apply the \GRF inequality to show \pref{eq:obj1} and \pref{eq:obj2}, and complete the proof of \pref{thm:lb-unit}.

We first represent $\wb{p^{(1)}}$ using $\crl{\bbP_{j,k}'}$ and represent $\wb{q^{(1)}}$ using $\crl{\bbQ_{j,k}'}$. Since $\bbP_{j,k}'(E_{j,k}^{(1)})=\bbP_{j,k}^{(1)}(E_{j,k}^{(1)})$ and $\bbQ_{j,k}'(E_{j,k}^{(1)})=\bbQ(E_{j,k}^{(1)})$ hold for all $j,k$, we have
\begin{equation*}
    \wb{p^{(1)}}=\frac{1}{(q(S,K)+2)K}\sum_{j\in[q(S,K)+2]}\sum_{k\in[K]}\bbP_{j,k}^{(1)}\prn*{E_{j,k}^{(1)}}=\frac{1}{(q(S,K)+2)K}\sum_{j\in[q(S,K)+2]}\sum_{k\in[K]}\bbP'_{j,k}\prn*{E_{j,k}^{(1)}},
\end{equation*}
\begin{equation*}
    \wb{q^{(1)}}=\frac{1}{(q(S,K)+2)K}\sum_{j\in[q(S,K)+2]}\sum_{k\in[K]}\bbQ\prn*{E_{j,k}^{(2)}}=\frac{1}{(q(S,K)+2)K}\sum_{j\in[q(S,K)+2]}\sum_{k\in[K]}\bbQ'_{j,k}\prn*{E_{j,k}^{(1)}}.
\end{equation*}
By the \GRF inequality (\pref{prop:grf}), we have
\begin{align}\label{eq:lbp1}
    \wb{p^{(1)}}
    &\ge \wb{q^{(1)}}-\sqrt{2\wb{q^{(1)}}\cdot\frac{1}{(q(S,K)+2)K}\sum_{j\in[q(S,K)+2]}\sum_{k\in[K]}D_{\rm re}\prn*{ \bbP'_{j,k} \parallel\bbQ'_{j,k}}}\notag\\
    &\overset{\rm(i)}{\ge}  \wb{q^{(1)}}-\sqrt{2\wb{q^{(1)}}\frac{\wb{q^{(1)}}}{4}}\notag\\
    &=\frac{2-\sqrt{2}}{2}\wb{q^{(1)}}\notag\\
    &\overset{\rm(ii)}{\ge} \frac{2-\sqrt{2}}{4}\frac{K-r(S,K)}{(q(S,K)+2)^2K}
\end{align}
and
\begin{align}\label{eq:lbp2}
    \wb{p^{(2)}}
    &\ge \wb{q^{(2)}}-\sqrt{2\wb{q^{(2)}}\cdot\frac{1}{(q(S,K)+2)K}\sum_{j\in[q(S,K)+2]}\sum_{k\in[K]}D_{\rm re}\prn*{ \bbP''_{j,k} \parallel\bbQ''_{j,k}}}\notag\\
    &\overset{\rm(iii)}{\ge}  \wb{q^{(2)}}-\sqrt{2\wb{q^{(2)}}\frac{\wb{q^{(2)}}}{4}}\notag\\
    &=\frac{2-\sqrt{2}}{2}\wb{q^{(2)}}\notag\\
    &\overset{\rm(iv)}{\ge} \frac{2-\sqrt{2}}{4}\frac{1}{(q(S,K)+2)^2},
\end{align}
where (i) follows from \pref{lm:close1}, (ii) follows from \pref{cor:first-second-class}, (iii) follows from \pref{lm:close2}, and (iv) follows from \pref{cor:first-second-class}. We thus prove  \pref{eq:obj1} and \pref{eq:obj2}.

Now we plug \pref{eq:lbp1} and \pref{eq:lbp2} into \cref{lm:reduction} and \cref{lm:reduction2}. We have
\begin{align*}
    R^{\pi}(T)&\ge \frac{1}{\abs{\Phi_1}}\sum_{\bmu\in\Phi_1}R^{\pi}_{\bmu}(T)\\
    &\ge\cR_{\rm bad}(S,K,T)\cdot\frac{1}{(q(S,K)+2)K}\sum_{j\in[q(S,K)+2]}\sum_{k\in[K]}\bbP_{j,k}^{(1)}\prn*{E_{j,k}^{(1)}}\\
    &\ge\cR_{\rm bad}(S,K,T)\cdot\frac{2-\sqrt{2}}{4}\frac{K-r(S,K)}{(q(S,K)+2)^2K}\\
    &= \frac{\prn*{K-r(S,K)}}{8\prn{q(S,K)+2}}{\prn*{\frac{T}{{K-r(S,K)}}}^{\frac{1}{2-2^{-q(S,K)}}}}\cdot\frac{2-\sqrt{2}}{4}\frac{K-r(S,K)}{(q(S,K)+2)^2K}\\
    &=\frac{2-\sqrt{2}}{32(q(S,K)+2)^3}\frac{\prn{K-r(S,K)}^{2-\frac{1}{2-2^{-q(S,K)}}}}{K}T^{\frac{1}{2-2^{-q(S,K)}}}\\
    &\ge\frac{2-\sqrt{2}}{32(\log_2\log_2(T/K))^3}\frac{\prn{K-r(S,K)}^{2-\frac{1}{2-2^{-q(S,K)}}}}{K}T^{\frac{1}{2-2^{-q(S,K)}}}\\
    &=\frac{2-\sqrt{2}}{160\log_2(T/K)}\frac{\prn{K-r(S,K)}^{2-\frac{1}{2-2^{-q(S,K)}}}}{K}T^{\frac{1}{2-2^{-q(S,K)}}}
\end{align*}
and
\begin{align*}
    R^{\pi}(T)&\ge\frac{1}{\abs{\Phi_2}}\sum_{\bmu\in\Phi_2}R_{\bmu}^\pi(T)\\
    &\ge\cR_{\rm bad2}(S,K,T)\cdot\frac{1}{(q(S,K)+2)K}\sum_{j\in[q(S,K)+2]}\sum_{k\in[K]}\bbP_{j,k}^{(2)}\prn*{E_{j,k}^{(2)}}\\
    & \ge \cR_{\rm bad2}(S,K,T)\cdot\frac{2-\sqrt{2}}{4}\frac{1}{(q(S,K)+2)^2}\\
    &=\frac{{K}}{4\prn{q(S,K)+2}}{\prn*{\frac{T}{{K}}}^{\frac{1}{2-2^{-q(S,K)-1}}}}\cdot\frac{2-\sqrt{2}}{4}\frac{1}{(q(S,K)+2)^2}\\
    &=\frac{2-\sqrt{2}}{16\prn{q(S,K)+2}^3}K^{1-{\frac{1}{2-2^{-q(S,K)-1}}}}T^{\frac{1}{2-2^{-q(S,K)-1}}}\\
    &\ge\frac{2-\sqrt{2}}{16\prn{\log_2\log_2(T/K)}^3}K^{1-{\frac{1}{2-2^{-q(S,K)-1}}}}T^{\frac{1}{2-2^{-q(S,K)-1}}}\\
    &\ge\frac{2-\sqrt{2}}{80\log_2(T/K)}K^{1-{\frac{1}{2-2^{-q(S,K)-1}}}}T^{\frac{1}{2-2^{-q(S,K)-1}}}.
\end{align*}
We thus complete the proof of \pref{thm:lb-unit}. $\hfill\Box$

\section{Proof of \pref{thm:glb}}\label{app:lb-gbwcs}

The proof of \pref{thm:glb} builds on the proof of \pref{thm:lb-unit}. In this proof, we emphasize the differences, which are mainly in  \pref{app:REdefG,app:combg,app:altg}, corresponding to the first three steps of the \lbargu method.

Given any $K>1$, $S\ge0$ and $T\ge 2K$, we focus on the setting of $\cD_k=\mathcal{N}(\mu_k,1)$ ($\forall k\in[K]$), as this is sufficient for us to prove the desired lower bound.  For simplicity, in this proof we will directly use the vector $\bmu$ to represent the environment.

For any environment $\bmu$, let $X_{\bmu}^{t}(k)\sim\mathcal{N}(\mu_k,1)$ denote the i.i.d. random reward of each action $k$ at round $t$ ($k\in[K],t\in[T]$). 
For any policy $\pi\in\Pi_S$, for any environment $\bmu$, for any $t\in[T]$, we use $a_t$ to denote the random action selected by policy $\pi$ at round $t$ under environment $\bmu$, and use $X^t_{\bmu}(a_t)$ to denote the random reward observed by policy $\pi$ at round $t$ under environment $\bmu$. Let $\cH_t\ldef\prn*{\prn*{a_1,X_{\bmu}^1(a_1)},\dots,\prn*{a_t,X_{\bmu}^t(a_t)}}$ be the history (of actions and observations) up to round $t$ (inclusive), whose value lies in $\Omega_t\ldef\prn*{[K]\times\bbR}^t$. Let $\cF_t\ldef\cB(\Omega_t)$ be the Borel $\sigma$-algebra on $\Omega_t$. Let $\bbP_{\bmu}^{\pi}$ be the probability measure induced by (i.e., the joint distribution of) $\cH_T$, and $\bbE_{\bmu}^\pi$ be the associated expectation operator. Let $R_{\bmu}^{\pi}(T)\ldef T\mu^*-\bbE_{\bmu}^{\pi}\brk*{\sum_{t=1}^T{\mu}_{a_t}}$ be policy $\pi$'s distribution-dependent regret under environment $\bmu$.

Similar to \pref{app:lower}, we argue that in our proof,  we only need to consider the case of $q''(S,G)+2\le \log_2\log_2(T)$. 
Suppose $q''(S,K)+2>\log_2\log_2(T)$, then we have 
\begin{align*}
T^{\frac{1}{2-2^{-q''(S,G)}}}\le\sqrt{2T},
\end{align*}
thus the lower bound in \pref{thm:glb} becomes $\Omega(\sqrt{T}/(K\log T))$ and can be directly obtained by applying the well-known $\Omega(\sqrt{KT})$ lower bound of the classical \MAB (see, e.g., \citealt{lattimore2020bandit}). Therefore, the really non-trivial case of \pref{thm:glb} is the case of $q''(S,G)+2\le \log_2\log_2(T)$, and we focus on this case in the rest of our proof.

Our goal is to explicitly construct a family of environments $\Phi$, such that for any $S$-switching-budget policy $\pi\in\Pi_S$, the ``worst-case regret'' $\max_{\bmu\in\Phi}R_{\bmu}^{\pi}(T)$ is lower bounded by
\begin{equation*}
    {\Omega}\prn*{\frac{1}{K\log T}T^{\frac{1}{2-2^{-q''(S,G)}}}}
\end{equation*}
Since the worst-case regret $R^{\pi}(T)$ is no less than the  $\max_{\bmu\in\Phi}R_{\bmu}^{\pi}(T)$, the above goal directly implies \pref{thm:glb}. 

Without loss of generality, we assume that $1\in\arg\max_{i\in[K]}\min_{j\ne i}c_{i,j}$. For notational simplicity, for all $j\in[q''(S,G)+1]$, we \textbf{redefine} the sequence $(t_j)_{j=0}^{q''(S,G)+1}$ as $t_0=0$ and
\begin{equation}\label{eq:redefine}
    t_j=\left\lfloor{T}^{\frac{2-2^{-(i-1)}}{2-2^{-q''(S,G)}}}\right\rfloor,~~\forall j=1,\dots,q''(S,G)+1.
\end{equation}
Note that the above definition is different from the definition of $(t_j)_{j=0}^{q'(S,G)+1}$ in \pref{alg:hsse} --- we only use the above definition in this proof, for the purpose of simplifying notations.

\subsection{Definitions of Risky Events}\label{app:REdefG}

For any policy $\pi\in\Pi_S$, for any environment ${\bmu}$, we make some key definitions below. 

1. For any $n_1,n_2\in[T]$, we define a random variable $S(n_1,n_2)$ to be the  total switching cost incurred in period $[n_1:n_2]$ (note that if there is a switch happening between round $n_1-1$ and round $n_1$, or between round $n_2$ and round $n_2+1$, we do not count its cost in $S(n_1,n_2)$).

2. Second, we define a stopping time 
\[
\tau\ldef\min\crl*{t\in[T]: \text{all of the actions in }[K]\text{ are chosen in period }[t_{q''(S,G)}:\tau]}
\]
if the set is non-empty and $\tau=\infty$ otherwise. %
Note that this definition is different from the definition used in \pref{app:REdef}. The idea of such ``tracking the covering time'' definition of $\tau$ originates in the preliminary version of this paper (\citealt{simchi2019phase}).

3. We define a class of \emph{risky} events as follows: for any $k\in[K]$, let

$E_{1,k}\ldef\crl*{\text{action }k \text{ is not chosen in period }\brk*{1:t_{1}}},$

$E_{j,k}\ldef\crl*{\text{action }k \text{ is not chosen in period }\brk*{t_{j-1}:t_{j}}},~~\forall j\in[2:q''(S,G)],$

$E_{q''(S,G)+1,k}\ldef\crl*{\text{action }k \text{ is not chosen in period }\brk*{t_{q''(S,G)}:\floor{\prn{t_{q''(S,G)}+T}/2}}},$

$E_{q''(S,G)+2,k}\ldef\crl*{\tau\le \floor{\prn{t_{q''(S,G)}+T}/2},\, a_\tau=k, \,S(1:t_{q''(S,G)})\ge q''(S,G)H}.$\\
By doing so, we get $\prn*{q''(S,G)+2}K$ risky events (of the form $E_{j,k}$) in total. Note that the time points $\prn{t_j}_{j=1}^{q''(S,G)+1}$ are fixed and given in \pref{eq:redefine}, and the events $\prn{E_{q''(S,G)+2,k}}_{k\in[K]}$ are defined based on the stopping time $\tau$. %

\subsection{Combinatorial Arguments and Lower Bounds for Risky Events (under a Single Environment)}\label{app:combg}
The main purpose of this subsection is to prove the following lemma using (non-trivial) combinatorial (and probabilistic) arguments. Compared with the second step in the proof of \pref{thm:lb-unit}, we develop new techniques here to deal with the general switching cost structure, while paying less attention to the order of $K$.

\begin{lemma}\label{lm:first-classg}For any policy $\pi\in\Pi_S$, for any environment $\bmu$, we have
\[
\sum_{j\in[q''(S,G)+2]}\sum_{k\in[K]}\bbP_{\bmu}^{\pi}\prn*{E_{j,k}}\ge 1.
\]
\end{lemma}

\cref{lm:first-classg}  leads to the following corollary, which will be utilized in subsequent subsections.

\begin{corollary}
\label{cor:first-second-classg}For any policy $\pi\in\Pi_S$, for any environment $\bmu$, we have
\[
\frac{1}{(q''(S,G)+2)K}\sum_{j\in[q''(S,G)+2]}\sum_{k\in[K]}\bbP_{\bmu}^{\pi}\prn*{E_{j,k}}\ge \frac{1}{(q''(S,G)+2)K},
\]
\end{corollary}

\cref{cor:first-second-classg} tells us the following fact: under any \emph{single} environment $\bmu$, the average probability of the  risky events is $\wt{\Omega}\prn*{\frac{1}{K}}$.

In the rest of this subsection, we provide a proof for \cref{lm:first-classg}.

\proof{Proof of \pref{lm:first-class}.} Since $H$ is the total weight of the shortest Hamiltonian path of $G$, for any $j\in[q''(S,G)]$, we have
\begin{align*}
\sum_{k\in[K]}\Ind\crl*{E_{j,k}}&=\text{number of actions that are not chosen in period }\brk*{t_{j-1}:t_{j}}\\
&\ge \Ind\crl*{\text{not all actions are chosen in period }\brk*{t_{j-1}:t_{j}}}\\
&\ge \Ind\crl*{S\prn*{t_{j-1}:t_j}<H}
\end{align*}
almost surely. Thus for any $j\in[q''(S,G)]$, we have
\begin{equation}
\begin{aligned}[b]\label{eq:lm6g}
\sum_{k\in[K]}\bbP_{\bmu}^{\pi}\prn*{E_{j,k}}& = \sum_{k\in[K]}\bbE_{\bmu}^{\pi}\brk*{\Ind\crl*{E_{j,k}}}\\
& =  \bbE_{\bmu}^{\pi}\brk*{\sum_{k\in[K]}\Ind\crl*{E_{j,k}}}\\
& \ge \bbE_{\bmu}^{\pi}\brk*{\Ind\crl*{S\prn*{t_{j-1}:t_j}<H}}.
\end{aligned}
\end{equation}
Summing \pref{eq:lm6g} over $j\in[q''(S,G)]$, we have
\begin{align*}
\sum_{j\in[q''(S,G)]}\sum_{k\in[K]}\bbP_{\bmu}^{\pi}\prn*{E_{j,k}}& \ge \bbE_{\bmu}^\pi\brk*{ \sum_{j\in[q''(S,G)]}\Ind\crl*{S\prn*{t_{j-1}:t_j}<H}}\\
&\overset{\rm(i)}{\ge}\bbE_{\bmu}^\pi\brk*{\prn*{1- \Ind\crl*{S\prn*{1:t_{q''(S,G)}}\ge {q''(S,G)}H}}}.
\end{align*}
Note that (i) follows from
\begin{align*}
\sum_{j\in[q''(S,G)]}\Ind\crl*{S\prn*{t_{j-1}:t_j}<H}&\ge\Ind\crl*{\bigcup_{j\in[q''(S,G)]}\crl*{S\prn*{t_{j-1}:t_j}<H}}\\
&\overset{\rm(ii)}{\ge}\Ind\crl*{S\prn*{1:t_{q''(S,G)}}< q''(S,G)H}\\
&=1-\Ind\crl*{S\prn*{1:t_{q''(S,G)}}\ge q''(S,G)H},
\end{align*}
where (ii) follows from the pigeonhole principle.

Now we define
\begin{align*}
E_{\sim,k} \ldef   \crl*{\text{action }k\text{ is not among the first }K-1\text{ (different) actions chosen in period }\brk*{t_{q''(S,G)}:T}}.
\end{align*}
If both $\crl*{S\prn*{1:t_{q''(S,G)}}\ge {q''(S,G)}H}$ and $ E_{\sim,k} $ happen, then since $\tau$ is the first time that all actions of $[K]$ have been chosen after round $t_{q''(S,G)}$, we know that  
either $E_{q''(S,G)+1,k}=\crl*{\text{action }k \text{ is not chosen in period }\brk*{t_{q''(S,G)}:\floor{\prn{t_{q''(S,G)}+T}/2}}}$ happens, or $E_{q''(S,G)+2,k}=\crl*{\tau\le \floor{\prn{t_{q''(S,G)}+T}/2},\, a_\tau=k, \, S\prn*{1:t_{q''(S,G)}}\ge q''(S,G)H}$ happens.
Therefore, we know that
\[
E_{q''(S,G)+1,k}\cup E_{q''(S,G)+2,k} \ \supset\   E_{\sim,k} \cap \crl*{S\prn*{1:t_{q''(S,G)}}\ge {q''(S,G)}H}.
\]
This implies that
\begin{align*}
    \sum_{k\in[K]}\bbP_{\bmu}^{\pi}\prn*{E_{q''(S,G)+1,k}\cup E_{q''(S,G)+2,k} }& \ge\sum_{k\in[K]}\bbP_{\bmu}^{\pi}\prn*{E_{\sim,k} \cap \crl*{S\prn*{1:t_{q''(S,G)}}\ge {q''(S,G)}H}}\\
    &= \sum_{k\in[K]}\bbE_{\bmu}^\pi\brk*{\Ind\crl*{E_{\sim,k}}\Ind{\crl*{S\prn*{1:t_{q''(S,G)}}\ge {q''(S,G)}H}}} \\
    & =\bbE_{\bmu}^{\pi}\brk*{\sum_{k\in[K]}\Ind\crl*{E_{\sim,k}}\Ind{\crl*{S\prn*{1:t_{q''(S,G)}}\ge{q''(S,G)}H}}}\\
    &\overset{\rm (iii)}{\ge} \bbE_{\bmu}^{\pi}\brk*{\Ind{\crl*{S\prn*{1:t_{q''(S,G)}}\ge {q''(S,G)}H}}},
\end{align*}
where (iii) follow from the  definition of $E_{\sim,k}$.

Combining the above two paragraphs, we have
\begin{align*}
    \sum_{j\in[q''(S,G)+2]}\sum_{k\in[K]}\bbP_{\bmu}^{\pi}\prn*{E_{j,k}}
    &\ge\sum_{j\in[q''(S,G)]}\sum_{k\in[K]}\bbP_{\bmu}^{\pi}\prn*{E_{j,k}}+   \sum_{k\in[K]}\bbP_{\bmu}^{\pi}\prn*{E_{q''(S,G)+1,k}\cup E_{q''(S,G)+2,k} }\\
    &\ge\bbE_{\bmu}^\pi\brk*{\prn*{1- \Ind\crl*{S\prn*{1:t_{q''(S,G)}}\ge {q''(S,G)}H}}}\\
    &~~~+\bbE_{\bmu}^{\pi}\brk*{\Ind{\crl*{S\prn*{1:t_{q''(S,G)}}\ge {q''(S,G)}H}}}\\
    &=1.
\end{align*}
$\hfill\Box$

\subsection{Alternative Environments, Bad Events, and Lower Bound Reductions}\label{app:altg}

Compared with the proof of \pref{thm:lb-unit}, the main difference in this step is that we define the {reference} and alternative environments in a new way. In particular, we make action 1 the \emph{unique} optimal action in the reference environment, and let all the alternative environments in the form of $\bmu_{q''(S,G)+2,k}$ ($k\ne 1$) be the same as the reference environments. Since any switch from or to action 1 incurs a cost at least $\max_{i}\min_{j\ne i}c_{i,j}$, such new techniques help us to make the quantity $\max_{i}\min_{j\ne i}c_{i,j}$ appear in the lower bound.

\underline{In the rest of the proof, we fix an arbitrary policy $\pi\in\Pi_S$.}

 For any $j\in[q''(S,G)+2]$, define a reward gap
\[
\Delta_{j}\ldef\begin{cases}
1, &\text{if }j=1,\\
\frac{1}{{2(q''(S,G)+2)}}\sqrt{\frac{{1}}{{t_{j-1}}}}, &\text{if }j\in[2:q''(S,G)+1],\\
-\frac{1}{{2(q''(S,G)+2)}}\sqrt{\frac{{1}}{{t_{q''(S,G)}}}}, &\text{if }j=q''(S,G)+2.
\end{cases}
\]
Note that $\abs*{\Delta_j}\in[0,1]$ for all $j\in[q''(S,G)+2]$.

Let $\alpha=(\frac{\Delta_{q''(S,G)+1}}{2},0,0,\dots,0)\in\bbR^K$ be the \emph{reference} environment. Let $\bbQ\ldef\bbP_{\alpha}^\pi$ denote the \emph{reference} measure.

For any $j\in[q''(S,G)+1], k\in[K]$, define an \emph{alternative} environment $\bmu_{j,k}\ldef \prn*{\mu_{j,k;1},\dots,\mu_{j,k;K}}\in\bbR^K$ where
\[
\mu_{j,k;i}\ldef\begin{cases}\alpha_i+\Delta_j, &\text{if }i=k,\\
\alpha_i, &\text{otherwise.}
\end{cases}
\]
Note that each alternative environment $\bmu_{j,k}$ define above only differs from the reference environment in terms of the mean reward of action $k$.

For $j=q''(S,G)+2$, for any $k\ne 1$, define an \emph{alternative} environment $\bmu_{q''(S,G)+2,k}\ldef\alpha$ which is the same as the reference environment. For $j=q''(S,G)+2$ and $k=1$, define an \emph{alternative} environment 
$\bmu_{q''(S,G)+2,1}\ldef \prn*{\mu_{q''(S,G)+2,1;1},\dots,\mu_{q''(S,G)+2,1;K}}\in\bbR^K$ where
\[
\mu_{q''(S,G)+2,1;i}\ldef\begin{cases}\alpha_i+\Delta_j, &\text{if }i=1,\\
\alpha_i, &\text{otherwise.}
\end{cases}
\]
Note that the above definitions are different from those in the proof of \pref{thm:lb-unit}.

For any  $j\in[q''(S,G)+2], k\in[K]$, let $\bbP_{j,k}\ldef\bbP^{\pi}_{\bmu_{j,k}}$ denote the \emph{alternative} measure associated with the alternative environment $\bmu_{j,k}$. 

We explicitly construct a class of environments $\Phi\ldef\crl*{\bmu_{j,k}\mid j\in[q''(S,G)+2], k\in[K]}$.

For any  $j\in[q''(S,G)+2], k\in[K]$, under environment $\bmu_{j,k}$, the risky event $E_{j,k}$ becomes a \emph{bad event} whose occurrence would lead to large regret. Specifically:
\begin{itemize}
    \item Suppose $j=1$. Since action $k$ is the unique optimal action under environment $\bmu_{1,k}$, choosing any action other than $k$ for one round incurs at least a $\Delta_1-\frac{\Delta_{q''(S,G)+1}}{2}\ge\frac{\Delta_1}{2}$ term in the policy's  regret, and the occurrence of $E_{1,k}=\crl*{\text{action }k \text{ is not chosen in period }\brk*{1:t_{1}}}$ incurs at least a ${t_1}\Delta_1/2$ term in the policy's  regret.
    \item Suppose $j\in[2:q''(S,G)]$. Since action $k$ is the unique optimal action under environment $\bmu_{j,k}$, choosing any action other than $k$ for one round incurs at least a $\Delta_j-\frac{\Delta_{q''(S,G)+1}}{2}\ge\frac{\Delta_j}{2}$ term in the policy's  regret, and the occurrence of $E_{j,k}=\crl*{\text{action }k \text{ is not chosen in period }\brk*{t_{j-1}:t_{j}}}$ incurs at least a $\prn*{t_j-t_{j-1}+1}\Delta_j/2$ term in the policy's  regret.
    \item Suppose $j=q''(S,G)+1$. Since action $k$ is the unique optimal action under environment $\bmu_{q''(S,G)+1,k}$, choosing any action other than $k$ for one round incurs at least a $\Delta_{q''(S,G)+1}-\frac{\Delta_{q''(S,G)+1}}{2}\ge\frac{\Delta_{q''(S,G)+1}}{2}$ term in the policy's  regret, and the occurrence of $E_{q''(S,G)+1,k}=\crl*{\text{action }k \text{ is not chosen in period }\brk*{t_{q''(S,G)}:\floor{\prn{t_{q''(S,G)}+T}/2}}}$ incurs at least a $\prn*{\floor{\prn{t_{q''(S,G)}+T}/2}-t_{q''(S,G)}+1}\Delta_{q''(S,G)+1}/2$ term in the policy's  regret.
    \item Suppose $j=q''(S,G)+2$ and $k=1$. Since action 1 is the worst action under environment $\bmu_{q''(S,G)+2,1}$, choosing action 1 for one round incurs at least a $-\Delta_{q''(S,G)+2}-\frac{\Delta_{q''(S,G)+1}}{2}=-\frac{\Delta_{q''(S,G)+2}}{2}$ term in the policy's regret. Furthermore, by the switching constraint $S(1:T)\le S<(q''(S,G)+1)H+\min_{j\ne 1}c_{1,j}$, 
    the occurrence of $E_{q''(S,G)+2,1}$ implies the occurrence of $\crl*{\text{no switch happens after round } \tau}$ (as the remaining switching budget does not allow for switching from action 1), thus essentially implies the occurrence of  $\crl*{\text{action }1\text{ is chosen in every round in period }[\floor{\prn{t_{q''(S,G)}+T}/2}:T]}$. As a result, the occurrence of  $E_{q''(S,G)+2,1}$ incurs at least a $-\prn*{T-\floor{\prn{t_{q''(S,G)}+T}/2}+1}\Delta_{q''(S,G)+2}/2$ term in the policy's  regret.
    \item Suppose $j=q''(S,G)+2$ and $k\ne 1$. Since action 1 is the unique optimal action under environment $\bmu_{q''(S,G)+2,k}=\alpha$, choosing action $k\ne 1$ for one round incurs at least a $\frac{\Delta_{q''(S,G)+1}}{2}$ term in the policy's regret. Furthermore, by the switching constraint $S(1:T)\le S<(q''(S,G)+1)H+\min_{j\ne 1}c_{j,1}$ (we utilize the symmetry of switching costs here), 
    the occurrence of $E_{q''(S,G)+2,1}$ implies the occurrence of $\crl*{\text{action 1 is never chosen after round } \tau}$ (as the remaining switching budget does not allow for switching to action 1), thus essentially implies the occurrence of  $\crl*{\text{action }1\text{ is not chosen in period }[\floor{\prn{t_{q''(S,G)}+T}/2}:T]}$. As a result, the occurrence of  $E_{q''(S,G)+2,1}$ incurs at least a $\prn*{T-\floor{\prn{t_{q''(S,G)}+T}/2}+1}\Delta_{q''(S,G)+1}/2=-\prn*{T-\floor{\prn{t_{q''(S,G)}+T}/2}+1}\Delta_{q''(S,G)+2}/2$ term in the policy's  regret.
\end{itemize}
Note that the arguments in the last two bullets are very different from what we have done in the proof of \pref{thm:lb-unit}. 
The above arguments lead to \cref{lm:badg}.

\begin{lemma}[From risky events to bad events]\label{lm:badg} For any $j\in[q''(S,G)+2], k\in[K]$, under environment $\bmu_{j,k}$, the risky event $E_{j,k}$ becomes a \emph{bad event} in the sense that
\[\bbE_{\bmu_{j,k}}^{\pi}\brk*{T\mu_{j,k;k}-\sum_{t=1}^T\mu_{j,k;a_t} \mid E_{j,k}}\ge \cR_{\rm bad}\prn{S,G,T},\]
where \[\cR_{\rm bad}(S,G,T)\ldef\frac{{1}}{16\prn{q''(S,G)+2}}{{{T}}^{\frac{1}{2-2^{-q''(S,G)}}}}\]
is a universal lower bound on the ``distribution-dependent regret conditional on the bad event.''
\end{lemma}

\proof{Proof of \cref{lm:badg}.} By the arguments in the previous paragraph, we have
\begin{align*}
\bbE_{\bmu_{j,k}}^{\pi}\brk*{T\mu_{j,k;k}-\sum_{t=1}^T\mu_{j,k;a_t} \mid E_{j,k}}
&\ge\bbE_{\bmu_{j,k}}^{\pi}\brk*{\prn*{t_j-t_{j-1}^{(2)}+1}\mu_{j,k;k}-\sum_{t\in \brk*{t_{j-1}:t_{j}}} \mu_{j,k;a_t} \mid E_{j,k}}\\
&\ge    \begin{cases}
        t_1\Delta_{1}/2, &\text{if }j=1,\\
        \prn*{t_j-t_{j-1}+1}\Delta_{j}/2, &\text{if }j\in[2:q''(S,G)],\\
        \prn*{\floor{\prn{t_{q''(S,G)}+T}/2}-t_{q''(S,G)}+1}\Delta_{q''(S,G)+1}/2, &\text{if }j=q''(S,G)+1,\\
        -\prn*{T-\floor{\prn{t_{q''(S,G)}+T}/2}+1}\Delta_{q''(S,G)+2}/2, &\text{if }j=q''(S,G)+2,
        \end{cases}\\
&\ge \frac{1}{16(q''(S,G)+2)}{{{T}}^{\frac{1}{2-2^{-q''(S,G)}}}},
\end{align*}
where the last inequality follows from the same algebra presented in the proof of \pref{lm:bad}.
$\hfill\Box$

Based on \cref{lm:badg}, we can reduce the task of proving a lower bound on the policy's (distribution-dependent) regret $R_{\bmu_{j,k}}^{\pi}(T)$ to the task of proving a lower bound on the bad event probability $\bbP_{j,k}\prn*{E_{j,k}}$. Consequently, we can reduce the task of proving a lower bound on  $\sup_{\bmu\in\Phi}R^{\pi}_{\bmu}(T)$ to the task of proving a lower bound on the ``average-case bad event probability'' $\frac{1}{(q''(S,G)+2)K}\sum_{j\in[q''(S,G)+2]}\sum_{k\in[K]}\bbP_{j,k}\prn*{E_{j,k}}$.
\begin{lemma}[Reducing regret lower bounds to bad event probability lower bounds]\label{lm:reductiong} For any $j\in[q''(S,G)+2], k\in[K]$, we have
\[
R_{\bmu_{j,k}}^{\pi}(T)\ge \cR_{\rm bad}(S,G,T)\cdot \bbP_{j,k}\prn*{E_{j,k}}.
\]
As a result, we have
\begin{align*}
    R^{\pi}(T)\ge \sup_{\bmu\in\Phi}R^{\pi}_{\bmu}(T)\ge\cR_{\rm bad}(S,K,T)\cdot\frac{1}{(q''(S,G)+2)K}\sum_{j\in[q''(S,G)+2]}\sum_{k\in[K]}\bbP_{j,k}\prn*{E_{j,k}}.
\end{align*}
\end{lemma}
\proof{Proof of \cref{lm:reductiong}.}
The proof is almost the same as the proof of \pref{lm:reduction}.
$\hfill\Box$

\subsection{Probability Space Changing Tricks}\label{sec:changingg}
\cref{lm:reductiong} indicates that, in order to prove the desired lower bound on the regret $R^\pi(T)$, it suffices to prove the following statement:
\begin{equation}\label{eq:obj1g}
    \wb{p}\ldef\frac{1}{(q''(S,G)+2)K}\sum_{j\in[q''(S,K)+2]}\sum_{k\in[K]}\bbP_{j,k}\prn*{E_{j,k}}=\wt{\Omega}\prn*{\frac{1}{K}},
\end{equation}
That is, we only need to establish tight lower bounds on the average probability $  \wb{p}$.

By \cref{cor:first-second-classg}, we have
\[
\wb{q}\ldef\frac{1}{(q(S,K)+2)K}\sum_{j\in[q(S,K)+2]}\sum_{k\in[K]}\bbQ\prn*{E_{j,k}}\ge\frac{1}{(q(S,K)+2)K}=\wt{\Omega}\prn*{\frac{1}{K}},
\]
Therefore, in order to show \cref{eq:obj1g}, it suffices to show that $\wb{p}$ is close to $\wb{q}$. Similar to the proof of \pref{thm:lb-unit}, we apply the probability space changing tricks. The arguments are very similar to the arguments in \pref{sec:changing} and are omitted here.

\subsection{Applying the GRF Inequality}\label{app:applyg}

Similar to  \pref{app:apply}, by applying the \GRF inequality, we can show that
\begin{align}\label{eq:lbp1g}
    \wb{p}
    &{\ge}  \wb{q}-\sqrt{2\wb{q}\frac{\wb{q}}{4}}\notag\\
    &=\frac{2-\sqrt{2}}{2}\wb{q}.
\end{align}
Now we plug \pref{eq:lbp1g}  into \cref{lm:reductiong}. We have
\begin{align*}
    R^{\pi}(T)&\ge \sup_{\bmu\in\Phi}R^{\pi}_{\bmu}(T)\\
    &\ge\cR_{\rm bad}(S,G,T)\cdot\frac{1}{(q(S,K)+2)K}\sum_{j\in[q(S,K)+2]}\sum_{k\in[K]}\bbP_{j,k}\prn*{E_{j,k}}\\
    &\ge\cR_{\rm bad}(S,G,T)\cdot\frac{2-\sqrt{2}}{2}\wb{q}\\
    &= \frac{1}{16\prn{q''(S,G)+2}}{{T}^{\frac{1}{2-2^{-q''(S,G)}}}}\cdot\frac{2-\sqrt{2}}{2}\frac{1}{(q(S,K)+2)K}\\
    &=\Omega\prn*{\frac{1}{K\log T}T^{\frac{1}{2-2^{-q''(S,G)}}}}.
\end{align*}
We thus complete the proof of \pref{thm:glb}. $\hfill\Box$

\section{Proof of the Lower Bound in \pref{thm:asym}}\label{app:lb-dbwsc}
The proof of the lower bound builds on the proof of \pref{thm:lb-unit}. In this proof, we emphasize the differences, which are mainly in  \pref{app:REdefd,app:combd}, corresponding to the first two steps of the \lbargu method.

Given any $K>1$, $S\ge0$ and $T\ge 2K$, we focus on the setting of $\cD_k=\mathcal{N}(\mu_k,1)$ ($\forall k\in[K]$), as this is sufficient for us to prove the desired lower bound.  For simplicity, in this proof we will directly use the vector $\bmu$ to represent the environment.

For any environment $\bmu$, let $X_{\bmu}^{t}(k)\sim\mathcal{N}(\mu_k,1)$ denote the i.i.d. random reward of each action $k$ at round $t$ ($k\in[K],t\in[T]$). 
For any policy $\pi\in\Pi_S$, for any environment $\bmu$, for any $t\in[T]$, we use $a_t$ to denote the random action selected by policy $\pi$ at round $t$ under environment $\bmu$, and use $X^t_{\bmu}(a_t)$ to denote the random reward observed by policy $\pi$ at round $t$ under environment $\bmu$. Let $\cH_t\ldef\prn*{\prn*{a_1,X_{\bmu}^1(a_1)},\dots,\prn*{a_t,X_{\bmu}^t(a_t)}}$ be the history (of actions and observations) up to round $t$ (inclusive), whose value lies in $\Omega_t\ldef\prn*{[K]\times\bbR}^t$. Let $\cF_t\ldef\cB(\Omega_t)$ be the Borel $\sigma$-algebra on $\Omega_t$. Let $\bbP_{\bmu}^{\pi}$ be the probability measure induced by (i.e., the joint distribution of) $\cH_T$, and $\bbE_{\bmu}^\pi$ be the associated expectation operator. Let $R_{\bmu}^{\pi}(T)\ldef T\mu^*-\bbE_{\bmu}^{\pi}\brk*{\sum_{t=1}^T{\mu}_{a_t}}$ be policy $\pi$'s distribution-dependent regret under environment $\bmu$.

Similar to \pref{app:lower}, we argue that in our proof,  we only need to consider the case of $q(S,\bm{c})+2\le \log_2\log_2(T)$. 
Suppose $q(S,\bm{c})+2>\log_2\log_2(T)$, then we have 
\begin{align*}
T^{\frac{1}{2-2^{-q(S,\bm{c})}}}\le\sqrt{2T},
\end{align*}
thus the lower bound in \pref{thm:glb} becomes $\wt\Omega(\sqrt{T})$ and can be directly obtained by applying the well-known $\Omega(\sqrt{KT})$ lower bound of the classical \MAB (see, e.g., \citealt{lattimore2020bandit}). Therefore, the really non-trivial case of \pref{thm:glb} is the case of $q(S,\bm{c})+2\le \log_2\log_2(T)$, and we focus on this case in the rest of our proof.

Our goal is to explicitly construct a family of environments $\Phi$, such that for any $S$-switching-budget policy $\pi\in\Pi_S$, the ``worst-case regret'' $\max_{\bmu\in\Phi}R_{\bmu}^{\pi}(T)$ is lower bounded by
\begin{equation*}
    {\Omega}\prn*{\frac{1}{K\log T}T^{\frac{1}{2-2^{-q(S,\bm{c})}}}}
\end{equation*}
Since the worst-case regret $R^{\pi}(T)$ is no less than the  $\max_{\bmu\in\Phi}R_{\bmu}^{\pi}(T)$, the above goal directly implies \pref{thm:glb}. 

Without loss of generality, we assume that $1\in\arg\max_{i\in[K]}\min_{j\ne i}c_{i,j}$. For notational simplicity, for all $j\in[q(S,\bm{c})+1]$, we \textbf{redefine} the sequence $(t_j)_{j=0}^{q(S,\bm{c})+1}$ as $t_0=0$ and
\begin{equation}\label{eq:redefined}
    t_j=\left\lfloor{T}^{\frac{2-2^{-(j-1)}}{2-2^{-q(S,\bm{c})}}}\right\rfloor,~~\forall j=1,\dots,q(S,\bm{c})+1.
\end{equation}
Note that the above definition is different from the definition of $(t_j)_{j=0}^{q(S,\bm{c})+1}$ in \pref{alg:asse} --- we only use the above definition in this proof, for the purpose of simplifying notations.

\subsection{Definitions of Risky Events}\label{app:REdefd}

For any policy $\pi\in\Pi_S$, for any environment ${\bmu}$, we make some key definitions below. 

1. For any $n_1,n_2\in[T]$, we define a random variable $S(n_1,n_2)$ to be the  total switching cost incurred in period $[n_1:n_2]$ (note that if there is a switch happening between round $n_1-1$ and round $n_1$, or between round $n_2$ and round $n_2+1$, we do not count its cost in $S(n_1,n_2)$).

2. Second, we define a stopping time 
\[
\tau\ldef\min\crl*{t\in[T]: \text{all of the actions in }[K]\text{ are chosen in period }[t_{q(S,\bm{c})}:\tau]}
\]
if the set is non-empty and $\tau=\infty$ otherwise. %
Note that this definition is different from the definition used in \pref{app:REdef}. The idea of such ``tracking the covering time'' definition of $\tau$ originates in the preliminary version of this paper (\citealt{simchi2019phase}).

3. We define a class of \emph{risky} events as follows: for any $k\in[K]$, let

$E_{1,k}\ldef\crl*{\text{action }k \text{ is not chosen in period }\brk*{1:t_{1}}},$

$E_{j,k}\ldef\crl*{\text{action }k \text{ is not chosen in period }\brk*{t_{j-1}:t_{j}}},~~\forall j\in[2:q(S,\bm{c})],$

$E_{q(S,\bm{c})+1,k}\ldef\crl*{\text{action }k \text{ is not chosen in period }\brk*{t_{q(S,\bm{c})}:\floor{\prn{t_{q(S,\bm{c})}+T}/2}}},$

$E_{q(S,\bm{c})+2,k}\ldef\crl*{\tau\le \floor{\prn{t_{q(S,\bm{c})}+T}/2},\, a_\tau=k, \,S(1:t_{q(S,\bm{c})})> S-\Sigma}.$\\
By doing so, we get $\prn*{q(S,\bm{c})+2}K$ risky events (of the form $E_{j,k}$) in total. Note that the time points $\prn{t_j}_{j=1}^{q(S,\bm{c})+1}$ are fixed and given in \pref{eq:redefined}, and the events $\prn{E_{q(S,\bm{c})+2,k}}_{k\in[K]}$ are defined based on the stopping time $\tau$. %

\subsection{Combinatorial Arguments and Lower Bounds for Risky Events (under a Single Environment)}\label{app:combd}
The main purpose of this subsection is to prove the following lemma using (non-trivial) combinatorial (and probabilistic) arguments. Compared with the second step in the proof of \pref{thm:lb-unit}, we develop new techniques here to deal with the departure cost structure, while paying less attention to the order of $K$.

\begin{lemma}\label{lm:first-classd}For any policy $\pi\in\Pi_S$, for any environment $\bmu$, we have
\[
\sum_{j\in[q(S,\bm{c})+2]}\sum_{k\in[K]}\bbP_{\bmu}^{\pi}\prn*{E_{j,k}}\ge 1.
\]
\end{lemma}

\cref{lm:first-classd}  leads to the following corollary, which will be utilized in subsequent subsections.

\begin{corollary}
\label{cor:first-second-classd}For any policy $\pi\in\Pi_S$, for any environment $\bmu$, we have
\[
\frac{1}{(q(S,\bm{c})+2)K}\sum_{j\in[q(S,\bm{c})+2]}\sum_{k\in[K]}\bbP_{\bmu}^{\pi}\prn*{E_{j,k}}\ge \frac{1}{(q(S,\bm{c})+2)K},
\]
\end{corollary}

\cref{cor:first-second-classd} tells us the following fact: under any \emph{single} environment $\bmu$, the average probability of the  risky events is $\wt{\Omega}\prn*{\frac{1}{K}}$.

In the rest of this subsection, we provide a proof for \cref{lm:first-classd}.

\proof{Proof of \pref{lm:first-class}.} We discuss two cases: $q(S,\bm{c})$ is odd, and $q(S,\bm{c})$ is even. Suppose that $q(S,\bm{c})$ is odd. For any $j\in\brk{q(S,\bm{c})}$, we have
\begin{align*}
\sum_{k\in[K]}\Ind\crl*{E_{j,k}}&=\text{number of actions that are not chosen in period }\brk*{t_{j-1}:t_{j}}\\
&\ge \Ind\crl*{\text{not all actions are chosen in period }\brk*{t_{j-1}:t_{j}}}\\
&= 1-\Ind\crl*{\text{all actions are chosen in period }\brk*{t_{j-1}:t_{j}}}
\end{align*}
almost surely. Thus for any $j\in\crl{1,3,\cdots,q(S,\bm{c})-2}$, we have
\begin{equation*}
\begin{aligned}
\sum_{k\in[K]}\bbP_{\bmu}^{\pi}\prn*{E_{j,k}}+\sum_{k\in[K]}\bbP_{\bmu}^{\pi}\prn*{E_{j+1,k}}& = \sum_{k\in[K]}\bbE_{\bmu}^{\pi}\brk*{\Ind\crl*{E_{j,k}}}+\sum_{k\in[K]}\bbE_{\bmu}^{\pi}\brk*{\Ind\crl*{E_{j+1,k}}}\\
& =  \bbE_{\bmu}^{\pi}\brk*{\sum_{k\in[K]}\Ind\crl*{E_{j,k}}}+\bbE_{\bmu}^{\pi}\brk*{\sum_{k\in[K]}\Ind\crl*{E_{j+1,k}}}\\
& \ge 1-\Ind\crl*{\text{all actions are chosen in period }\brk*{t_{j-1}:t_{j}}}\\
&~~~+1-\Ind\crl*{\text{all actions are chosen in period }\brk*{t_{j}:t_{j+1}}}\\
& \ge1-\Ind\crl*{\text{all actions are chosen in both period }\brk*{t_{j-1}:t_{j}} \text{ and period } \brk*{t_{j}:t_{j+1}}}\\
&=1-\Ind\crl*{S\prn*{t_{j-1}:t_{j+1}}\ge 2\Sigma-c^{(1)}-c^{(2)}}\\
&=\Ind\crl*{S\prn*{t_{j-1}:t_{j+1}}< 2\Sigma-c^{(1)}-c^{(2)}}.
\end{aligned}
\end{equation*}
Moreover, for $j=q(S,\bm{c})$, we can show that
\begin{align*}
    \sum_{k\in[K]}\bbP_{\bmu}^{\pi}\prn*{E_{q(S,\bm{c}),k}}\ge1-\Ind\crl*{S\prn*{t_{j-1}:t_{j+1}}\ge \Sigma-c^{(1)}}=\Ind\crl*{S\prn*{t_{j-1}:t_{j+1}}< \Sigma-c^{(1)}}.
\end{align*}
Therefore, by the pigeonhole principle and the definition of $q(S,\bm{c})$, we have
\begin{align*}
&~~~\sum_{j\in[q(S,\bm{c})]}\sum_{k\in[K]}\bbP_{\bmu}^{\pi}\prn*{E_{j,k}}\\& \ge 
\bbE_{\bmu}^\pi\brk*{ \sum_{j\in\crl{1,3,\cdots,q(S,\bm{c})-2}}\Ind\crl*{S\prn*{t_{j-1}:t_{j+1}}< 2\Sigma-c^{(1)}-c^{(2)}}+\Ind\crl*{S\prn*{t_{q(S,\bm{c})-1}:t_{q(S,\bm{c})+1}}< \Sigma-c^{(1)}}}\\
&{\ge}\bbE_{\bmu}^\pi\brk*{\prn*{1- \Ind\crl*{S\prn*{1:t_{q(S,\bm{c})}}\ge {q(S,\bm{c})}\Sigma-\ceil*{\frac{q(S,\bm{c})}{2}}c^{(1)}-\floor*{\frac{q(S,\bm{c})}{2}}c^{(2)}}}}\\
&\ge\bbE_{\bmu}^\pi\brk*{\prn*{1- \Ind\crl*{S\prn*{1:t_{q(S,\bm{c})}}>S-\Sigma}}}.
\end{align*}
Suppose that $q(S,\bm{c})$ is even. Then using similar arguments, we can still show that
\begin{align*}
\sum_{j\in[q(S,\bm{c})]}\sum_{k\in[K]}\bbP_{\bmu}^{\pi}\prn*{E_{j,k}}
\ge\bbE_{\bmu}^\pi\brk*{\prn*{1- \Ind\crl*{S\prn*{1:t_{q(S,\bm{c})}}>S-\Sigma}}}.
\end{align*}
Therefore, the above inequality always holds.

Now we define
\begin{align*}
E_{\sim,k} \ldef   \crl*{\text{action }k\text{ is not among the first }K-1\text{ (different) actions chosen in period }\brk*{t_{q(S,\bm{c})}:T}}.
\end{align*}
If both $\crl*{S\prn*{1:t_{q(S,\bm{c})}}>S-\Sigma}$ and $ E_{\sim,k} $ happen, then since $\tau$ is the first time that all actions of $[K]$ have been chosen after round $t_{q(S,\bm{c})}$, we know that  
either $E_{q(S,\bm{c})+1,k}=\crl*{\text{action }k \text{ is not chosen in period }\brk*{t_{q(S,\bm{c})}:\floor{\prn{t_{q(S,\bm{c})}+T}/2}}}$ happens, or $E_{q(S,\bm{c})+2,k}=\crl*{\tau\le \floor{\prn{t_{q(S,\bm{c})}+T}/2},\, a_\tau=k, \, {S\prn*{1:t_{q(S,\bm{c})}}>S-\Sigma}}$ happens.
Therefore, we know that
\[
E_{q(S,\bm{c})+1,k}\cup E_{q(S,\bm{c})+2,k} \ \supset\   E_{\sim,k} \cap \crl*{S\prn*{1:t_{q(S,\bm{c})}}>S-\Sigma}.
\]
This implies that
\begin{align*}
    \sum_{k\in[K]}\bbP_{\bmu}^{\pi}\prn*{E_{q(S,\bm{c})+1,k}\cup E_{q(S,\bm{c})+2,k} }& \ge\sum_{k\in[K]}\bbP_{\bmu}^{\pi}\prn*{E_{\sim,k} \cap \crl*{S\prn*{1:t_{q(S,\bm{c})}}>S-\Sigma}}\\
    &= \sum_{k\in[K]}\bbE_{\bmu}^\pi\brk*{\Ind\crl*{E_{\sim,k}}\Ind{\crl*{S\prn*{1:t_{q(S,\bm{c})}}>S-\Sigma}}} \\
    & =\bbE_{\bmu}^{\pi}\brk*{\sum_{k\in[K]}\Ind\crl*{E_{\sim,k}}\Ind{\crl*{S\prn*{1:t_{q(S,\bm{c})}}>S-\Sigma}}}\\
    &\overset{\rm (iii)}{\ge} \bbE_{\bmu}^{\pi}\brk*{\Ind{\crl*{S\prn*{1:t_{q(S,\bm{c})}}>S-\Sigma}}},
\end{align*}
where (iii) follow from the  definition of $E_{\sim,k}$.

Combining the above two paragraphs, we have
\begin{align*}
    \sum_{j\in[q(S,\bm{c})+2]}\sum_{k\in[K]}\bbP_{\bmu}^{\pi}\prn*{E_{j,k}}
    &\ge\sum_{j\in[q(S,\bm{c})]}\sum_{k\in[K]}\bbP_{\bmu}^{\pi}\prn*{E_{j,k}}+   \sum_{k\in[K]}\bbP_{\bmu}^{\pi}\prn*{E_{q(S,\bm{c})+1,k}\cup E_{q(S,\bm{c})+2,k} }\\
    &\ge\bbE_{\bmu}^\pi\brk*{\prn*{1- \Ind\crl*{S\prn*{1:t_{q(S,\bm{c})}}>S-\Sigma}}}\\
    &~~~+\bbE_{\bmu}^{\pi}\brk*{\Ind{\crl*{S\prn*{1:t_{q(S,\bm{c})}}>S-\Sigma}}}\\
    &=1.
\end{align*}
$\hfill\Box$

\subsection{Alternative Environments, Bad Events, and Lower Bound Reductions}\label{app:altd}

\underline{In the rest of the proof, we fix an arbitrary policy $\pi\in\Pi_S$.}

 For any $j\in[q(S,\bm{c})+2]$, define a reward gap
\[
\Delta_{j}\ldef\begin{cases}
1, &\text{if }j=1,\\
\frac{1}{{2(q(S,\bm{c})+2)}}\sqrt{\frac{{1}}{{t_{j-1}}}}, &\text{if }j\in[2:q(S,\bm{c})+1],\\
-\frac{1}{{2(q(S,\bm{c})+2)}}\sqrt{\frac{{1}}{{t_{q(S,\bm{c})}}}}, &\text{if }j=q(S,\bm{c})+2.
\end{cases}
\]
Note that $\abs*{\Delta_j}\in[0,1]$ for all $j\in[q(S,\bm{c})+2]$.

Let $\bmu=(0,\dots,0)\in\bbR^K$ be the \emph{reference} environment. Let $\bbQ\ldef\bbP_{\bm{0}}^\pi$ denote the \emph{reference} measure.

For any $j\in[q(S,\bm{c})+2], k\in[K]$, define an \emph{alternative} environment $\bmu_{j,k}\ldef \prn*{\mu_{j,k;1},\dots,\mu_{j,k;K}}\in\bbR^K$ where
\[
\mu_{j,k;i}\ldef\begin{cases}0+\Delta_j, &\text{if }i=k,\\
0, &\text{otherwise.}
\end{cases}
\]
Note that each alternative environment $\bmu_{j,k}$  only differs from the reference environment in terms of the mean reward of action $k$.

For any  $j\in[q(S,\bm{c})+2], k\in[K]$, let $\bbP_{j,k}\ldef\bbP^{\pi}_{\bmu_{j,k}}$ denote the \emph{alternative} measure associated with the alternative environment $\bmu_{j,k}$. 

We explicitly construct a class of environments $\Phi\ldef\crl*{\bmu_{j,k}\mid j\in[q(S,\bm{c})+2], k\in[K]}$.

For any  $j\in[q(S,\bm{c})+2], k\in[K]$, under environment $\bmu_{j,k}$, the risky event $E_{j,k}$ becomes a \emph{bad event} whose occurrence would lead to large regret. Specifically:
\begin{itemize}
    \item Suppose $j=1$. Since action $k$ is the unique optimal action under environment $\bmu_{1,k}$, choosing any action other than $k$ for one round incurs at least a $\Delta_1$ term in the policy's  regret, and the occurrence of $E_{1,k}=\crl*{\text{action }k \text{ is not chosen in period }\brk*{1:t_{1}}}$ incurs at least a ${t_1}\Delta_1$ term in the policy's  regret.
    \item Suppose $j\in[2:q(S,\bm{c})]$. Since action $k$ is the unique optimal action under environment $\bmu_{j,k}$, choosing any action other than $k$ for one round incurs at least a $\Delta_j$ term in the policy's  regret, and the occurrence of $E_{j,k}=\crl*{\text{action }k \text{ is not chosen in period }\brk*{t_{j-1}:t_{j}}}$ incurs at least a $\prn*{t_j-t_{j-1}+1}\Delta_j$ term in the policy's  regret.
    \item Suppose $j=q(S,\bm{c})+1$. Since action $k$ is the unique optimal action under environment $\bmu_{q(S,\bm{c})+1,k}$, choosing any action other than $k$ for one round incurs at least a $\Delta_{q(S,\bm{c})+1}$ term in the policy's  regret, and the occurrence of $E_{q(S,\bm{c})+1,k}=\crl*{\text{action }k \text{ is not chosen in period }\brk*{t_{q(S,\bm{c})}:\floor{\prn{t_{q(S,\bm{c})}+T}/2}}}$ incurs at least a $\prn*{\floor{\prn{t_{q(S,\bm{c})}+T}/2}-t_{q(S,\bm{c})}+1}\Delta_{q(S,\bm{c})+1}$ term in the policy's  regret.
    \item Suppose $j=q(S,\bm{c})+2$ and $k=1$. Since action 1 is the worst action under environment $\bmu_{q(S,\bm{c})+2,1}$, choosing action 1 for one round incurs at least a $-\Delta_{q(S,\bm{c})+2}-\frac{\Delta_{q(S,\bm{c})+1}}{2}=-\frac{\Delta_{q(S,\bm{c})+2}}{2}$ term in the policy's regret. Furthermore, by the switching constraint $S(1:T)\le S<(q(S,\bm{c})+1)H+\min_{j\ne 1}c_{1,j}$, 
    the occurrence of $E_{q(S,\bm{c})+2,1}$ implies the occurrence of $\crl*{\text{no switch happens after round } \tau}$ (as the remaining switching budget does not allow for switching from action 1), thus essentially implies the occurrence of  $\crl*{\text{action }1\text{ is chosen in every round in period }[\floor{\prn{t_{q(S,\bm{c})}+T}/2}:T]}$. As a result, the occurrence of  $E_{q(S,\bm{c})+2,1}$ incurs at least a $-\prn*{T-\floor{\prn{t_{q(S,\bm{c})}+T}/2}+1}\Delta_{q(S,\bm{c})+2}/2$ term in the policy's  regret.
    \item Suppose $j=q(S,\bm{c})+2$. Since action $k$ is the worst action under environment $\bmu_{q(S,\bm{c})+2,k}^{(1)}$, choosing action $k$ for one round incurs at least a $-\Delta_{q(S,K)+2}^{(1)}$ term in the policy's regret. Furthermore, since the occurrence of $E_{q(S,\bm{c})+2,k}^{(1)}$ implies the occurrence of $\crl*{\text{action }k\text{ is chosen in every round in }[\floor{\prn{t_{q(S,\bm{c})}^{(1)}+T}/2}:T]}$ (because of the switching constraint), it incurs at least a $-\prn*{T-\floor{\prn{t_{q(S,\bm{c})}^{(1)}+T}/2}+1}\Delta_{q(S,\bm{c})+2}^{(1)}$ term in the policy's  regret.
\end{itemize}
The above arguments lead to \cref{lm:badd}.

\begin{lemma}[From risky events to bad events]\label{lm:badd} For any $j\in[q(S,\bm{c})+2], k\in[K]$, under environment $\bmu_{j,k}$, the risky event $E_{j,k}$ becomes a \emph{bad event} in the sense that
\[\bbE_{\bmu_{j,k}}^{\pi}\brk*{T\mu_{j,k;k}-\sum_{t=1}^T\mu_{j,k;a_t} \mid E_{j,k}}\ge \cR_{\rm bad}\prn{S,G,T},\]
where \[\cR_{\rm bad}(S,G,T)\ldef\frac{{1}}{8\prn{q(S,\bm{c})+2}}{{T}^{\frac{1}{2-2^{-q(S,\bm{c})}}}}\]
is a universal lower bound on the ``distribution-dependent regret conditional on the bad event.''
\end{lemma}

\proof{Proof of \cref{lm:badd}.} The proof is the same as the proof of \pref{lm:bad}.
$\hfill\Box$

Based on \cref{lm:badd}, we can reduce the task of proving a lower bound on the policy's (distribution-dependent) regret $R_{\bmu_{j,k}}^{\pi}(T)$ to the task of proving a lower bound on the bad event probability $\bbP_{j,k}\prn*{E_{j,k}}$. Consequently, we can reduce the task of proving a lower bound on  $\sup_{\bmu\in\Phi}R^{\pi}_{\bmu}(T)$ to the task of proving a lower bound on the ``average-case bad event probability'' $\frac{1}{(q(S,\bm{c})+2)K}\sum_{j\in[q(S,\bm{c})+2]}\sum_{k\in[K]}\bbP_{j,k}\prn*{E_{j,k}}$.
\begin{lemma}[Reducing regret lower bounds to bad event probability lower bounds]\label{lm:reductiond} For any $j\in[q(S,\bm{c})+2], k\in[K]$, we have
\[
R_{\bmu_{j,k}}^{\pi}(T)\ge \cR_{\rm bad}(S,G,T)\cdot \bbP_{j,k}\prn*{E_{j,k}}.
\]
As a result, we have
\begin{align*}
    R^{\pi}(T)\ge \sup_{\bmu\in\Phi}R^{\pi}_{\bmu}(T)\ge\cR_{\rm bad}(S,K,T)\cdot\frac{1}{(q(S,\bm{c})+2)K}\sum_{j\in[q(S,\bm{c})+2]}\sum_{k\in[K]}\bbP_{j,k}\prn*{E_{j,k}}.
\end{align*}
\end{lemma}
\proof{Proof of \cref{lm:reductiond}.}
The proof is the same as the proof of \pref{lm:reduction}.
$\hfill\Box$

\subsection{Probability Space Changing Tricks}\label{sec:changingd}
\cref{lm:reductiond} indicates that, in order to prove the desired lower bound on the regret $R^\pi(T)$, it suffices to prove the following statement:
\begin{equation}\label{eq:obj1d}
    \wb{p}\ldef\frac{1}{(q(S,\bm{c})+2)K}\sum_{j\in[q''(S,K)+2]}\sum_{k\in[K]}\bbP_{j,k}\prn*{E_{j,k}}=\wt{\Omega}\prn*{\frac{1}{K}},
\end{equation}
That is, we only need to establish tight lower bounds on the average probability $  \wb{p}$.

By \cref{cor:first-second-classd}, we have
\[
\wb{q}\ldef\frac{1}{(q(S,K)+2)K}\sum_{j\in[q(S,K)+2]}\sum_{k\in[K]}\bbQ\prn*{E_{j,k}}\ge\frac{1}{(q(S,K)+2)K}=\wt{\Omega}\prn*{\frac{1}{K}},
\]
Therefore, in order to show \cref{eq:obj1d}, it suffices to show that $\wb{p}$ is close to $\wb{q}$. Similar to the proof of \pref{thm:lb-unit}, we apply the probability space changing tricks. The arguments are very similar to the arguments in \pref{sec:changing} and are omitted here.

\subsection{Applying the GRF Inequality}\label{app:applyd}

Similar to  \pref{app:apply}, by applying the \GRF inequality, we can show that
\begin{align}\label{eq:lbp1d}
    \wb{p}
    &{\ge}  \wb{q}-\sqrt{2\wb{q}\frac{\wb{q}}{4}}\notag\\
    &=\frac{2-\sqrt{2}}{2}\wb{q}.
\end{align}
Now we plug \pref{eq:lbp1d}  into \cref{lm:reductiond}. We have
\begin{align*}
    R^{\pi}(T)&\ge \sup_{\bmu\in\Phi}R^{\pi}_{\bmu}(T)\\
    &\ge\cR_{\rm bad}(S,G,T)\cdot\frac{1}{(q(S,K)+2)K}\sum_{j\in[q(S,K)+2]}\sum_{k\in[K]}\bbP_{j,k}\prn*{E_{j,k}}\\
    &\ge\cR_{\rm bad}(S,G,T)\cdot\frac{2-\sqrt{2}}{2}\wb{q}\\
    &= \frac{1}{8\prn{q(S,\bm{c})+2}}{{T}^{\frac{1}{2-2^{-q(S,\bm{c})}}}}\cdot\frac{2-\sqrt{2}}{2}\frac{1}{(q(S,K)+2)K}\\
    &=\Omega\prn*{\frac{1}{K\log T}T^{\frac{1}{2-2^{-q(S,\bm{c})}}}}.
\end{align*}
We thus complete the proof of the lower bound. $\hfill\Box$

\vspace{2em}
\noindent{\large \bf Part IV. Supplements for \pref{sec:conclusion}}

\section{On the Impact of Unknown Horizon}\label{app:unknown}
Consider the standard \MAB problem. %
An algorithm is called \emph{anytime} if it does not require the knowledge of $T$. An anytime algorithm $\phi$ is called a \emph{sublinear-rate}  algorithm, if and only if there exist constants $C>0, \delta>0$ such that $R^{\phi}(T)\le C\sqrt{K}T^{1-\delta}$ for all $T\ge K$. Note that ensuring a sublinear regret rate is a very basic requirement --- an anytime algorithm failing to meet this requirement is not really meaningful (at least when we use regret as the performance metric). 
The following proposition shows that any anytime algorithm with a fixed switching budget independent of $T$ \emph{cannot} be a sublinear-rate algorithm.

\begin{proposition}\label{prop:unknown}Let $K\in\bbN_{>0}$ and $S>0$ be arbitrary constants. 
Let $\phi$ be an anytime algorithm whose expected number of switches under any environment is always no larger than $S$ (this condition is weaker than the switching constraint). Then $\phi$ cannot be a sublinear-rate algorithm.
\end{proposition}

The proof of \pref{prop:unknown} builds on the proof of \pref{thm:lb-unit}, but is much easier because \pref{prop:unknown} does not require precise control of the incurred switching cost --- in fact, \pref{prop:unknown} holds even if one changes the switching budget within a constant factor. The technique to deal with an unknown $T$ builds on Theorem 3 of \cite{dong2020multinomial}, which is a result similar to \pref{prop:unknown} but  proved under a different problem setting (the MNL bandit problem) with slightly different focus (i.e., it focuses on regret rates ranging from $T^{2/3}$ to $T^{1/2}$). The proof of \pref{prop:unknown} can be found in \pref{app:subunknown}.

We make two remarks. First, \pref{prop:unknown} indicates that an anytime algorithm  cannot be really meaningful (i.e., attain a sublinear regret rate) if its switching budget does not grow with $T$. This shows that the knowledge of $T$ is somehow necessary if we do not want the switching budget to grow with $T$.

Second, a simple adaption of the proof of \pref{prop:unknown} indicates that $\Omega(\frac{K\log T}{\log\log T})$ switches are necessary for achieving $\wt\cO({\sqrt{KT}})$ regret in the classical \MAB problem without the knowledge of $T$.  Note that the celebrated \textsf{UCB2} algorithm (\citealt{auer2002finite}) achieves $\wt\cO({\sqrt{KT}})$ regret in the classical \MAB problem with $\cO(K\log T)$ switches, and it is an anytime algorithm. Obtaining a tighter characterization on the trade-off between the switching budget (defined as an exact function of $T$) and the optimal regret rate without knowing $T$ is an interesting future direction.

\subsection{Proof of \pref{prop:unknown}.}\label{app:subunknown}
Assume that there exist constants $C>0, \delta>0$ such that $R^\phi(T)\le C\sqrt{K}T^{1-\delta}$ for all $T\ge K$. Let $\eps<\delta$ be a constant, and let $T_0\in\bbN_{>0}$ be the smallest integer such that $\frac{1}{64}\sqrt{K}T^{1-\epsilon}>C\sqrt{K}T^{1-\delta}$. Our goal is to find a finite $T\ge T_0$ such that $R^{\phi}(T)\ge \frac{1}{48}\sqrt{K}T^{1-\eps}$ (then there is a contradiction).

Let $\eta\ldef \frac{1}{2\eps}-1$. 
Without loss of generality, assume $T_0\ge\max\crl*{ 2^{\frac{1}{\eta}},(1+\eta)^2}$.  Let $q'\ldef \ceil{8S/K}\ge 1$. Define a sequence $(T_j)_{j=0}^{q'}$ such that $T_j=\floor*{(T_{j-1})^{{\frac{1}{2\eps}}}}$ for all $j\in[1:q']$ (since $q'$ is a constant, this sequence is well-defined). We have $T_{q'}\le {T_0}^{\frac{1}{2^{q'}\eps^{q'}}}$ and $T_j\ge 2 T_{j-1}-1$ for all $j\in[q']$. Moreover, for all $j\in[q']$,
\begin{equation}\label{eq:seq}
\frac{T_j-T_{j-1}+1}{\sqrt{T_{j-1}/K}}\ge \frac{(T_j+1)/2}{\sqrt{(T_j+1)^{2\eps}/K}}\ge\frac{1}{2}\sqrt{K}(T_j)^{1-\eps}.
\end{equation}

We adopt the notation system introduced in \pref{app:lower}. Below is a simple application of \lbargu.

For any $T\in[T_0:T_{q'}]$, under environment $\bm{0}$, algorithm $\phi$'s expected number of switches is no larger than $S\le{Kq'}/{6}$. Thus there must exist $q_0\in[q']$ such that algorithm $\phi$'s expected number of switches  in period $[T_{q_0-1}:T_{q_0}]$ under environment $\bm{0}$ is no larger than $K/6$. We then know that $\bbP_{\bm{0}}^{\phi}\prn*{S(T_{q_0-1}:T_{q_0})\le K/3}\ge\frac{1}{2}$. We define a class of risky events as follows: for any $k\in[K]$, define
$E_{k}\ldef\crl*{\text{action }k \text{ is not chosen in period }\brk*{T_{q_0-1}:T_{q_0}}}$. 
By algebra similar to \pref{eq:lm7}, we have \[\sum_{k\in[K]}\bbP_{\bm{0}}^\phi(E_{j,k})\ge \prn*{K-1-\floor*{\frac{K}{3}}}\bbE_{\bm{0}}^\phi\brk*{\Ind\crl*{S(T_{q_0-1}:T_{1_0})\le\frac{K}{4}}}\ge\frac{1}{2}\prn*{K-{\frac{2K}{3}}}\ge\frac{K}{6}.\]

Let $\bm{0}$ be the reference measure. Define $K$ alternative environments $\bmu_1,\dots\bmu_K\in\bbR^K$ such that
\[
\mu_{k;i}\ldef\begin{cases}\frac{1}{6}\sqrt{\frac{K}{T_{q_0-1}}}, &\text{if }i=k,\\
0, &\text{otherwise.}
\end{cases}
\]
Using \pref{eq:seq}, we can easily show that
\[
R^{\pi}(T_{q_0})\ge\frac{1}{2}\sqrt{K}(T_j)^{1-\eps}\cdot{\frac{1}{K}\sum_{k\in[K]}\bbP_{\bmu_k}^\phi(E_{j,k})}.
\]

For any $k\in[K]$, let $\bbP'_k$ be the probability measure induced by the joint random variable
\begin{equation*}
\prn*{\prn*{a_t,X_{\bmu_{k}}^t(a_t)}_{t\in[1:T_{q_0-1}-1]},\prn*{a_t,X_{\bmu_{k}}^t(a_t)\Ind\crl*{a_t\ne k} }_{t\in[T_{q_0-1}:T_{q_0}]}}.
\end{equation*}
We have $\sum_{k\in[K]}D_{\rm re}(\bbP_k'\parallel \bbP_{\bm{0}}^{\phi})=\sum_{k\in[K]}D_{\rm KL}(\bbP_k'\parallel \bbP_{\bm{0}}^{\phi})\le\frac{1}{2}\prn*{\frac{1}{6}\sqrt{\frac{K}{T_{q_0-1}}}}^2T_{q_0-1}=\frac{K}{72}$. By the \GRF inequality, we have 
\[
{\frac{1}{K}\sum_{k\in[K]}\bbP_{\bmu_k}^\phi(E_{j,k})}=\frac{1}{K}\sum_{k\in[K]}\bbP_{k}'(E_{j,k})\ge\frac{2-\sqrt{2}}{2}\frac{1}{K}\sum_{k\in[K]}\bbP_{\bm{0}}^\phi(E_{j,k})\ge\frac{2-\sqrt{2}}{12}\ge\frac{1}{24}.
\]
Thus $R^\pi(T_{q_0})\ge\frac{1}{48}\sqrt{K}(T_{q_0})^{1-\eps}$ and we are done. $\hfill\Box$ %

\section{On the Optimal Regret Rate of G-BwSC}\label{app:tight}

Recall that our upper and lower bounds in \pref{sec:asym} do now always match. A natural question is whether one of them is tight. In this section, we show that neither of the bounds are tight in $T$ for general $G$ (even for a fixed $K$). Specifically, we establish  \pref{thm:exist}, which  successfully recovers the exact dependency on $T$ of the optimal regret of \GB when $K$ is a fixed constant. However, this result  is only of theoretical interest, for two reasons. First, the gap between the upper and lower bounds employed in our proof is bad --- it is of the order of $(k!)^2$. Also, computing the key index $q(S,G)$ is difficult. Obtaining bounds with better dependence on $K$ and finding better interpretations of the index are  interesting open problems.

\begin{theorem}\label{thm:exist}Let $K\ge1$ be an arbitrary given constant. For any given switching graph $G$ such that $|G|=K$, for any $S\ge0$,
$$R_S^*(T)=\wt{\Theta}(T^{\frac{1}{2-2^{-q(S,G)}}}),$$
where $q(S,G)$ is an integer completely determined by the switching graph $G$. %
Moreover, there exist $S,G$ such that $q(S,G)\ne q'(S,G)$, and there exit $S,G$ such that $q(S,G)\ne q''(S,G)$.
\end{theorem}

In \pref{app:index}, we give a more detailed description on the key index $q(S,G)$.

\subsection{The Key Index $m_G(S)$}\label{app:index}

We start with the following definition.

\begin{definition} (Adversarial TSP)
Given a switching graph $G$ and a starting vertex $i\in[K]$, the adversarial traveling salesman problem is defined as follows. There is a destination set $U$ and a visited vertices set $V$. Initially, $U=[K]$ and $V=\{i\}$. An agent starts a walk on $G$ from vertex $i$. Each time she visits a new vertex $j\notin V$, $V$ becomes $V\cup\{j\}$, then an adversary selects one vertex $l\in U\cap V$ and eliminates $l$ from $U$, i.e., $U$ becomes $U\setminus\{l\}$. The agent then decides her next visit based on $U$ and $V$. Obviously, when the agent has visited all vertices, i.e., $V=[K]$, there will be only one vertex left in the destination set $U$. The agent is required to first make $V=[K]$ then ends her walk at the final vertex in $U$. The adversarial traveling salesman problem seeks to minimize the worst-case length of the agent's walk, whose optimal value is denoted by $W_{G,i}^{(1)}$. 
\end{definition}

The adversarial TSP is a generalization of the classical TSP. When the adversary is known to always eliminate the newly visited vertex from $U$, the agent solves exactly a TSP. However, since the adversarial elimination rule is not known in advance, the agent has to consider the worst case among all adversarial elimination rules. Thus, the solution to the classical TSP (i.e., the length of the shortest Hamiltonian cycle in $G$) serves as a natural lower bound for $W_{G,i}^{(1)}$ for all $i\in[k]$. Meanwhile, $H+\max_{i,j\in[k]}c_{i,j}$ serves as a natural upper bound for $W^{(1)}_G:=\min_{i\in[k]}W_{G,i}^{(1)}$. This implies that $W^{(1)}_G\in[H+\min_{j\ne i}c_{i,j},H+\max_{i,j}c_{i,j}]$. %

The adversarial TSP is a pure computational problem that can be solved by a dynamic program (still, it is NP-hard). Specifically, let $W_{G,i}^{U,V}$ denote the minimum worst-case length of the agent's \textit{remaining} walk when her destination set is $U$, her visited vertices set is $V$, and her current position is $i\in V$, then the solution to the adversarial TSP can be found by calculating $W_{G,i}^{(1)}=W_{G,i}^{[K],\{i\}}$ via the following dynamic program:
\begin{equation}\label{eq:dp1}
W_{G,i}^{U,V}=\min_{j\in{[K]\setminus{V}}}\left\{c_{i,j}+\max_{l\in U \cap(V\cup\{j\})}W_{G,j}^{U\setminus\{l\},V\cup\{j\}}\right\},~~~~\forall i\in V, V\ne[K].
\end{equation}
$$
W_{G,i}^{\{j\},[K]}=c_{i,j},~~~~\forall i,j\in[K].
$$

Given the value of $W_{G,i}^{(1)}$ for $i\in[K]$, we further define a series of quantities via dynamic programming. Let $H_{i,j}$ denote the length of the shortest Hamiltonian path in $G$ that starts from $i$ and ends at $j\ne i$. We calculate
\begin{equation}\label{eq:dp2}
W_{G,i}^{(n)}=\min\limits_{j\in [K]\setminus{\{i\}},l\in[K]} \left\{H_{i,j}+c_{j,l}+W_{G,l}^{(n-1)}\right\},~~~~\forall i\in[K], n=2,3\dots
\end{equation}
and define $W_G^{(n)}:=\min_{i\in[K]}W_{G,i}^{(n)}$ for $n\in\mathbb{Z}_{\ge1}$. Therefore, for any given switching graph $G$, by the above procedure, we obtain a series of quantities
$$
W_G^{(1)}\le W_G^{(2)}\le W_G^{(3)}\le\cdots
$$
 that are completely determined by $G$. 
Let $W_G^{(0)}=0$. For any $S\ge0$, we define
$$
q(S,G):=\sup\{n\mid S\ge W_G^{(n)}\}.
$$
This is the key index stated in \pref{thm:exist}.

\end{appendices}

\end{document}